\documentclass[11pt]{article}

\usepackage{tikz}
\usetikzlibrary{arrows.meta, positioning}

\usepackage{amssymb}
\usepackage{amsmath,amsfonts,amsthm}
\usepackage{subcaption}
\usepackage{multirow}
\usepackage{graphicx}
\usepackage{caption}
\usepackage{float}
\usepackage{bm}
\usepackage{amssymb}
\usepackage[round]{natbib}
\usepackage{url}
\usepackage[normalem]{ulem}
\usepackage{cases}
\usepackage[colorlinks,linkcolor=red,citecolor=blue,urlcolor=magenta]{hyperref}
\usepackage{hyperref}       % hyperlinks
\usepackage{url}            % simple URL typesetting
\usepackage{booktabs}       % professional-quality tables
\usepackage{amsfonts}       % blackboard math symbols
\usepackage{nicefrac}       % compact symbols for 1/2, etc.
\usepackage{xcolor}         % colors
\usepackage{amsmath}
\usepackage{multirow}
\usepackage{graphicx}
\usepackage{amsthm}
\usepackage{makecell}
\usepackage{algorithm}
\usepackage{algorithmic}
\usepackage{cases}
\usepackage{rotating}
\usepackage{comment}
\usepackage{varwidth}
\newtheorem{assumption}{Assumption}

\newtheorem{proposition}{Proposition}
\newtheorem{remark}{Remark}
\newtheorem{lemma}{Lemma}
\newtheorem{theorem}{Theorem}

\newcommand{\mE}{\mathbb{E}}
\newcommand{\tbf}{\textbf}
\newcommand{\filterF}{\mathcal{F}}
\newcommand{\mR}{\mathbb{R}}
\newcommand{\bX}{\mathbf{X}}
\newcommand{\bY}{\mathbf{Y}}
\newcommand{\bZ}{\mathbf{Z}}

\newcommand{\bx}{\mathbf{x}}

\newcommand{\bu}{\mathbf{u}}

\newcommand{\by}{\mathbf{y}}
\newcommand{\bz}{\mathbf{z}}
\newcommand{\bc}{\mathbf{c}}

\newcommand{\bW}{\mathbf{W}}
\newcommand{\bM}{\mathbf{M}}
\newcommand{\bA}{\mathbf{A}}
\newcommand{\bSimga}{\boldsymbol{\Sigma}}
\newcommand{\bB}{\mathbf{B}}
\newcommand{\bC}{\mathbf{C}}
\newcommand{\bU}{\mathbf{U}}

\newcommand{\tar}{{\text{target}}}
\newcommand{\cov}{{\text{Cov}}}

\newcommand{\genX}{\hat\bX}
\newcommand{\genP}{\hat P}
\newcommand{\KL}{\text{KL}}
\newcommand{\TV}{\text{TV}}
\newcommand{\intd}{\mathrm{d}}
\allowdisplaybreaks
\title{Generating solution paths of Markovian stochastic differential equations using diffusion models\footnote{A preliminary version of this work, titled ``Data-driven generative simulation of SDEs using diffusion model'' \citep{gao2025datadriven}, was presented at the NeurIPS 2025 MLxOR Workshop: Mathematical
Foundations and Operational Integration of Machine Learning for Uncertainty-Aware Decision-Making}.}

\author{
	Xuefeng Gao\thanks{Department of Systems Engineering and Engineering Management, The Chinese University of Hong Kong, Hong Kong, China. E-mail: xfgao@se.cuhk.edu.hk}
	\and
	Jiale Zha\thanks{Department of Systems Engineering and Engineering Management, The Chinese University of Hong Kong, Hong Kong, China. E-mail: jialezha@link.cuhk.edu.hk }
	\and
	Xun Yu Zhou\thanks{Department of Industrial Engineering and Operations Research and The Data Science Institute, Columbia University, New York, NY 10027, USA. Email: xz2574@columbia.edu}
}

\begin{document}

\maketitle

\begin{abstract}
   This paper introduces a new approach to generating sample paths of unknown Markovian stochastic differential equations (SDEs) using diffusion models, a class of generative AI methods commonly employed in image and video applications. Unlike the traditional Monte Carlo methods for simulating SDEs, which require explicit specifications of the drift and diffusion coefficients, ours takes a model-free, data-driven approach. Given a finite set of sample paths from an SDE, we utilize conditional diffusion models to generate new, synthetic paths of the same SDE. Numerical experiments show that our method consistently outperforms two alternative methods in terms of the Kullback--Leibler (KL) divergence between the distributions of the target SDE paths and the generated ones. Moreover, we present a theoretical error analysis deriving an explicit bound on the said KL  divergence.
    Finally, in simulation and empirical studies,  we leverage these synthetically generated sample paths to boost the performance of reinforcement learning algorithms for continuous-time mean--variance portfolio selection, hinting promising applications of our study in financial analysis and decision-making.
\end{abstract}


\section{Introduction}

Stochastic differential equations (SDEs) are an important class of equations for continuous-time stochastic models that have been widely employed in numerous fields, including finance, physics, operations research and machine learning. In many applications, generating sample paths of SDEs is crucial \citep{glasserman2004monte, kloedennumerical, songscore}. When an underlying SDE is known and given, i.e., its drift and diffusion coefficients are specified, Monte Carlo simulation is the classical approach for sample path generation with various methods proposed and developed in the literature; see e.g.  \cite{kloedennumerical} and the references therein. This paper instead studies the problem of simulating a Markovian SDE (i.e., the coefficients of the equation depend on the current, instead of the historical, state of the solution) when the underlying equation is {\it unknown}, with access only to a finite set of sample paths from the unknown SDE. Beyond theoretical curiosity, such a problem is motivated by many applications where one discretely observes only collections of time series whose dynamics can be reasonably captured by Markovian, otherwise totally unknown, SDEs.

Traditional approaches to generating paths from an unknown SDE largely rely on {\it model-based} methods, where the SDE is assumed to belong to a specific class such as geometric Brownian motions. Parameters of the equation are then estimated using statistical inference techniques \citep{iacus2008simulation, kessler2012statistical}. More recent studies use neural networks to represent drift and diffusion coefficients of SDEs, and construct appropriate loss functions to train these networks; see e.g. \cite{kidger2021neural, zhu2024dyngma}. Once such a neural SDE is learned, additional sample paths can then be generated from it by Monte Carlo. On the other hand, several recent papers explore {\it model-free} methods, leveraging generative models to generate samples directly from unknown SDEs, skipping the step of learning the SDEs themselves.\footnote{This is in the same spirit as in image generation, where one generates more cat pictures based on a set of sample cat pictures without attempting to learn the underlying unknown distribution of the samples.}  For instance, \cite{van2023monte} utilizes GANs \citep{goodfellow2014generative} to simulate unknown SDEs, while \cite{yang2024pseudoreversible} employs normalizing flows to generate samples from SDEs.

In this paper, we introduce a new model-free (up to the Markovian, diffusion SDE structure), data-driven approach to generate sample paths of SDEs using {\it conditional} diffusion models, a cutting-edge class of generative AI models that have achieved remarkable success in image and video applications whose performances surpass those of GANs and other generative models \citep{dhariwal2021diffusion, ho2020denoising, rombach2022high}.
The essential idea of diffusion models is to use a forward process to gradually turn the unknown target distribution into a simple noise distribution, and then reverse this process to generate new samples from the target distribution. 
Conditional diffusion models, in particular, can generate data samples that are conditioned on specific input variables; see Section~\ref{sec:CDM-review} for a concise review.
Employing  conditional diffusion models, this paper makes the following main contributions:
\begin{itemize}
\item First, we propose to train a sequence of conditional diffusion models to generate new sample paths from an unknown SDE, given a finite set of discrete-time observations (time series) of the SDE. The key observation is that for {\it Markovian} SDEs, the distribution of the next stochastic increment depends solely on the current state and time.
Thus, the state and time naturally become the conditioning input variables, and we train conditional diffusion models to
generate a sample of the next increment. In a sequential and autoregressive fashion, we paste these increment samples over time to generate a new sample path from the unknown SDE.

\item Second, we demonstrate the effectiveness of our approach by conducting experiments on various types of SDEs and comparing its performance with two benchmark methods including the neural SDE method in \cite{kidger2021neural} and a recent alternative generative approach in \cite{liu2025training}. We show that
our approach consistently and significantly outperforms the two benchmarks across nearly all experiments, in terms of the Kullback--Leibler (KL) divergence between the generated synthetic SDE paths and the given sample training paths.

\item Third, we carry out a rigorous theoretical analysis on the errors associated with our approach. Specifically, we establish an explicit bound on the KL divergence between the target and generated SDE path distributions. As a by-product, this KL bound also yields a bound on the total variation distance via Pinsker's inequality. The resulting bound offers a quantitative measure of the accuracy of our method. To our best knowledge, this is the first theoretical error bound on SDE path generation using conditional diffusion models.

\item Finally, to illustrate the potential applications of our framework in decision making, we carry out both a simulation study and an empirical one to demonstrate how we can make use of the synthetically generated sample paths
to enrich the dataset for training model-free reinforcement learning (RL) algorithms in continuous-time mean--variance portfolio selection \citep{ jia2023q} and to enhance their performance. Both studies show that
when synthetic paths are added in the RL training process, the resulting investment policies achieve the highest Sharpe ratios in the out-of-sample test, primarily due to reductions of the terminal wealth variances.  Moreover,
 our empirical study reveals that incorporating synthetic data does {\it not} enhance the model-based or plug-in policies, where one first estimates the model parameters and then plugs the estimators into the optimal policies for the portfolio selection problem. The implications from these results are that synthetic paths do not improve the accuracy of estimating model parameters because the generated paths do not contain more information, probabilistically or statistically, than the original sample paths, but they enable the RL agent to explore more possible market scenarios (even if fictitious) and thereby to learn more robust investment strategies. So, {\it AI-generated paths do not help us understand the environment better, but do help us perform better.}

\end{itemize}

This paper is related to a variety of topics in literature.
In terms of SDE path generation, it is closely related to \cite{liu2025training}, as both use conditional diffusion models to autoregressively sample SDE increments.
The crucial difference lies in the estimation of the score function, a key component of the diffusion models \citep{ho2020denoising, songscore}. While \cite{liu2025training} design a training-free Monte Carlo empirical estimator for the score
function and show that their approach outperforms GANs, we opt for training score neural networks which offer significantly greater scalability as demonstrated in our experiments. Moreover, our work differs from \cite{liu2025training} in two other important aspects: (1) we provide a rigorous error analysis for our diffusion-based generative method; and (2) we explore the practical applications of SDE path generation to portfolio selection problems in terms of improving decision making performance. Such an error analysis and applications to real-world problems were absent in \cite{liu2025training}.

For error analysis, our work is related to a growing body of literature on convergence theory and error bounds for {\it general} diffusion models;  see, e.g., \cite{bentonnearly, chen2023improved, conforti2025kl, chen2023sampling, fu2024unveil, gao2025wasserstein, li2024towards, li2025d, tang24} and the references therein. These studies derive various bounds on errors between target distributions and generated ones, informing the accuracy of their methods in modeling and generating {\it static} data distributions.
However, they do not consider the specific problem of SDE path generation and hence the resulting error propagation issue that arises from applying conditional diffusion models in a {\it dynamic} and auto-regressive manner. Consequently, our error analysis diverges substantially from these prior papers.

When it comes to financial applications, our research is related to two recent working papers \citep{chen2025diffusion, aghapour2025solving}. \cite{chen2025diffusion} apply unconditional diffusion models to simulate high-dimensional asset returns and conclude that, in their specific setting,  generated data improve the accuracy of parameter estimation (e.g., mean and covariance) and thus help solve a static portfolio optimization problem in a model-based manner.\footnote{On p. 22 of \cite{chen2025diffusion}, it is stated that ``This result reflects improvements in both the mean and covariance estimation from diffusion-generated
data, but most of the improvements come from the improved covariance estimation, which is not
surprising given the very design of our diffusion factor model." However, the authors have not offered an intuitive explanation as to why generated data provide more information than the observed ones, leading to better estimation.} We take a different direction -- we consider continuous-time portfolio selection with model-free RL and enrich the training dataset with synthetic paths. The implication of our results is quite different from that of \cite{chen2025diffusion}: at least in our setting, synthetic paths help RL policies but not plug-in ones, which has simple intuitions as discussed earlier.
When finalizing our paper, we became aware of a recent working paper \cite{aghapour2025solving}, which applies diffusion models to discrete-time mean--variance portfolio selection. Specifically, the authors directly model and generate stock price time series using conditional diffusion models, and then utilize the resulting synthetic data as a training environment for a policy gradient algorithm (known as TD3). This seems similar to our approach; however, there are differences in several aspects. They concentrate on discrete-time portfolio selection problems without involving  SDEs, whereas we focus on the SDE generation task and demonstrate its application through continuous-time portfolio selection. Moreover, their error bounds are established in a variant of the 2-Wasserstein distance, while ours are in terms of KL divergence.

Lastly, our work is related to a broader and more extensive literature on time series generation or forecasting using generative models such as GANs and diffusion models; see, just to name a few,  \cite{narasimhan2024time, liao2024sig, wiese2020quant, yoon2019time}. Due to the absence of the Markov property in general time series, conditioning on the entire past history or its embedded representation is often required for effective generation and forecasting. Although these studies report promising empirical results, they often fall short of rigorous theoretical analyses of the associated errors. By contrast, our work provides a distinctive contribution by focusing on the generation of sample paths from Markovian SDEs with theoretical guarantees. Furthermore, we showcase the application of our results to RL for dynamic portfolio optimization, an important problem in finance where it is standard to use Markovian SDEs to model dynamic asset prices.

The remainder of the paper is organized as follows. Section~\ref{eq:problem} formulates the problem, outlines the diffusion-based methodology for SDE path generation, and presents the related algorithms. Section~\ref{sec: numerical_experiment} is devoted to numerical experiments on four different types of SDEs, including a high-dimensional equation, to compare the performances between our method and two benchmark ones.  Section~\ref{sec:error} focuses on a theoretical error analysis of our SDE generative approach. In Section~\ref{sec:application}, we present an application to continuous-time mean--variance portfolio selection and conduct both simulated and empirical experiments. Finally, Section~\ref{sec:conclusion} concludes. Proofs and other supplementary materials are placed in the appendices.

\section{Problem Formulation and Methodology}\label{eq:problem}

\textbf{Notation}. We adopt the following notation throughout this paper.
The identity matrix in $\mathbb{R}^d$ is denoted by $I_d$, where $d$ is the dimension.
We use $\mathcal{N}(\boldsymbol{\mu}, \boldsymbol{\Sigma})$ to denote the multivariate Gaussian distribution with mean vector $\boldsymbol{\mu}$ and covariance matrix $\boldsymbol{\Sigma}$, and $\mathcal{N}(\cdot; \boldsymbol{\mu}, \boldsymbol{\Sigma})$
the corresponding Gaussian density function.
We denote by $\|\cdot\|$ the $L_2$ norm/spectral norm of vectors/matrices, and by $\|\cdot\|_F$ the Frobenius norm of matrices. A function  $f$ is called $L$-Lipschitz, where $L>0$, if $||f( \bx) - f( \bx')||_2 \le L||\bx - \bx'||_2$ for any $\bx,\bx'$.
We employ the notation $x \lesssim y$ to indicate that $x \leq Cy$ for some positive constant $C > 0$, and define $x \gtrsim y$ analogously.
Furthermore, we write $x \asymp y$ if there exist constants $C_1, C_2 > 0$ such that $C_1y \leq x \leq C_2y$.
For a random variable $\bX$, its distribution is denoted by $P_\bX$. We write
$\bX \overset{d}{=} \bY$ if $\bX$ and $\bY$ have the same distribution, and $P_\bX \ll P_\bY$ if $P_\bX$ is absolutely continuous with respect to (w.r.t.) $P_\bY$. Two distributions $P_\bX$ and $P_\bY$ are said to be mutually absolutely continuous, or equivalent, if $P_\bX \ll P_\bY$ and $P_\bY \ll P_\bX$. The Kullback--Leibler (KL) between probability distributions $P_\bX$ and $P_\bY$ is
   $ \KL(P_\bX || P_\bY) := \int \log(\frac{\intd P_\bX}{\intd P_\bY}) \intd P_\bX$ if $P_\bX \ll P_\bY$, and is defined to be $+\infty$ otherwise.

\subsection{Problem Formulation}

Consider a $d$-dimensional \textit{target} SDE defined on a filtered probability space $\big(\Omega, \filterF, \mathbb{P}, \{\filterF_t\}_{0\le t\le T}\big)$:\begin{equation}\label{eq: target_SDE}
	\intd\bX(t) = \mu(t, \bX(t))\intd t + \sigma(t, \bX(t))\intd\bW(t),\ t \in [0, T],
\end{equation}
where $\mu: [0, T] \times \mR^d \to \mR^d$ is the drift coefficient, $\sigma: [0, T] \times \mR^d \to \mR^{d\times m}$ is the diffusion coefficient, and $\bW$ is a standard $m$-dimensional Brownian motion. We assume that the equation starts from a fixed, known initial state $\bX(0)\equiv \bx(0)\in\mR^d$ and has a unique strong solution. However, both functions $\mu$ and $\sigma$ are \textit{unknown,} and we have access only to a set of i.i.d. sample paths of the SDE \eqref{eq: target_SDE} on $[0, T]$. Our goal, in mathematical terms,  is to generate additional sample paths that exhibit the same distributional properties as those of the {\it unknown} SDE \eqref{eq: target_SDE}.

In practice, one only observes or generates discrete-time states of \eqref{eq: target_SDE}. Without loss of generality, we consider a uniform grid $\mathcal{T} := \{t_n: t_n = n\Delta t\ \text{for}\ n=0, 1, ..., N_T\}$ with a fixed time step $\Delta t > 0$ and assume that the number of discrete epochs $N_T = T / \Delta t$ is an integer.\footnote{Our approach applies to the case  when the initial condition $\bX(0)$ follows an unknown distribution and the time grid $\mathcal{T}$ is non-uniform. In particular, in Example 2 of the numerical study (Section~\ref{sec: numerical_experiment}), we consider a non-uniform time grid $\mathcal{T}$. However,  for ease of presentation, we will focus mainly on the uniform time grid throughout the paper.}  The dataset available to us, therefore,  consists of $H > 1$ sample paths of $\{\bX(t): t \in [0, T]\}$ observed at grid points in $\mathcal{T}$ (i.e. time series), denoted by
\begin{align}\label{eq: obs_data}
	\left\{\bx^{(i)}(t_0), \bx^{(i)}(t_1), \cdots, \bx^{(i)}(t_{N_T}),\quad i=1, ..., H\right\}.
\end{align}
Our aim is to generate more time series at the same grid points of the sample paths of \eqref{eq: target_SDE} that are distributionally as close to those of \eqref{eq: obs_data} as possible.\footnote{While the original goal was to generate paths following the same distributions of the solution to \eqref{eq: target_SDE}, the only data available to us are the collection of series in \eqref{eq: obs_data}. Hence, to generate new series close to \eqref{eq: obs_data} is the best we could do. This is analogous to generating more cat pictures based on a set of existing cat pictures, a typical generative AI (GenAI) task.}
 To do so, we leverage the Markov property of the SDE \eqref{eq: target_SDE} and train conditional diffusion models to simulate the stochastic {\it increments} of \eqref{eq: target_SDE}. Note that $\Delta t$ is not necessarily small in many applications, in which case the resulting stochastic increments of the SDE  are generally non-Gaussian. This rules out the method of directly sampling Gaussian increments, a generally easy endeavor even in high dimensions. On the other hand, in high-dimensional settings where $d$ is large, classical density estimation methods such as kernel density estimation become impractical, begging for alternative generative models like ours.

\subsection{Review of Conditional Diffusion Models}\label{sec:CDM-review}

For the reader's convenience, we now give a brief review of the conditional diffusion models, based on the \textit{continuous-time score-based diffusion models} in \cite{songscore}.
Let $\bC$ be an unknown random variable (e.g., a covariate). Conditional diffusion models aim to generate new samples from an unknown (conditional) target distribution $p_\tar(\cdot | \bc):= q_0(\cdot | \bc) $ on $\mR^{d_\tar}$ given $\bC = \bc \in \mR^{d_c}$,
when one has access to a set of independent training data $\{(\bc_i, \by_i(0)): \bc_i\sim\bC, \by_i(0)\sim q_0(\cdot | \bc_i), i=1,\cdots N\}$. To achieve this, the conditional diffusion model first noises the target distribution with a (non-homogeneous) Ornstein--Uhlenbeck (OU) process:
\begin{align}\label{eq: forward_SDE_DDPM}
    \intd\bY(\tau) = -f(\tau)\bY(\tau)\intd\tau + g(\tau) \intd\bB(\tau),\ \bY(0)\sim q_0(\cdot | \bc),\ \tau\in [0, T_g],
\end{align}
where $\bB$ is a standard Brownian motion in $\mR^{d_\tar}$ that is independent of $\bY(0)$, and $T_g$ is a fixed horizon length (e.g. $T_g =1$). Throughout this paper, we
consider the popular choice of $f(\tau): = \frac{1}{2}\beta(\tau),\ g(\tau): = \sqrt{\beta(\tau)},\ \beta(\tau) := a\tau + b$, where $a$ and $b$ are pre-specified positive constants \citep{ho2020denoising, songscore}. The noising process \eqref{eq: forward_SDE_DDPM} is called the \textit{forward process} which admits an analytical solution
\begin{align*}%\label{eq: forward_SDE_DDPM_solution}
    \bY(\tau) &= e^{-\int_0^\tau f(v)\intd v}\bY(0) + \int_{0}^\tau e^{-\int_u^\tau f(v)\intd v}g(u) \intd \bB(u).
\end{align*}
Given a condition $\bc$, we let the marginal density of $\bY(\tau)$ be $q_\tau(\cdot | \bc)$, which is unknown due to the unknown $p_\tar(\cdot | \bc)$. For suitable choices of $T_g, a$ and $b$, the distribution $q_{T_g}(\cdot | \bc)$ is close to Gaussian by the convergence property of the OU process. When $\bY(0) = \by(0)$ is given, the conditional distribution of $\bY(\tau)$ is clearly Gaussian and known, which we denote by $q_{\tau|0}(\cdot |\by(0),  \bc)$.

Consider the reverse (in time) denoising process $\{\tilde\bY(\tau):= \bY(T_g - \tau), \tau\in[0, T_g]\}$.  Under mild assumptions \citep{Anderson1982, cattiaux2023time, haussmann1986time}, $\tilde\bY$ satisfies the reverse SDE starting from $\tilde\bY(0)\sim q_{T_g}(\cdot | \bc)$:
\begin{align}\label{eq: reverse_SDE_DDPM}
    \intd\tilde{\bY}(\tau) = [f(T_g - \tau)\tilde\bY(\tau) + g^2(T_g - \tau) \nabla\log q_{T_g - \tau}(\tilde \bY(\tau) | \bc)]\intd\tau + g(T_g - \tau) \intd\tilde\bB(\tau),
\end{align}
where $\nabla\log q_{\tau}(\cdot | \bc)$ is called the (conditional) score function and is unknown. Note that $\tilde\bY(T_g) = \bY(0)\sim p_\tar(\cdot | \bc)$, which is the target conditional distribution we want to sample from. Hence, the conditional diffusion model generates new samples by simulating the reverse SDE \eqref{eq: reverse_SDE_DDPM}. This requires (a) replacing the unknown $q_{T_g}(\cdot | \bc)$ by its approximation, which is Gaussian and hence easy to sample from;  (b) learning the score function $\nabla\log q_{\tau}(\cdot | \bc)$; and (3) discretizing \eqref{eq: reverse_SDE_DDPM} for sample generations.

To learn the unknown conditional score, one can consider training a neural network $s_\theta$ parameterized by $\theta$ to minimize $\int_{0}^{T_g}\mE\|s_\theta(\tau, \bY(\tau), \bc) - \nabla\log q_\tau(\bY(\tau) | \bc)\|^2\intd \tau$. This optimization problem is intractable because the objective involves the unknown score. In this paper, we apply the following denoising score matching procedure \citep{songscore, vincent2011connection}:
\begin{align}\label{eq: denosing_score_matching}
    \min_{\theta}\mE_{\bc\sim\bC}\mE_{\tau\sim \text{Unif}[0, T_g]} \mE_{\bY(0)} \mE_{\bY(\tau)|(\bY(0), \bc)}\bigg[\lambda(\tau)\big\| s_\theta(\tau, \bY(\tau), \bc) - \nabla\log q_{\tau|0}(\bY(\tau) | \bY(0), \bc)\big\|^2\bigg],
\end{align}
where $\text{Unif}[0, T_g]$ is the uniform distribution on $[0, T_g]$, $\bY(0)\sim p_\tar(\cdot | \bc)$, $\bY(\tau) | (\bY(0), \bc) \sim q_{\tau|0}(\cdot | \bc)$ which is Gaussian, and $\lambda: [0, T_g]\to \mR_{+}$ is some positive weighting function.  The stochastic gradient
descent algorithm and its variants can be applied to solve the empirical version of \eqref{eq: denosing_score_matching} given samples $\{(\bc_i, \by_i(0)): \bc_i\sim\bC, \by_i(0)\sim q_0(\cdot | \bc_i), i=1,\cdots N\}$.\footnote{We do not employ other popular training techniques for conditional diffusion models, such as the classifier-free guidance \citep{ho2021classifier} commonly used for image generation tasks.
This is because the data distribution in our SDE path generation task is fundamentally different from images, and applying classifier-free guidance does not yield performance gains in our experiments.}

With the conditional score function trained and learned, the sampling process is to simulate \eqref{eq: reverse_SDE_DDPM}
in $K$ steps, using the Euler--Maruyama discretization scheme with a uniform time step size of $\Delta\tau =T_g / K$. Specifically, let $\tau_k = k\Delta \tau$ and $\tilde{\bZ}(0) \sim \mathcal{N}(0, I_d)$. For $k=0,1,\cdots,K-1$, set
\begin{align}\label{eq: reverse_SDE_DDPM_simulation}
    \tilde{\bZ}(\tau_{k + 1}) = \tilde{\bZ}(\tau_{k}) + [\tilde f(\tau_k)\tilde\bZ(\tau_k) + \tilde g^2(\tau_k) s_\theta(T_g - \tau_k, \tilde{\bZ}(\tau_{k})), \bc)]\Delta\tau + \tilde g(\tau_k)\sqrt{\Delta\tau} \cdot \zeta_k,
\end{align}
where $\tilde f(\tau) = f(T_g - \tau), \tilde g(\tau) = g(T_g - \tau)$, and $\zeta_k$'s are i.i.d. standard Gaussian random vectors. The terminal state $\tilde\bZ(\tau_K)$ is a generated, desired sample.

\subsection{Generation of SDE Paths via Conditional Diffusion Models}
We now apply the general conditional diffusion model to generate sample paths of an unknown SDE.
Due to the Markov property of the SDE \eqref{eq: target_SDE}, we construct the desired generative model in an autoregressive manner. Specifically, we generate samples of  the stochastic increment $\Delta\bX(t; \bx) :=\bX^{t, \bx}(t + \Delta t) - \bx$ defined by the solution to \eqref{eq: target_SDE} conditional on $\bX^{t, \bx}(t) = \bx$. For each $t_n$, we set condition $\bc_n = (t_n, \bx(t_n))$ sampled from $\bC_n = (t_n, \bX(t_n))$ and train a conditional diffusion model $G(t_n, \bx; \theta_n)$ where $\bx(t_n)=\bx$, such that $G(t_n, \bx; \theta_n)$ consists of a score network $s_{\theta_n}$ trained via \eqref{eq: denosing_score_matching} and generates new samples $\tilde\bZ_n(\tau_K)$ distributed as $\Delta\bX(t_n; \bx)$ by running \eqref{eq: reverse_SDE_DDPM_simulation}, i.e.,
\begin{align}\label{eq: objective_AR_gen}
	\tilde\bZ_n(\tau_K)\overset{d}{\approx}  \Delta \bX(t_n; \bx) := \bX(t_n + \Delta t) - \bx =  \int_{t_n}^{t_n + \Delta t}\mu(t, \bX(t))\intd t + \int_{t_n}^{t_n + \Delta t}\sigma(t, \bX(t))\intd\bW(t).
\end{align}
The entire training process is stipulated in Algorithm \ref{alg: training_process}, where $N_T$ score networks $\{s_{\theta_n}: n=0, \cdots, N_T - 1\}$ are trained in parallel.
Next, to generate new sample paths (time series) from the unknown SDE, we start from $\hat \bx(t_0) = \bx(0)$ and recursively produce
\begin{align}
	\hat\bX(t_{n+1}) = \hat\bx(t_{n }) + \tilde\bZ_n(\tau_K),\quad  \text{for}\ n = 0, ..., N_T-1. \label{eq: AR_generation}
\end{align}
The detailed generation process of SDE paths is summarized in Algorithm \ref{alg: generation_process}.

\begin{remark}
Instead of using conditional diffusion models to generate SDE paths in a sequential and autoregressive fashion, an alternative approach is to train a single diffusion model to generate samples of the entire trajectory $(\bX(t_1), \cdots, \bX(t_{N_T}))$ directly.
In such a case, the dimension of the target distribution is $d\times N_T$, which becomes large with a high-dimensional SDE (large $d$) and/or a large number of observation slots (large $N_T$). While diffusion models can indeed deal with high-dimensional distributions, they nonetheless require more complex network architectures to train the score functions and longer training times. The autoregressive approach we choose instead reduces the dimension and complexity of the generation task, which allows us to use a simple neural network such as a multilayer perceptron (MLP) for training the conditional score function, leading to a more efficient solution.
\end{remark}

% \newpage

\begin{algorithm}[H]
\small
\caption{Training process of the SDE generative model
$\{G(t_n, \cdot; \theta_n): n=0, \cdots, N_T - 1\}$}
\begin{algorithmic}\label{alg: training_process}
    \REQUIRE Training data $\{(\bx^{(i)}(t_n), \Delta\bx^{(i)}(t_n; \bx^{(i)}(t_n))): i=1,\cdots, H, n=0, \cdots, N_T - 1\}$, score networks $\{s_{\theta_n}\}_{n=0}^{N_T - 1}$, batch size $m\le H$, learning rate function $\alpha(\cdot)$, noise schedulers $f(\tau)$ and $g(\tau)$, diffusion horizon $T_g$, number of diffusion steps $K$, diffusion time step $\Delta\tau = T_g/K$
    \FOR{episode $j = 1,..., J$}
    \FOR{SDE slot $t_n= 0, \Delta t, 2\Delta t, ..., (N_T - 1)\Delta t$}
    \STATE 1. Randomly sample a batch of data at $t_n$:
    \vspace{-2mm}
        $$\big\{(\bc^{(i_l)}, \by^{(i_l)}(0)) = ((t_n, \bx^{(i_l)}(t_n)), \Delta\bx^{(i_l)}(t_n; \bx^{(i_l)}(t_n))): l=1,\cdots, m\big\}$$
    \vspace{-7mm}
    \STATE 2. Randomly sample a diffusion step $k\sim \text{Unif}[1, K]$ and set $\tau_k = k\Delta\tau$
    \STATE 3. Generate noised samples $\{\by^{(i_l)}(\tau_k): l=1,\cdots, m\}$, where
    \vspace{-3mm}
    \[\by^{(i_l)}(\tau_k) = e^{-\int_0^{\tau_k} f(v)\intd v}\by^{i_l}(0) + \bigg(\int_{0}^{\tau_k} e^{-\int_u^{\tau_k} 2f(v)\intd v}g^2(u) \intd u\bigg)^{1/2}\zeta_l,\ \zeta_l\sim\mathcal{N}(0, I_d)
    \]
    \vspace{-5mm}
    \STATE 4. Compute the denosing score matching loss $L(\theta_n)$\footnotemark:
    \vspace{-2mm}
    \begin{align*}
    \frac{1}{m}\sum_{l=1}^m \int_{0}^{\tau_k} e^{-\int_u^{\tau_k} 2f(v)\intd v}g^2(u) \intd u\cdot\big\| s_{\theta_n}(\tau_k, \by^{(i_l)}(\tau_k), \bc^{(i_l)}) - \nabla\log q_{\tau_k|0}(\by^{(i_l)}(\tau_k) | \by^{(i_l)}(0), \bc^{(i_l)})\big\|^2
    \end{align*}
    \vspace{-4mm}
    \STATE 5. Update $\theta_n \leftarrow \theta_n - \alpha(j)\nabla_{\theta_n} L(\theta_n)$
    \ENDFOR
    \ENDFOR
    \RETURN score network $s_{\theta_n}$ of $G(t_n, \cdot; \theta_n)$ for $n=0, \cdots, N_T-1$
\end{algorithmic}
\end{algorithm}
\footnotetext{This loss function corresponds to the training objective of Denoising Diffusion Probabilistic Models in \cite[Section 3.4]{ho2020denoising}. }

% \vspace{-10mm}
\begin{algorithm}[htp]
\small
\caption{Generation of synthetic SDE paths}
\begin{algorithmic}\label{alg: generation_process}
    \REQUIRE network $\{s_{\theta_n}\}_{n=0}^{N_T - 1}$, initial position $\hat\bx(t_0)=\bx(0)$, noise schedulers $f(\tau)$ and $g(\tau)$, diffusion horizon $T_g$, number of diffusion steps $K$, diffusion time step $\Delta\tau = T_g/K$
    \FOR{SDE slot $t_n= 0, \Delta t, 2\Delta t, ..., (N_T - 1)\Delta t$}
    \STATE 1. Set the condition $\bc_n = (t_n, \hat\bx(t_n))$
    \STATE 2. Simulate $\tilde\by(\tau_0)\sim\mathcal{N}(0, I_d)$
    \STATE 3. Simulate the reverse process of the conditional diffusion model:
    \FOR{diffusion step $\tau_k = 0, \Delta\tau, 2\Delta\tau, \cdots, K\Delta\tau$}
    \vspace{-5mm}
        \STATE
        \[\tilde{\by}(\tau_{k + 1}) = \tilde{\by}(\tau_{k}) + [ f(T_g -\tau_{k})\tilde\by(\tau_{k}) +  g^2(T_g -\tau_{k}) s_{\theta_n}(T_g - \tau_k, \tilde\by(\tau_{k}), \bc_n)]\Delta\tau + g(T_g - \tau_k) \sqrt{\Delta\tau}\zeta_k
        \]
    % \vspace{-2mm}
        where $\zeta_k\sim\mathcal{N}(0, I_d)$
    \ENDFOR
    \STATE 4. Set and store the next state $\hat\bx(t_{n + 1}) = \hat\bx(t_n) + \tilde\by(\tau_{K})$
    \ENDFOR

    \RETURN generated path $(\hat\bx(t_0), \hat\bx(t_1), \cdots, \hat\bx(t_{N_T}))$
\end{algorithmic}
\end{algorithm}
% \vspace{10mm}

\section{Numerical Study For Path Generation}\label{sec: numerical_experiment}

In this section, we numerically evaluate our generative model using training sample paths that are simulated from the underlying SDEs. We test our method on four classical examples, three of which are one-dimensional SDEs including a time-homogeneous OU process, a time-inhomogeneous geometric Brownian motion (GBM) and a time-homogeneous Cox–Ingersoll–Ross (CIR) process, and the last is a 100-dimensional time-homogeneous GBM. We choose these SDEs because, for the purpose of generating training data, the solutions to these SDEs can be simulated {\it exactly} in the sense that the joint distribution of simulated values coincides with that of the continuous-time SDE on the simulation time grid \citep[Chapter 3]{glasserman2004monte}.

We benchmark our approach against two alternative methods. The first one (termed as SDM-MC), proposed in \cite{liu2025training}, is also an autoregressive generative model built on diffusion models but with a Monte Carlo score function estimator (so that score training is not required). The second method is based on Neural SDE proposed in \cite{kidger2021neural}. We compare the three approaches with metrics including the KL-divergences between the distribution of the training path and that of the generated synthetic SDE path.

For this numerical study, we display the network structure and other configurations of our diffusion models  $\{G(t_n, \cdot; \theta_n)\}_{n=0}^{N_T - 1}$ in Appendix \ref{sec: experiment_settings}.
For the two benchmarks, we directly follow their hyperparameter settings in \cite{liu2025training} and \cite{kidger2021neural}, respectively.

\paragraph{Example 1.}
A (one-dimensional) time-homogeneous OU process is governed by
\begin{align}\label{eq: SDE_OU}
    \intd X(t) = \theta (\mu - X(t)) \intd t + \sigma \intd W(t),\ X(0) = x(0).
\end{align}
Set $x(0) = 1.5, \mu=1.2, \theta = 1, \sigma=0.3$, and $T = 1, \Delta t = 0.05$, resulting in $N_T = 20$ observation slots in total. We collect $H = 100$ ``real'' OU trajectories
in the training dataset \eqref{eq: obs_data}, each simulated from \eqref{eq: SDE_OU} via Monte Carlo. Then, in each experiment, we generate 100 synthetic OU paths with each of the three methods, and compute the corresponding KL divergences between the joint distributions of the (discrete observations of) 100 real and synthetic paths using the method from \cite{wang2009divergence}. We repeat 5,000 such experiments.

\begin{table}[htp]
    \centering
    \begin{tabular}{c | c c c}
    \toprule
     Method & Ours & SDM-MC & Neural SDE\\
     \midrule
     KL divergence $\downarrow$ & \tbf{0.0676 $\pm$ 0.004} & 0.2652 $\pm$ 0.008 & 0.9243 $\pm$ 0.014\\
    \bottomrule
    \end{tabular}
    \caption{The 95\% confidence intervals of KL-divergences between 100 real and synthetic OU paths on time grid $\mathcal{T}$.}
    \label{table: OU_KL}
\end{table}

Table \ref{table: OU_KL} presents the 95\% confidence intervals of the resulting KL-divergence values for the three methods. The results show that our model produces paths with distributions much closer to those of the target OU process, significantly outperforming both benchmarks.

\begin{figure}[htp]
    \centering
    \includegraphics[scale=0.44]{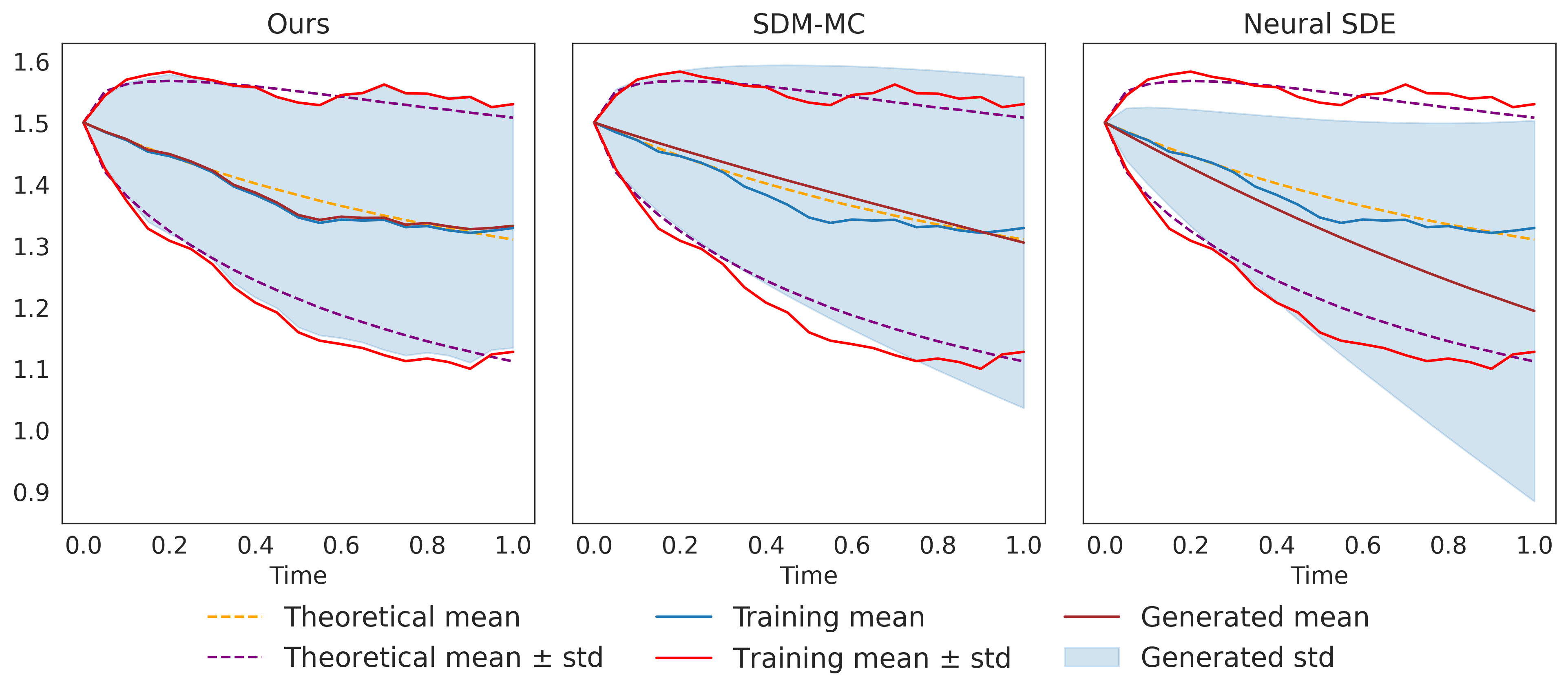}
    \caption{One-dimensional OU process: comparison of mean and standard deviation (std) of solutions obtained by three generative models with 100 synthetic paths. Theoretical and training mean and std are also plotted.}
    \label{fig: 1D_OU_trend}
\end{figure}

Next, following \cite{liu2025training}, we evaluate the performance of the three generative models by comparing the mean and standard deviation (std) of the solutions obtained using 100 synthetic paths. The results are presented in Figure \ref{fig: 1D_OU_trend}, which also includes the theoretical (i.e., oracle) mean and std of the OU process $X(t)$ from \eqref{eq: SDE_OU} for reference. As shown, both SDM-MC and our method effectively capture the evolution trend of the training paths, although the latter seems to fit better both the theoretical/training mean and std. By contrast, the Neural SDE method performs poorly.

Finally, a maximum likelihood estimation (MLE) method can be applied to estimate the parameters $\mu, \theta$ and $\sigma$ of the target OU process \eqref{eq: SDE_OU} based on a given set of solution paths, as described in \cite{tang2009parameter}. We apply it to compare the performance (in terms of the estimation accuracy) of the three methods based on 1,000 repeated experiments each on 100 synthetic paths from the corresponding generative models. We also include the parameter estimations based on the given 100 real OU paths for training as reference points, and report the comparison results in Table \ref{table: 1D_OU_params}.

There are several notable observations from Table \ref{table: 1D_OU_params}. First, none of the four can estimate the mean reversion speed parameter $\theta$ accurately. This nevertheless is consistent with the common knowledge that  the mean reversion speed of OU processes is inherently hard to estimate with a small sample size. Moreover, none of the three generative methods performs meaningfully better  than the one using the 100 OU training trajectories for estimating any parameters. The reason is that generated paths do not give any more probabilistic/statistical information than those already contained in the original sample/training paths, not to mention that the former inherently contain bias and deviations from the latter.
This, in turn, has a significant implication: {\it generative methods do not really generate new information.} However, for decision making, generative synthetic data provides more scenarios to train and improve action {\it strategies}. More on this in Section \ref{sec:application}, where we present a concrete application of our method. The other observation is that our method consistently yields more accurate estimates than those from the two benchmark models, with large margins in some cases. For the drift parameters $\theta$ and $\mu$, even though estimation accuracy is not good in general, our method still significantly outperforms the other two and is the closest to the ones obtained by the training paths.

\begin{table}[htp]
    \centering
    \begin{tabular}{c c | c | c c c}
    \toprule
    Parameter & Truth & Training & Ours & SDM-MC & Neural SDE\\
    \hline
     $\mu$ & 1.2 & 1.1830  & 1.1827 $\pm$ 0.0031 & 0.7669 $\pm$ 0.5506 & 0.6888 $\pm$ 0.5810\\
     % \hline
     $\theta$ & 1 & 0.8699 & 0.8708 $\pm$ 0.0105 & 0.2120 $\pm$ 0.0100 & 0.4151 $\pm$ 0.0177\\
     % \hline
     $\sigma$ & 0.3  & 0.2987 & 0.2900 $\pm$ 0.0003 & 0.2934 $\pm$ 0.0003 & 0.3492 $\pm$ 0.0008\\
    \bottomrule
    \end{tabular}
    \caption{Parameters and 95\% confidence intervals estimated from 100 training OU paths and 100 synthetic paths obtained from three generative methods.}
    \label{table: 1D_OU_params}
\end{table}

\paragraph{Example 2.}
A (one-dimensional) time-inhomogeneous GBM is driven by the following SDE:
\begin{align}\label{eq: SDE_TGBM}
    dX(t) = \mu(t)X(t) dt + \sigma(t)X(t) dW(t),\ X(0) = x(0),
\end{align}
where we choose  $\mu(t) = 4t, \sigma(t) = \sqrt{4t}$ and $x(0) = 1$. In this example, we set $T=\frac{1}{2}$ and consider a non-uniform time grid $\mathcal{T}$ to demonstrate the flexibility and generalizability of our approach. Specifically, we generate $N_T=100$ observation time points \textit{at random} on $[0,T]$, and simulate $H=100$ trajectories of  \eqref{eq: SDE_TGBM} on $\mathcal{T}$ to construct the training dataset.

\begin{table}[htp]
    \centering
    \begin{tabular}{c | c c c}
    \toprule
     Method & Ours & SDM-MC & Neural SDE\\
     \midrule
     KL divergence $\downarrow$ & \tbf{1.1075 $\pm$ 0.047} & N.A. & 6.4322 $\pm$ 0.0773\\
    \bottomrule
    \end{tabular}
    \caption{The 95\% confidence intervals of KL-divergences between 100 real and synthetic time-inhomogeneous GBM paths on time grid $\mathcal{T}$.}
    \label{table: TGBM_KL}
\end{table}

\begin{figure}[htp]
    \centering
    \includegraphics[scale=0.44]{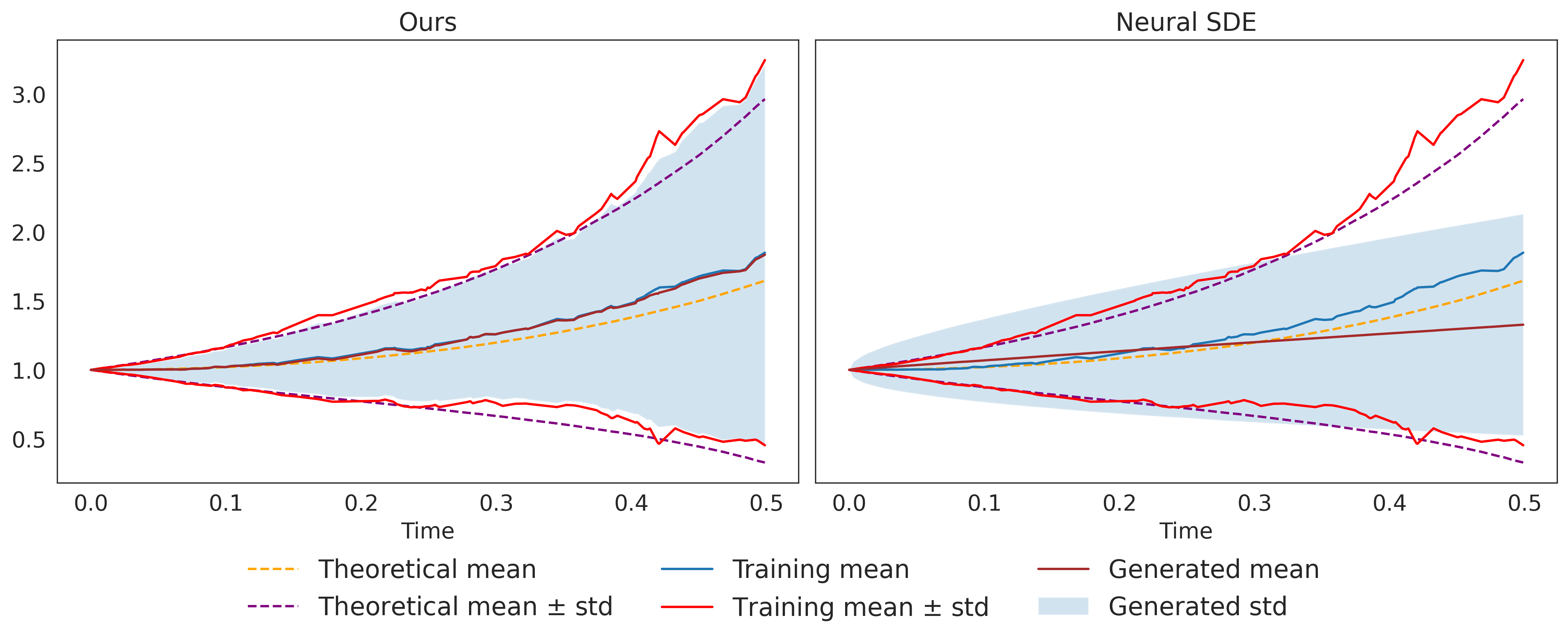}
    \caption{One-dimensional time-inhomogeneous GBM: comparison of mean and standard deviation (std) of solutions obtained by two generative models with 100 synthetic paths. Theoretical and training mean and std are also plotted.}
    \label{fig: 1D_TGBM_trend}
\end{figure}

Since the SDM-MC method is designed for time-homogeneous SDEs, we compare our approach with Neural SDE only in this example. As shown in Table \ref{table: TGBM_KL}, the KL divergence between the joint distribution of the real GBM paths and that of our synthetically generated paths, is significantly smaller than the Neural SDE counterpart. 
Furthermore, Figure \ref{fig: 1D_TGBM_trend} illustrates the various means and standard deviations of the solutions generated by the two approaches, revealing that our method accurately captures the evolution trend of the training paths, whereas Neural SDE fails to do so. These results further demonstrate the superiority of our approach over Neural SDE.

\vspace{-10mm}
\paragraph{Example 3.}
A (one-dimensional) time-homogeneous Cox–Ingersoll–Ross (CIR) process is
\begin{align*}%\label{eq: SDE_CIR}
    dX(t) = \alpha(b - X(t))dt + \sigma\sqrt{X(t)}dW(t),\ X(0) = x(0).
\end{align*}
We take $x(0) = 0.5, \alpha = 0.2, b = 0.05, \sigma=0.1$ and the training data are $H=100$ real CIR paths with $T=0.5$ and $\Delta t= 0.01$.

Table \ref{table: CIR_KL} presents a comparison of the KL divergences between the real CIR paths and the synthetic ones generated by the three different models (based on 5000 repeated experiments). Again, the Neural SDE method exhibits considerably poorer performance, whereas our model beats SDM-MC.

\begin{table}[htp]
    \centering
    \begin{tabular}{c | c c c}
    \toprule
     Method & Ours & SDM-MC & Neural SDE\\
     \midrule
     KL divergence $\downarrow$ & \tbf{0.0917 $\pm$ 0.008} & 0.4541 $\pm$ 0.019 & 0.7047 $\pm$ 0.023\\
    \bottomrule
    \end{tabular}
    \caption{The 95\% confidence intervals of KL-divergences between  100 real and synthetic CIR paths on time grid $\mathcal{T}$.}
    \label{table: CIR_KL}
\end{table}

\vspace{-5mm}
\paragraph{Example 4.}
A multivariate GBM in $\mR^{d}$ with $d=100$ is defined as:
\begin{align*}%\label{eq: SDE_hd_GBM}
	\intd\bX_j(t) = \mu_j\bX_j(t)\intd t + \sigma_j\bX_j(t) \intd\bW_j(t),\ \bX_j(0) = x_j(0),\quad j=1,\cdots, 100,
\end{align*}
where $\bW_j$ is a one-dimensional standard Brownian motion, and $\bW_j(t)$ and $\bW_l(t)$ have a correlation $\rho_{jl}\in[-1, 1]$ for $j, l=1,..., 100$. We randomly generate $\boldsymbol{\mu}, \boldsymbol{\sigma}, \boldsymbol{\rho}$, the corresponding vector or matrices consisting of those model parameters.
Set $T=7$ and $\Delta t=1$ so that the time step is ``large''. We use $H=10,000$ real paths to train the three generative models.

\begin{table}[htp]
    \centering
    \begin{tabular}{c | c c c}
    \toprule
     Method & Ours & SDM-MC & Neural SDE\\
     \midrule
     KL divergence $\downarrow$ & \tbf{3.84 $\pm$ 0.106} & 247.41 $\pm$ 0.347 & 822.38 $\pm$ 0.316\\
    \bottomrule
    \end{tabular}
    \caption{The 95\% confidence intervals of KL-divergences between 10,000 real and synthetic 100-dimensional time-homogeneous  GBM paths on time grid $\mathcal{T}$.}
    \label{table: HD_GBM_KL}
\end{table}

\begin{figure}[htp]
	\centering
	\includegraphics[scale=0.35]{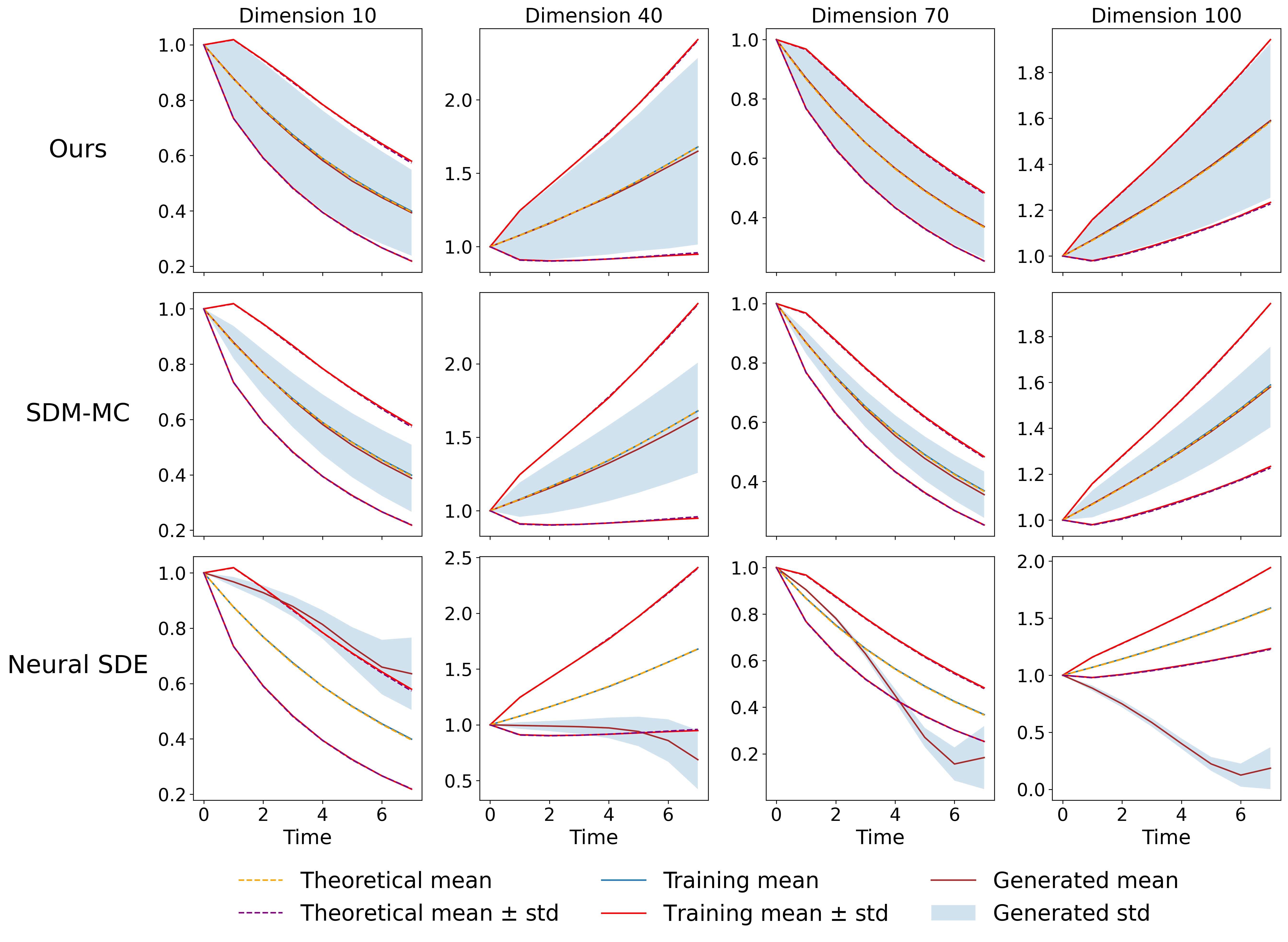}
	\caption{100-dimensional time-homogeneous GBM: comparison of the mean and the standard deviation (std) of solutions (across 4 different dimensions) obtained by three generative models with 10,000 synthetic paths. Theoretical and training mean and std are also plotted.}
	\label{fig: HD_GBM_trend_DDPM}
\end{figure}

Table \ref{table: HD_GBM_KL} reports the KL divergence results based on 100 repeated experiments. This time, our method outperforms the other two {\it massively}.
Figure \ref{fig: HD_GBM_trend_DDPM} further provides a visual confirmation that our synthetic paths closely track the training trajectories in terms of mean and standard deviation at multiple dimensions. By contrast, the other two methods fail to achieve this level of accuracy.

In summary, our approach consistently outperforms the two benchmarks across nearly all experiments in generating synthetic SDE paths. The SDM-MC method requires an extensive amount of data to construct an accurate Monte-Carlo score estimator for a high-quality generative model. By contrast, our model excels even with limited training trajectories (relative to the dimension) and is applicable to time-inhomogeneous SDEs, offering greater flexibility. Moreover, our model scales effectively to high-dimensional generation tasks, whereas SDM-MC struggles to do so. Meanwhile,  the Neural SDE method is often plagued by unstable training dynamics and exhibits high sensitivity to hyperparameter choices, a phenomenon also observed in other studies \citep{issa2023non, liao2024sig}. This inherent instability ultimately contributes to its poor performance in our experiments.

\section{Error Analysis}\label{sec:error}
In this section, we provide an error analysis of our SDE generative model. Specifically,  consider the time series corresponding to the target unknown SDE \eqref{eq: target_SDE} on grid $\mathcal{T}$: $(\bX(t_0), \cdots, \bX(t_{N_T}))$.
Given a fixed $\bX(t_0) \equiv \bx(0)$, we denote by $P_{1:N_T|0}$ the joint distribution of the target path $(\bX(t_1), \cdots, \bX(t_{N_T}))$, and by $\genP_{1:N_T|0}$ that of the generated path $(\genX(t_1), \cdots, \genX(t_{N_T}))$.
Our goal is to derive an explicit bound in the KL-divergence between $P_{1:N_T|0}$ and $\genP_{1:N_T|0}$, so as to quantify
the aggregated error of our generative model. In the following, we discuss the well-posedness and decomposition of the KL error in Section~\ref{sec:decompKL}, the error bound for conditional diffusion models in Section~\ref{sec:error-cdm}, and the overall error bound for the SDE generative model in Section~\ref{sec:bound-SDE}.

\subsection{Decomposition of the KL Error}\label{sec:decompKL}

The very existence of the KL-divergence $\KL\big(P_{1:N_T|0} ||\genP_{1:N_T|0}\big)$ requires $P_{1:N_T|0}$ to be absolutely continuous w.r.t. $\genP_{1:N_T|0}$. To guarantee such a condition is met and thereby $\KL\big(P_{1:N_T|0} ||\genP_{1:N_T|0}\big)$ is well-defined, we make the following assumption on the target SDE.

\begin{assumption}\label{assumption: SDE_solution_supp_abs}
    (a) The coefficients $\mu$ and $\sigma$ of the target SDE \eqref{eq: target_SDE} satisfy the global Lipschitz and linear growth conditions:
    \begin{align*}
        ||\mu(t, \bx) - \mu(t, \bx')||_2 + ||\sigma(t, \bx) - \sigma(t, \bx')||_2 &\le C_L||\bx - \bx'||_2,\\
        ||\mu(t, \bx)||_2 + ||\sigma(t, \bx)||_2 &\le C_L(1 + ||\bx||_2),
    \end{align*}
    for every $t\in[0, T], \bx\in\mR^d, \bx'\in\mR^d$, where $C_L$ is a positive constant independent of $(t, \bx,\bx')$.

   (b) For any $0 \le s \le t \le T$,  the conditional distribution of $\bX(t)$ given $\bX(s) = \bx(s)$ is absolutely continuous w.r.t. the Lebesgue measure, and it admits a probability density function that is positive almost everywhere on $\mR^d$.
\end{assumption}

Assumption \ref{assumption: SDE_solution_supp_abs}(a) is the classical (sufficient) condition for a general SDE to have a unique strong solution \citep{karatzas2014brownian}. Moreover, under this assumption the solution to \eqref{eq: target_SDE} admits the following moment bounds \citep[Theorem 4.5.4]{kloedennumerical}: there exists a positive constant $C_M$ depending only on $C_L$ such that $$\mE\|\bX(t)\|^2 \le (1 + \mE\|\bX(0)\|^2)e^{tC_M}$$ for any $t\in[0, T],$ and
\begin{align} \label{eq: target_SDE_moment_bound}
    \mE\|\bX(t_{n + 1}) - \bX(t_n)\|^2 &\le M_2(t_n; T) \cdot e^{\Delta tC_M}\Delta t
\end{align}
for any $t_n\in\mathcal{T}$, where
\begin{align*}
   M_2(t_n; T)&:= D_1 \cdot (T - t_n + 1)(1 + \mE\|\bX(t_n)\|^2) \nonumber \\
   & \le D_2\cdot (T - t_n + 1)(1 + \mE\|\bX(0)\|^2)e^{t_{n}C_M },
\end{align*}
and $D_1$ and $D_2$ are some positive constants depending only on $C_L$. These moment bounds will be useful for the subsequent error analysis.

Assumption \ref{assumption: SDE_solution_supp_abs}(b) is imposed to ensure that the joint distributions $P_{1:N_T|0}$ and $\genP_{1:N_T|0}$ share the same support and the absolute continuity condition holds. Drifted Brownian motions, OU processes, and Langevin SDEs are classical examples satisfying Assumption \ref{assumption: SDE_solution_supp_abs}(b).
General sufficient conditions for Assumption \ref{assumption: SDE_solution_supp_abs}(b) to hold have also been extensively studied in Malliavian calculus and support theorem of SDE; see, e.g.  \cite{gyongy1990approximation, Michel1990, nualart2006malliavin, stroock1972support}.  Some SDEs, such as GBMs, do not satisfy Assumption \ref{assumption: SDE_solution_supp_abs}(b) because the marginal distribution is not supported on the entire space $\mR^d$. For such SDEs, one may conduct some transformations (e.g., the logarithmic one) on their states so that the transformed SDEs have transition densities supported on $\mR^d$.

We now introduce a few more notations to facilitate exposition. For any $t_n\in\mathcal{T}$, denote by $P_{\Delta\bX(t_n; \bx(t_n))}$ the conditional distribution of the increment $\Delta\bX(t_n; \bx(t_n))$ of the target SDE given  $\bX(t_n) = \bx(t_n)\in\mR^d$. Let $\Delta\genX(t_n; \bx(t_n))$ be the corresponding generated increment (i.e. $\tilde\bZ_n(\tau_K)$ in \eqref{eq: objective_AR_gen}), with $\genP_{\Delta\genX(t_n; \bX(t_n))}$ its distribution.
 Under Assumption \ref{assumption: SDE_solution_supp_abs}, we have the following useful result, the proof of which is deferred to Appendix \ref{sec: path_KL_proof}.

\begin{proposition}\label{prop: path_KL_decomposition}
     Suppose that Assumption \ref{assumption: SDE_solution_supp_abs} holds. Then,
     \begin{enumerate}
         \item both $P_{\Delta\bX(t_n; \bx(t_n))}$ and $\genP_{\Delta\genX(t_n; \bX(t_n))}$ are equivalent to the Lebesgue measure and admit density functions that are positive almost everywhere on $\mR^d$.

        \item  Both $P_{1:N_T|0}$ and $\genP_{1:N_T|0}$ are equivalent to the Lebesgue measure and admit density functions that are positive almost everywhere on $\mR^{d\times N_T}$. Moreover, we have
        \begin{align}\label{eq: path_KL_decomposition}
            \KL\big(P_{1:N_T|0} ||\genP_{1:N_T|0}\big) = \sum_{n=0}^{N_T - 1}\mE_{\bX(t_n)\sim P_{n|0}}\bigg[\KL\big(P_{\Delta\bX(t_n; \bX(t_n))} \big|\big| \genP_{\Delta\genX(t_n; \bX(t_n))}\big)\bigg],
        \end{align}
        where $P_{n|0}$ is the marginal distribution of the SDE solution $\bX(t_n)$ conditional on $\bX(t_0) \equiv \bx(t_0)$. 
     \end{enumerate}
\end{proposition}

The decomposition in \eqref{eq: path_KL_decomposition} arises from the chain rule for KL divergence \citep[Exercise 3.2]{wainwright2019high}  together with the Markovian nature of the target and generated paths. This fundamental property of chain rule is the primary technical motivation for us to use the KL divergence for the error analysis. As a direct consequence of \eqref{eq: path_KL_decomposition}, we can simplify the task of bounding the KL-divergence between the target and generated paths. Specifically, it suffices to derive an error bound for {\it each} conditional diffusion model of producing the increment $\Delta \genX(t_n; \bX(t_n))$. We will delve into this crucial step in the next section.

\subsection{Error Bound for Conditional Diffusion Models}\label{sec:error-cdm}
In this section, we establish a KL error bound for the general conditional diffusion model discussed in Section~\ref{sec:CDM-review}, providing a theoretical guarantee on its performance. The result actually goes beyond the specific application of SDE path generation.
%While its proof is similar in spirit to the general error results for the unconditional diffusion model \citep{bentonnearly}, there are certain subtle yet important differences, which is the reason why we present it for the reader's convenience.

 Recall that $p_\tar(\cdot | \bc)$ denotes the unknown target conditional distribution. We denote by $\tilde p^\theta_{\tau_K}(\cdot | \bc)$ the marginal distribution of the generated sample $\tilde\bZ(\tau_K)$ in \eqref{eq: reverse_SDE_DDPM_simulation} given condition $\bc$. The following result provides the desired KL error bound for the conditional diffusion model. Its proof is lengthy, which is delayed to Appendix \ref{sec: convergence_proof}.

\begin{proposition}\label{prop: condition_diffusion_KL_bound}
Suppose the following conditions hold:
\begin{enumerate}
    \item for any given condition $\bc$, the target conditional distribution admits a twice continuously differentiable density function $p_\tar(\cdot | \bc)$, which is positive almost everywhere on $\mR^{d_\tar}$ and the corresponding score function $\nabla\log p_\tar(\cdot | \bc)$ is $L(\bc)$-Lipschitz continuous with
    \begin{align}\label{eq: convergence_lipschitz}
        \mE_{\bc\sim\bC}[L(\bc)] = L_1 < \infty.
    \end{align}

    \item The target distribution has a finite second moment, i.e.
    \begin{align}\label{eq: convergence_finite_moment}
        \mE_{\bc\sim\bC}\big[\mE_{\bY(0)\sim p_\tar(\cdot | \bc)}||\bY(0)||^2\big] < M_2
    \end{align}
    for some positive constant $M_2$.

    \item There exists a small constant $\varepsilon_{\text{score}}>0$ such that
    \begin{align}\label{eq: convergence_score_matching}
        \frac{1}{K}\sum_{k=1}^K\mE_{\bc\sim\bC}\bigg[\mE_{\bY(\tau_k)\sim q_{\tau_k}(\cdot | \bc)}\big[||\nabla\log q_{\tau_k}(\bY(\tau_k) | \bc) - s_\theta(\tau_k, \bY(\tau_k), \bc)||^2\big]\bigg] \le \varepsilon_{\text{score}}^2.
    \end{align}
\end{enumerate}

Then, we have
\begin{align}
    &\mE_{\bc\sim\bC}\left[\KL\big(p_\tar(\cdot | \bc) || \tilde p^\theta_{\tau_K}(\cdot | \bc)\big)\right]\nonumber\\
    &\lesssim \underbrace{e^{-\frac{a}{2}T_g^2 - bT_g}\big[M_2 + de^{-\frac{a}{2}T_g^2 - bT_g}\big]}_{{\text(i)}} + \underbrace{(T_g + 1)^3 \varepsilon_{\text{score}}^2}_{\text{(ii)}} + \underbrace{\Delta\tau(M_2 + d)(T_g + 1)^5}_{\text{(iii)}} \nonumber\\
    &\qquad + \underbrace{M_2\Delta\tau (T_g + 1)^2 + d\Delta\tau \big[L_1 + (T_g + 1)^4 \big]}_{\text{(iv)}} + \underbrace{d(\Delta\tau)^2(L_1 + T_g + 1)}_{\text{(v)}}.\label{eq: condition_diffusion_KL_bound} 
\end{align}
\end{proposition}

In the following, we provide interpretations of the error bound components in \eqref{eq: condition_diffusion_KL_bound}, and discuss the conditions imposed in Proposition \ref{prop: condition_diffusion_KL_bound}:
\begin{enumerate}
    \item The term (i) of \eqref{eq: condition_diffusion_KL_bound} is the \textit{initialization error}, caused by sampling the reverse SDE \eqref{eq: reverse_SDE_DDPM} from a Gaussian noise rather than the terminal state of the forward SDE \eqref{eq: forward_SDE_DDPM}. This error will be small if $e^{-\frac{a}{2}T_g^2 - bT_g}$ is small. For image generation tasks, one typical choice is to set $T_g=1$, $a=19.9$ and $b=0.1$ \cite[Appendix C]{songscore}, resulting in a very small value of $e^{-\frac{a}{2}T_g^2 - bT_g}$. This is also the setting we take in this paper, and it is different from the one in the prior theoretical studies \citep{bentonnearly, chen2023improved} where $a=0$ and $b=1$, in which case one needs to choose a large $T_g$ to make the initialization error small.

    \item The term (ii) of \eqref{eq: condition_diffusion_KL_bound} is the \textit{score matching error}, which measures the quality of the neural network $s_\theta$ for approximating the unknown score function $\nabla\log q_\tau$ at each $\tau=\tau_k$. This term will be small, if $\varepsilon_{\text{score}}$ in
    \eqref{eq: convergence_score_matching} is small. Meanwhile, the condition \eqref{eq: convergence_score_matching} quantifies the score matching error averaged over $K$ steps of the generation process. This condition is standard and used in many studies on convergence analysis of diffusion models; see e.g. \cite{chen2023sampling, bentonnearly, chen2023improved, gao2025wasserstein}.

    \item The terms (iii), (iv) and (v) represent the \textit{discretization error} of simulating the reverse SDE \eqref{eq: reverse_SDE_DDPM} by the Euler--Maruyama scheme in \eqref{eq: reverse_SDE_DDPM_simulation}. In particular, for $\tau\in[\tau_k, \tau_{k + 1})$,
    \begin{enumerate}
        \item[3(a)] the term (iii) corresponds to the state discretization error induced by approximating $f(\tau)\bY(\tau)$ with $f(\tau_k)\bY(\tau_k)$,
        \item[3(b)] the term (iv) is the score function discretization error caused by using $g(\tau_k)\nabla\log q_{\tau_k}(\bY(\tau_k)|\bc)$ to replace $g(\tau)\nabla\log q_{\tau}(\bY(\tau)|\bc)$, and
        \item[3(c)] the term (v) captures the approximation error of simulating the diffusion term $g(T_g - \tau)\intd\bB(\tau)$ by $g(T_g - \tau_k)\sqrt{\Delta\tau}\zeta_k$. 
    \end{enumerate}
    Clearly,  the discretization error can be made small if $\Delta \tau$ is small, i.e., if the number of sampling steps $K$ in the generation process is large, so that the Euler--Maruyama scheme approximates the reverse SDE \eqref{eq: reverse_SDE_DDPM} well. Because the drift of \eqref{eq: reverse_SDE_DDPM} involves the score function, we impose the first (Lipschitz score) condition
in Proposition \ref{prop: condition_diffusion_KL_bound} to control the discretization error.
\end{enumerate}

Our proof of Proposition \ref{prop: condition_diffusion_KL_bound} builds on two recent papers on KL error analysis of unconditional diffusion models \citep{bentonnearly, chen2023improved}, which treat a simplified case with $f=\frac{1}{2}$ and $g=1$ in the forward process \eqref{eq: forward_SDE_DDPM}. By contrast, we examine conditional diffusion models where $f$ and $g$ are time-dependent functions. This introduces several technical complexities in our theoretical analysis, as the error characteristics of the discretized sample generation process differ substantially from those in the constant case (i.e., $f=\frac{1}{2}$ and $g=1$). In particular, the Girsanov-based method used in \cite{bentonnearly, chen2023improved} to bound the KL distance between the path measures of the true and approximate reverse process is not applicable to our analysis. This is because the true reverse process $\tilde \bY$ in \eqref{eq: reverse_SDE_DDPM} and its approximation $\tilde\bZ$ in \eqref{eq: reverse_SDE_DDPM_simulation} do not share the same diffusion coefficient ($\tilde g(\tau)$ vs $\tilde g(\tau_k)$), precisely due to the presence of a non-constant function $g$ in our setting. 
It is also worth noting that  \cite{bentonnearly} and \cite{chen2023improved} do not require the Lipschitz score condition owing to an early-stopping technique: the reverse process \eqref{eq: reverse_SDE_DDPM_simulation} is terminated at time $T_g - \Delta \tau$ instead of $T_g.$ By doing so, they can bound the KL error between the distribution of generated samples and the forward process distribution at time $\Delta \tau$, which corresponds to a small noise perturbation of the original data distribution.
We do not pursue the early stopping technique in this paper for two main reasons. First, we prefer a simpler and cleaner presentation. Second, applying early stopping to our autoregressive SDE generation raises subtle technical issues when we apply conditional diffusion models sequentially over the time grid. This will be discussed in more detail in Section~\ref{sec:bound-SDE}.

We also note that a recent working paper \citep{fu2024unveil} presents an error analysis of conditional diffusion models with classifier-free guidance \citep{ho2021classifier} in total variation (TV) distance. Proposition \ref{prop: condition_diffusion_KL_bound} in the present paper offers a complementary result, exploring the error bound in KL divergence (which is stronger than TV by Pinsker's inequality) for conditional diffusion models without classifier-free guidance. Moreover, in contrast to \citep{fu2024unveil} which focus on the statistical theory (i.e. sample complexity bounds) of conditional diffusion models without addressing discretization, Proposition \ref{prop: condition_diffusion_KL_bound} provides an error bound that explicitly incorporates discretization errors.

\subsection{Error Bound}\label{sec:bound-SDE}
In this section, we derive the error bound for our SDE generative model, which is the main theoretical result of the paper.
According to Proposition \ref{prop: path_KL_decomposition}, we only need to apply Proposition \ref{prop: condition_diffusion_KL_bound} to bound the KL-error for each conditional diffusion model $G(t_n, \bx(t_n); \theta_n)$. The following two assumptions ensure the conditions of Proposition \ref{prop: condition_diffusion_KL_bound} hold.

\begin{assumption}[Lipschitz SDE increment]\label{assumption: SDE_increment_Lipschitz}
    For any $t_n\in\mathcal{T}$, the increment $\Delta\bX(t_n; \bx(t_n))$ of the target SDE \eqref{eq: target_SDE} conditional on $\bX(t_n) = \bx(t_n)$ admits a twice continuously differentiable density function $P_{\Delta\bX(t_n; \bx(t_n))}$, and $\nabla\log P_{\Delta\bX(t_n; \bx(t_n))}$ is $L_n(\bx(t_n))$-Lipschitz continuous with 
\begin{align*}
    \mE_{\bX(t_n)\sim P_{n|0}}[L_n(\bX(t_n))] = L_1(t_n) < \infty.
\end{align*}
\end{assumption}

\begin{assumption}[Score matching error]\label{assumption: score_matching_error}
    There exists $\varepsilon_{\text{score}} > 0$ such that for each $t_n$ in $\mathcal{T}$, the corresponding learned score function $s_{\theta_n}(\tau_k, \cdot, t_n, \bx(t_n))$ of $G(t_n, \bx(t_n); \theta_n)$ satisfies
    \begin{align*}
       &\frac{1}{K_n}\sum_{k=1}^{K_n}\mE_{\bc_n\sim\bC_n}\bigg[\mE_{\bY(\tau_k)\sim q_{\tau_k}(\cdot | \bc_n)}\big[||\nabla\log q_{\tau_k}(\bY(\tau_k) | \bc_n) - s_{\theta_n}(\tau_k, \bY(\tau_k), \bc_n)||^2\big]\bigg]\le \varepsilon^2_{\text{score}},
    \end{align*}
    where $\bC_n = (t_n, \bX(t_n))$ with $\bX(t_n)\sim P_{n|0}$ as in Proposition \ref{prop: path_KL_decomposition}, the score function $\nabla\log q_{\tau_k}(\cdot | \bc_n)$ is defined based on the forward process \eqref{eq: forward_SDE_DDPM} for which the target distribution is that of $\Delta\bX(t_n; \bx(t_n))$ in \eqref{eq: objective_AR_gen}, and $K_n$ is the number of the diffusion steps of $G(t_n, \bx(t_n); \theta_n)$.
\end{assumption}

In the context of generating SDE paths, Assumption \ref{assumption: SDE_increment_Lipschitz} can be verified to hold for certain special equations such as OU processes. For a more general target SDE, if it is observed at a high frequency, i.e., the time intervals $t_{n+1} - t_n$ are small, then the increment $\Delta\bX(t_n; \bx(t_n))$ is approximately Gaussian by virtue of the Euler--Maruyama method. In this case, Assumption \ref{assumption: SDE_increment_Lipschitz} is also likely to hold. As mentioned earlier, applying the early stopping technique to relax Assumption \ref{assumption: SDE_increment_Lipschitz} is particularly challenging for autoregressive SDE path generation. Specifically, if each increment $\Delta\bX(t_n; \bx(t_n))$ is perturbed by a small amount of noise due to the early stopping, these noises will propagate and accumulate over time along the SDE path, making it difficult to establish a bound on the KL-divergence between the perturbed target SDE paths and generated paths. Hence, we opt out of this approach.

We now state our main theoretical result, which provides an explicit bound on the KL-divergence between the distributions of $P_{1:N_T|0}$ and $\genP_{1:N_T|0}$.

\begin{theorem}\label{thm: SDE_diffusion_convergence}
    Consider a target SDE \eqref{eq: target_SDE} satisfying Assumption \ref{assumption: SDE_solution_supp_abs} and \ref{assumption: SDE_increment_Lipschitz}. If Assumption \ref{assumption: score_matching_error} holds for the conditional diffusion model $G(t_n, \cdot; \theta_n)$ at each $t_n\in\mathcal{T}$, then
    \begin{align*}
    &\KL(P_{1:N_T|0} || \genP_{1:N_T|0}) \\
    & \lesssim  \sum_{n=0}^{N_T - 1} \bigg\{\underbrace{e^{-\frac{a}{2}T_g^2(t_n) - bT_g(t_n)}\big[M_2(t_{n}; T) e^{\Delta tC_M}\Delta t + de^{-\frac{a}{2}T_g^2(t_n) - bT_g(t_n)}\big]}_{initialization\ error} + \underbrace{[T_g(t_n) + 1]^3\varepsilon_{\text{score}}^2}_{score\ matching\ error} \\
    &\quad \qquad \quad+ \underbrace{\frac{1}{K_n}\big(M_2(t_{n}; T) e^{\Delta tC_M}\Delta t + d\big)(T_g(t_n) + 1)^6 + \frac{d \cdot T_g(t_n) \cdot L_1(t_n)}{K_n}}_{discretization\ error}\bigg\},
    \end{align*}
    where $\Delta t$ is the time step of observing \eqref{eq: target_SDE}, $M_2(t_n; T)$ and $C_M$ are defined in \eqref{eq: target_SDE_moment_bound}, and $T_g(t_n)=K_n\Delta\tau_n$ is the diffusion time horizon of $G(t_n, \bx; \theta_n)$.
\end{theorem}

The proof of Theorem~\ref{thm: SDE_diffusion_convergence} directly follows from Proposition~\ref{prop: path_KL_decomposition}, Proposition~\ref{prop: condition_diffusion_KL_bound} and the moment bound of the SDE increment in \eqref{eq: target_SDE_moment_bound}. We can also immediately infer from Theorem~\ref{thm: SDE_diffusion_convergence} and Pinsker's inequality to obtain an error bound in
the total variation (TV) distance between $P_{1:N_T|0}$ and $\genP_{1:N_T|0}$:
    \begin{align*}
        \TV\big(P_{1:N_T|0} || \genP_{1:N_T|0}\big) \le \sqrt{\frac{1}{2} \KL\big(P_{1:N_T|0} || \genP_{1:N_T|0}\big)}.
    \end{align*}
    These obtained bounds quantify the accuracy of our method.

\section{Application to Mean--Variance Portfolio Selection}\label{sec:application}
In this section, we conduct both a simulation study (Section~\ref{sec: application_simulation}) and an empirical one (Section~\ref{sec: application_empirical}) to demonstrate how our generative approach can be useful in financial portfolio optimization.

\subsection{Simulation Study}\label{sec: application_simulation}
We first briefly review the continuous-time mean--variance portfolio selection problem where SDE modeling is essential; see e.g. \cite{wang2020continuous, zhou2000continuous}. An agent invests in a risk-free asset (e.g., a saving account) whose interest rate is  $r$ and a risky asset (e.g., a stock), the price of which is governed by an SDE $\{S(t): t\ge 0\}$, which is a geometric Brownian motion with drift $\mu$ and
volatility $\sigma>0$. The agent
has a fixed investment horizon $T$ and an initial endowment $x_0$.
She rebalances her portfolio continuously with a strategy $a = \{a(t), t\in[0, T]\}$ in a self-financing fashion, where $a(t)$ is the discounted dollar value invested in the risky asset at time $t$. Then her discounted wealth process satisfies \cite[Section 7.1]{jia2023q}:
\begin{align}\label{eq: MV_wealth_process}
    \intd X^a(t) = a_t\frac{\intd \big(e^{-rt}S(t)\big)}{e^{-rt}S(t)}, \ X^a(0) = x(0).
\end{align}
The problem is to find a strategy $a$ that minimizes the variance of the terminal wealth for a given target mean wealth level $z$, i.e.,
\begin{align}\label{eq: MV_objective}
    \min_a \text{Var}(X^a(T)),\ s.t.\ \mathbb{E}[X^a(T)] = z.
\end{align}

Recently, \cite{jia2023q} apply a general model-free continuous-time reinforcement learning (RL) approach, termed as $q-$learning, to solve the above problem when the dynamic of the risky asset $S(\cdot)$ is {\it unknown} (hence following an unknown SDE). 
We now show how synthetic asset price paths can help improve their $q-$learning algorithm by enriching the policy training.

\begin{sidewaystable}[htp]
    \centering
    \small
    % \hskip-1.5cm
    \addtolength{\tabcolsep}{-4.5pt}
    \begin{tabular}{c c|| c c | c c | c c | c c | c c}
    \toprule
       \multicolumn{2}{c||}{setup}  & \multicolumn{2}{c|}{$H (40) + H_s(0)$} & \multicolumn{2}{c|}{$H (40) + H_s(1)$} & \multicolumn{2}{c|}{$H (40) + H_s(10)$} & \multicolumn{2}{c}{$H (40) + H_s(40)$} & \multicolumn{2}{c}{$H (40) + H_s(360)$} \\\hline
        $\mu$ & $\sigma$ & Mean(Var) & Sharpe &   Mean(Var) & Sharpe & Mean(Var) & Sharpe & Mean(Var) & Sharpe & Mean(Var) & Sharpe  \\\hline
        0.5 & 0.4 
        & 1.163 (0.034) & 0.8905 & 1.046 (0.002) & 0.9226 & 1.052 (0.003) & 0.9317 & 1.054 (0.003) & \tbf{0.9324} & 1.054 (0.003) & \tbf{0.9324}
        \\

        0.3 & 0.4 
        & 1.070 (0.020) & 0.4992 & 1.127 (0.059) & 0.5198 & 1.122 (0.055) & \textbf{0.5213} & 1.121 (0.054) & \textbf{0.5213} & 1.060 (0.013) & \textbf{0.5213}
        \\

        0.1 & 0.4 
        & 1.062 (0.241) & 0.1264 & 1.079 (0.301) & \tbf{0.1444} & 1.078 (0.289) & \tbf{0.1444} & 1.077 (0.283) & 0.1443 & 1.079 (0.303) & 0.1441
        \\

        0 & 0.4 
        & 1.094 (7.306) & 0.0347 & 1.020 (0.325) & \textbf{0.0348} & 1.005 (0.019) & 0.0347 & 1.018 (0.274) & \textbf{0.0348} & 1.004 (0.013) & \textbf{0.0348}
        \\

        -0.1 & 0.4
        & 1.220 (2.117) & 0.1510 & 1.204 (1.660) & 0.1582 & 1.192 (1.388) & 0.1626 & 1.188 (1.314) & \tbf{0.1636} & 1.194 (1.413) & \tbf{0.1636}
        \\

        -0.3 & 0.4
        & 1.168 (0.107) & 0.5121 & 1.164 (0.103) & 0.5135 & 1.161 (0.098) & 0.5143 & 1.160 (0.097) & 0.5146 & 1.164 (0.101) & \tbf{0.5150}
        \\

        -0.5 & 0.4 
        & 1.271 (0.112) & 0.8079 & 1.195 (0.054) & 0.8367 & 1.171 (0.041) & 0.8465 & 1.165 (0.038) & \tbf{0.8488} & 1.165 (0.038) & \tbf{0.8488}
        \\ &&&&&&& \\

        0.5 & 0.3
        & 1.193 (0.020) & 1.3797 & 1.200 (0.019) & 1.4503 & 1.202 (0.018) & 1.4914 & 1.203 (0.018) & \textbf{1.4968} & 1.205 (0.019) & \textbf{1.4968}
        \\

        0.3 & 0.3
        & 1.139 (0.043) & 0.6686 & 1.154 (0.052) & 0.6768 & 1.155 (0.052) & 0.6786 & 1.156 (0.053) & 0.6792 & 1.156 (0.053) & \tbf{0.6793}
        \\

        0.1 & 0.3
        & 1.086 (0.273) & 0.1766 & 1.215 (1.444) & 0.1787 & 1.189 (1.100) & 0.1803 & 1.179 (0.980) & \tbf{0.1810} & 1.175 (0.939) & \tbf{0.1810}
        \\

        0 & 0.3 
        & 1.062 (3.990) & 0.0311 & 1.063 (4.106) & 0.0311 & 1.051 (2.236) & 0.0340 & 1.050 (2.138) & \tbf{0.0342} & 1.051 (2.263) & \textbf{0.0342}
        \\

        -0.1 & 0.3
        & 1.228 (1.024) & 0.2257 & 1.207 (0.798) & 0.2317 & 1.199 (0.724) & 0.2341 & 1.199 (0.721) & 0.2341 & 1.198 (0.718) & \tbf{0.2342}
        \\

        -0.3 & 0.3
        & 1.258 (0.151) & 0.6621 & 1.216 (0.102) & 0.6749 & 1.189 (0.076) & 0.6836 & 1.184 (0.072) & \tbf{0.6856} & 1.182 (0.071) & \tbf{0.6856}
        \\

        -0.5 & 0.3 
        & 1.274 (0.063) & 1.0871 & 1.231 (0.044) & 1.1065 & 1.198 (0.031) & 1.1198 & 1.190 (0.029) & \tbf{1.1229} & 1.189 (0.028) & \tbf{1.1229}
        \\ &&&&&&& \\

        0.5 & 0.2 
        & 1.190 (0.003) & 3.3755 & 1.190 (0.003) & 3.5470 & 1.192 (0.003) & 3.7618 & 1.191 (0.002) & \tbf{3.9119} & 1.191 (0.002) & \tbf{3.9119}
        \\

        0.3 & 0.2
        & 1.205 (0.030) & 1.1776 & 1.205 (0.027) & 1.2524 & 1.205 (0.026) & 1.2770 & 1.207 (0.025) & \tbf{1.3135} & 1.210 (0.026) & \tbf{1.3135}
        \\

        0.1 & 0.2
        & 1.158 (0.302) & 0.2885 & 1.209 (0.522) & 0.2899 & 1.227 (0.614) & 0.2903 & 1.224 (0.595) & \tbf{0.2905} & 1.220 (0.572) & 0.2904
        \\

        0 & 0.2
        & 1.116 (3.187) & 0.0649 & 1.103 (2.519) & 0.0650 & 1.167 (6.557) & 0.0652 & 1.158 (5.851) & \tbf{0.0655} & 1.054 (0.691) & 0.0654
        \\

        -0.1 & 0.2
        & 1.155 (0.133) & 0.4251 & 1.214 (0.238) & 0.4390 & 1.223 (0.258) & 0.4390 & 1.224 (0.261) & \tbf{0.4391} & 1.232 (0.278) & \textbf{0.4391}
        \\

        -0.3 & 0.2
        & 1.246 (0.059) & 1.0115 & 1.176 (0.029) & 1.0412 & 1.160 (0.023) & 1.0485 & 1.155 (0.022) & \tbf{1.0508} & 1.155 (0.022) & \tbf{1.0508}
        \\

        -0.5 & 0.2 
        & 1.270 (0.027) & 1.6316 & 1.173 (0.010) & 1.6972 & 1.162 (0.009) & 1.7053 & 1.151 (0.008) & \tbf{1.7131} & 1.151 (0.008) & \tbf{1.7131}
        \\
        \bottomrule

    \end{tabular}
    \addtolength{\tabcolsep}{1pt}
    \caption{Out-of-sample performance of the $q-$Learning algorithm (Algorithm \ref{alg: qL_mv}) in terms of mean, variance and Sharpe ratio. Here, $H=40$ ``real'' asset price paths and $H_s$ synthetic price paths are used in each learning episode of Algorithm \ref{alg: qL_mv}.}
    \label{table: qL_mv_simulation_more}
\end{sidewaystable}

\begin{table}[htp]
    \centering
    \begin{tabular}{c || c | c | c | c}
        \toprule
        Improvement (\%) & $H_s: 0 \Rightarrow  1$ & $H_s : 0 \Rightarrow 10$ & $H_s: 0 \Rightarrow 40$ & $H_s : 0 \Rightarrow  360$\\\hline
        Maximum & 14.2405 & 14.2405 & 15.8909 & 15.8908\\
        Minimum & 0 & 0 & 0.2882 & 0.2882\\
        Median & 2.7973 & 3.6898 & 3.8035 & 3.8257 \\
        Average & 3.1940 & 4.7423 & \tbf{5.3237} & 5.3136\\
        \bottomrule
    \end{tabular}
    \caption{Sharpe ratio improvement percentage for $q-$learning policy trained by Algorithm \ref{alg: qL_mv} when different numbers of synthetic price paths are added in each learning episode.}
    \label{table: qL_mv_simulation_improvement}
\end{table}

In our simulation study, we take $X^a(0)=1$, $r=2\%$, $T=0.5$, $S(0)=1$, and $z=1.2$. 
We consider different parameter configurations with drift $\mu\in\{0, \pm 0.1, \pm 0.3, \pm 0.5\}$ and volatility $\sigma\in\{0.2, 0.3, 0.4\}$ for simulating the market environment. For each pair of $(\mu, \sigma)$, we are given only $H=40$ ``real'' GBM paths simulated via Monte Carlo with a time step $\Delta t =1/252$ representing one trading day.
These 40 GBM paths constitute the training dataset to train our SDE generative model, whose specific settings are shown in Appendix \ref{sec: experiment_settings}. We implement the $q-$learning algorithm in \cite{jia2023q} with various choices of $H_s$, the number of synthetic GBM paths generated by our model at each episode to augment the original 40 training paths. This includes $H_s = 0, 1, 10, 40$ and $360$. The details of the RL algorithm we implement are given as Algorithm \ref{alg: qL_mv} in Appendix \ref{sec: application_simulation_more}, which is a direct extension of Algorithm~5 in \cite{jia2023q} by including synthetic price paths.
After training the RL policies with Algorithm \ref{alg: qL_mv}, we perform out-of-sample tests on 100,000 GBM paths (simulated from the oracle model via Monte Carlo) to empirically estimate $\mathbb{E}[X^a(T)]$ and Var$(X^a(T))$ and calculate the corresponding Sharpe ratio, which is $(\mE[X^a(T)] - 1)/\sqrt{\text{Var}(X^a(T))}$.

The results are presented in Tables \ref{table: qL_mv_simulation_more} and \ref{table: qL_mv_simulation_improvement}, which compare the performances of the RL algorithm, Algorithm \ref{alg: qL_mv}, with various numbers of additional synthetic price paths ($H_s = 1, 10, 40,$ and $360$) together with the case where no synthetic paths are used ($H_s=0$). We summarize the key findings as follows:
\begin{itemize}
\item Incorporating synthetic paths into the RL training dataset substantially enhances the Sharpe ratio of the resulting RL policies across nearly all parameter configurations (21 in total).
\item Even adding one single synthetic path per learning episode ($H_s=1$) yields a notable improvement, with an average Sharpe ratio increase of approximately $3.19\%$ across the 21 parameter settings.
\item The most significant performance gain is observed when $H_s=40$, with an average Sharpe ratio improvement of around $5.32\%$ over the case of pure ``real" paths. As indicated in Table \ref{table: qL_mv_simulation_more}, this improvement is largely attributed to reduced variance of terminal wealth with similar mean return when synthetic paths are included in training.
\item Increasing $H_s$ to 360 does not provide additional benefits, while doubling the computation time compared to the case of $H_s=40$. This suggests that, in practice, a moderate number of synthetic paths (e.g., on the same order as the number of training paths) is sufficient to train effective investment policies using Algorithm \ref{alg: qL_mv}. 
\end{itemize}

\subsection{Empirical Study}\label{sec: application_empirical}

We now conduct an empirical study to demonstrate how our generative approach can be useful practically in portfolio optimization, despite the fact that real financial time series may not be discrete observations of SDEs and may not even be Markovian.

We still consider the mean--variance portfolio selection problem in \eqref{eq: MV_objective}, with S\&P 500 index as the risky asset and a riskless asset with interest rate $r= 2\%$. The investment horizon is set to be half a year in the empirical study.
A classical model-based solution to \eqref{eq: MV_objective} is to use historical data of S\&P 500 to estimate a GBM model and then plug the estimated model parameters in the analytical expression of the optimal deterministic policy of \eqref{eq: MV_objective}, which is known from \cite{zhou2000continuous}. The resulting policy (called  \textit{plug-in policy}) has been shown to be inferior compared with the model-free \textit{RL policy} proposed by \cite{ jia2023q}.

We now describe the setup of our empirical study. We first split daily observations of S\&P 500 data from the start of 1990 to the end of 2009 into 40 half-year trajectories (henceforth called \textit{split paths}), and normalize each split path to start from 1. These paths are used to train our initial generative model. As in the SDE setting, we simulate the daily increment of S\&P 500, and autoregressively generate a synthetic index path for half a year. These synthetic paths can be pooled together with the original 40 split paths to either estimate the GBM coefficients (for plug-in policies) or train RL policies (similar to those in Section~\ref{sec: application_simulation}).
In contrast to the simulation study where multiple test paths can be simulated from a known ground-truth GBM model, out-of-sample tests of different policies in our empirical analysis need to be based on a {\it single} realized path of the S\&P 500 index. We consider the test period from 2010 to 2019 and  employ a rolling window approach to test the performance of different policies.
 Every half a year, we apply the plug-in and RL policies respectively derived from historical data to the upcoming half-year S\&P 500 index trajectory. Meanwhile, the newly observed index trajectory is added to our pool of historical split paths to be used to re-train the generative model and re-estimate the GBM parameters for updating the two policies in the next cycle.

Specifically, after the $k-$th half-year from the beginning of 2010 , $0\le k\le 19$, we re-train our generative model based on $(40+k)$ split paths to generate synthetic paths.
Two plug-in policies are then respectively developed when only the $(40 + k)$ split paths are used and when the $(40 + k)$ split and $(40 + k)$ synthetic paths are combined to estimate the parameters of GBM. For the RL policy trained via Algorithm \ref{alg: qL_mv}, we consider three different training sets in each learning episode: the first one consists of the $(40 + k)$ split paths only, the second one has $(40 + k)$ bootstrap paths similarly as in \cite{jia2023q},\footnote{Precisely, to generate
$B$ bootstrap paths, we randomly sample $B$ starting points from the S\&P 500 price trajectory between the beginning of 1990 and six months prior to the current date, and then append six months of subsequent index data to each starting point and normalizing the starting point to 1.} and the last one is based on the $(40 + k)$ split and $(40 + k)$ synthetic paths.\footnote{The finding from Section~\ref{sec: application_simulation} suggests that we take the same number of synthetic paths.} Then, we test the resulting two plug-in policies and three RL policies on the following $(k + 1)$th half-year S\&P 500 trajectory. We repeat this procedure until the end of 2019, and obtain 20 pairs of initial and terminal wealth for each policy. Note that in this empirical analysis, we also incorporate a practical consideration: if the wealth process associated with a trading strategy reaches zero at any point in time, it remains at zero thereafter, registering a bankruptcy event.
We evaluate the performance of different policies by calculating the sample mean and variance of the return (i.e., the percentage change between the initial and terminal wealth) and the corresponding Sharpe ratio.

\begin{table}[htp]
    \centering
    \begin{tabular}{c | c | l || c c c}
        \toprule
        Target & Policy & SDE Paths & Mean Return & Variance & Sharpe Ratio $\uparrow$ \\
        \hline
        \multirow{5}{*}{$z = 1.10$} & \multirow{2}{*}{Plug-in} & Split & 0.1676 & 0.0371 & 0.8703\\
        && Split + Synthetic & 0.1709 & 0.0339 & 0.9282 \\
        \cline{2-6}
        &\multirow{3}{*}{RL} & Split & 0.1075 & 0.0019 & 2.4503\\
        && Bootstrap & 0.1143 & 0.0019 & 2.6110\\
        && Split + Synthetic & 0.1241 & 0.0016 & 3.1033\\
        \hline
        \multirow{5}{*}{$z = 1.20$} & \multirow{2}{*}{Plug-in} & Split & -0.8029 & 0.3283 & -1.4014\\
        && Split + Synthetic & -0.8051 & 0.3100 & -1.4461 \\
        \cline{2-6}
        &\multirow{3}{*}{RL} & Split & 0.2143 & 0.0079 & 2.4104\\
        && Bootstrap & 0.2268 & 0.0076 & 2.6064\\
        && Split + Synthetic & 0.2469 & 0.0064 & 3.0783\\
          \hline
         -- & Buy and hold &  & 0.0572 & 0.0061 & 0.7331\\
        \bottomrule
    \end{tabular}
    \caption{Performances of different portfolio strategies with and without synthetic paths, along with the buy-and-hold benchmark.}
    \label{table: qL_mv_empirical}
\end{table}

\begin{figure}[htp]
    \centering
    \includegraphics[scale=0.8]{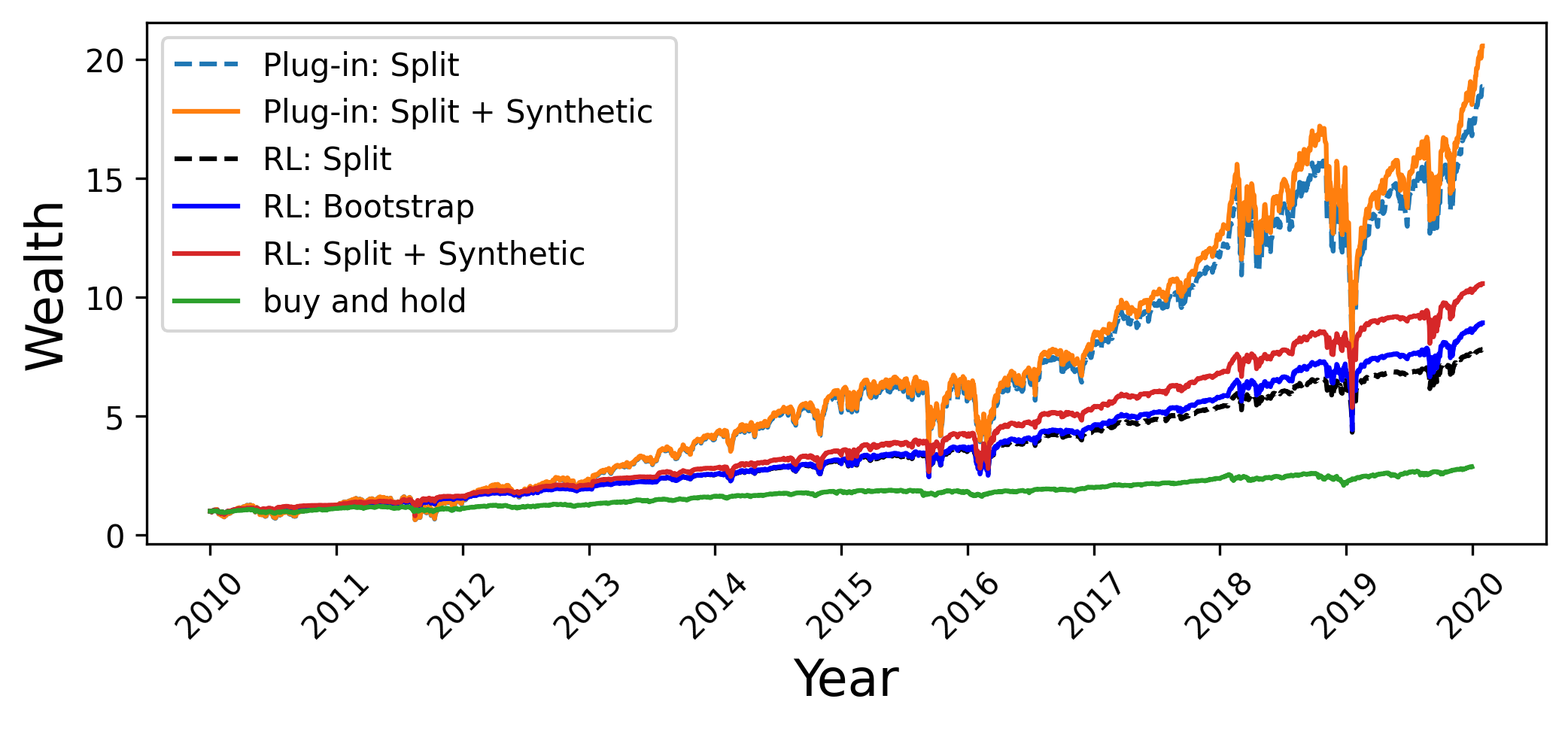}
    \caption{Wealth trajectories of two plug-in policies and three RL policies in the out-of-sample test when the target level $z=1.1$. The initial wealth is assume to be one. The ``buy and hold" trajectory is the S\&P 500 index normalized by its price at the beginning of 2010.}
    \label{fig: qL_mv_empirical_wealth_z11}
\end{figure}

\begin{figure}[htp]
\begin{minipage}{0.48\linewidth}
    \centering
    \includegraphics[scale=0.7]{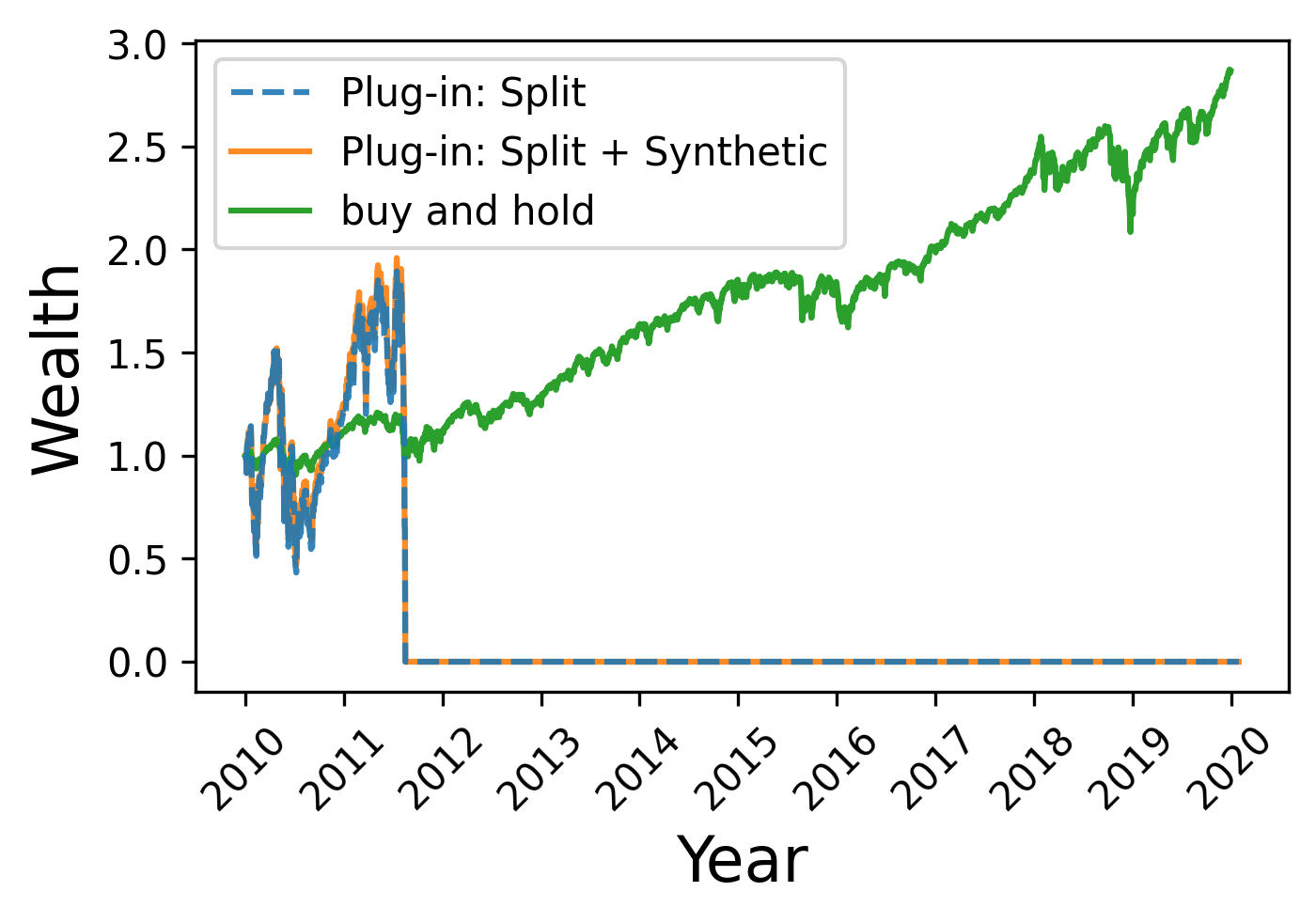}
\end{minipage}
\hfill
\begin{minipage}{0.48\linewidth}
    \centering
    \includegraphics[scale=0.7]{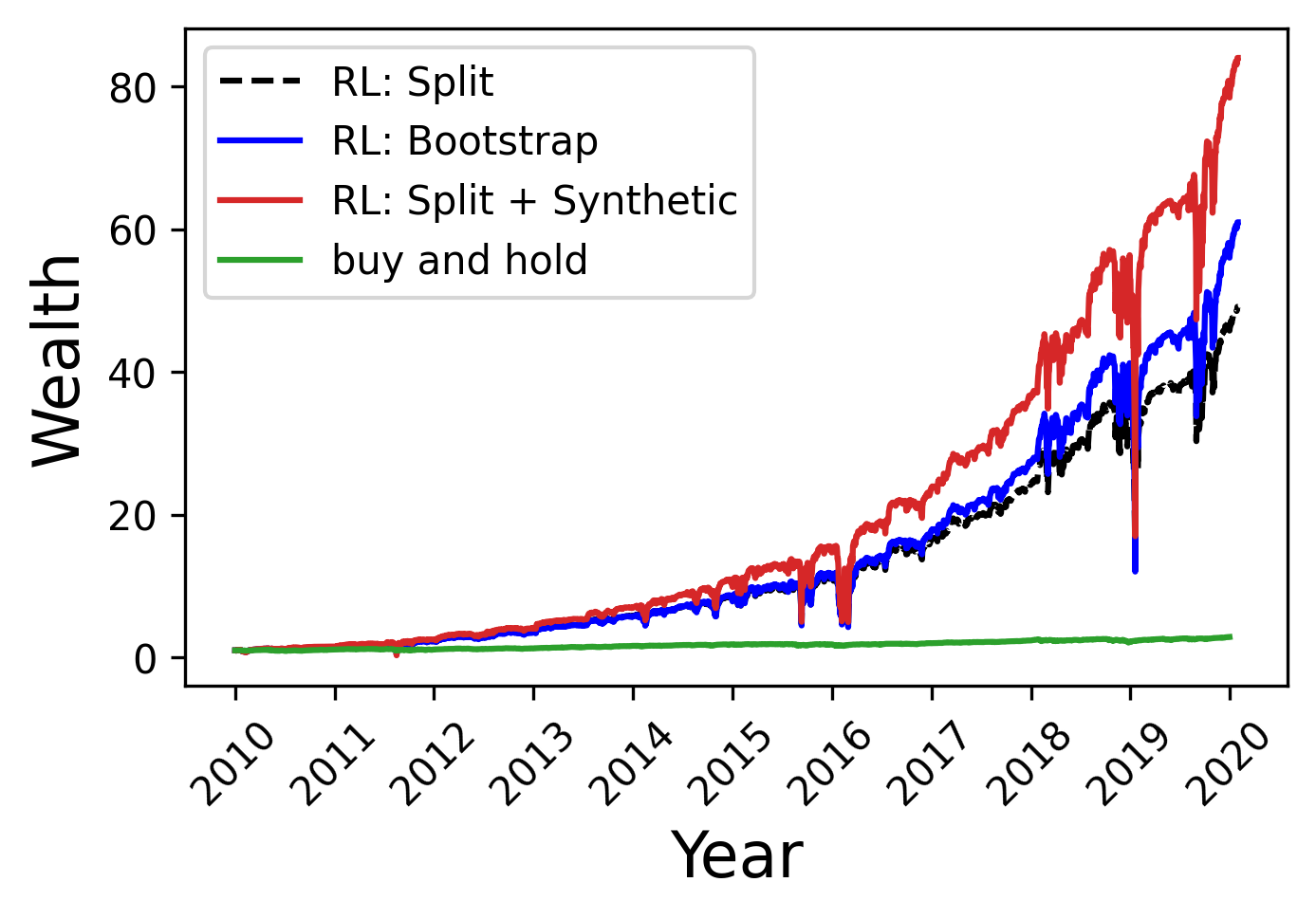}
\end{minipage}
\vspace{-2mm}
\caption{Wealth trajectories of two plug-in policies and three RL policies in the out-of-sample test when the target level $z=1.2$. The initial wealth is assume to be one. The ``buy and hold" trajectory is the S\&P 500 index normalized by its price at the beginning of 2010.}
\label{fig: qL_mv_empirical_wealth_z12}
\end{figure}

We present the results in Table \ref{table: qL_mv_empirical} where the numbers are rounded to four decimal places, and plot the wealth trajectories of those portfolio strategies when $z=1.1$ and $z=1.2$ in Figure \ref{fig: qL_mv_empirical_wealth_z11} and Figure \ref{fig: qL_mv_empirical_wealth_z12}, respectively. We also include the ``buy-and-hold" strategy of S\&P 500 index as a benchmark for the market.  There are several notable observations.
First, including synthetic paths may not necessarily improve the plug-in policy in terms of the Sharpe ratio as seen in the case of $z=1.2$ (i.e. the target return is $20\%$). This reinforces our earlier finding (Example 1) that generated paths are not helpful in improving the estimation of structural model parameters (drift and volatility). Note that with the aggressive target $z=1.2$, Figure \ref{fig: qL_mv_empirical_wealth_z12} shows that both plug-in policies suffered total ruin (zero wealth) in August 2011 following a sudden market/index downturn. This can be attributed to parameter misspecification: the strong bull market in the first half of 2011 led to an overestimate of the GBM drift. As a result, the plug-in strategy took overly aggressive leverage positions in pursuit of the high return target, leaving it vulnerable to a sudden market reversal. The subsequent market downturn ultimately led to the complete depletion of wealth.
Importantly,  augmenting real data with synthetic paths did not help -- indeed it made the Sharpe ratio even lower and failed to mitigate the risk of bankruptcy.

Second, the RL strategies significantly outperformed the model-based plug-in ones in Sharpe ratio,  consistently achieving/exceeding the given target returns throughout the test period and massively reducing terminal variances.
Moreover, all the RL strategies maintained positive wealth throughout the test period for both target levels $z=1.1$ and $z=1.2$, demonstrating their robustness and resilience. Finally and most importantly, adding synthetic paths to the RL training dataset provides a further boost in performance to the RL method with bootstrapped paths in \cite{jia2023q} with approximately 18\% increase in Sharpe ratio for both $z=1.1$ and $z=1.2$. The grand conclusion is that AI-generated data cannot be employed to better understand the unknown environment, but provides more training ``playgrounds'' to train better decision policies with more robust performances.

%%%%%%%%%%%%%%%%%%%%%%%%%%%%%%%%%%%%%%%%%%%%%%%%%%%%%%%%%%%%%%%%%

\section{Conclusions}\label{sec:conclusion}

In this paper, we provide a model-free, data-driven approach for generating sample paths of unknown Markovian SDEs via conditional diffusion models. Numerical experiments demonstrate the superiority of our method over two benchmark approaches.  Moreover, we establish a KL error bound for our path generation approach, giving a theoretical guarantee on its accuracy. As a concrete application, we make use of the generated sample paths to enhance the performance of RL algorithms in model-free continuous-time mean--variance portfolio selection, illustrating the potential of synthetic paths in creating more diverse scenarios for training decision-making strategies. A key insight from the study is that AI-generated paths do not help model parameter estimation, but enhance decision capabilities.

This work opens the gate to several directions of future research. First and foremost, it will be important to remove the Markovian assumption for both theoretical and practical reasons. In this case, the conditioning variables will be all the historical time--state pairs, resulting in a much higher dimensional problem. The theoretical error analysis will need to involve {\it functional} (or path-dependent) SDEs. Next, it will be interesting to study path generations for SDEs with jumps or those driven by fractional Brownian motions, the latter being an instance of non-Markovian equations. Moreover, relaxing the assumptions on the target SDE and establishing error bounds for alternative metrics, such as the Wasserstein distance, will further strengthen the theoretical foundation of our method. Another intriguing direction is to investigate the potential of diffusion models in what-if simulation analysis (e.g., what if the volatility increases by $10\%$) and counterfactual generations. Finally, exploring the use of synthetic data or paths in other decision-making problems may yield more valuable insights. In sum, the application of diffusion models to strategic data generations represents a largely uncharted territory where many exciting researches await.

\bibliographystyle{plainnat}
\bibliography{sample}

\newpage
\appendix

 \section{Proof of Proposition \ref{prop: path_KL_decomposition}}\label{sec: path_KL_proof}

\begin{proof}[Proof of Proposition \ref{prop: path_KL_decomposition}]

Throughout this proof, if a probability distribution is absolutely continuous w.r.t the Lebesgue measure, we simply say this distribution is ACL.

We first consider the target paths. For any $t_n\in\mathcal{T}$, denote by $P_{n + 1| n}(\cdot | \bx(t_n))$ the conditional distribution of $\bX(t_{n + 1})$ given $\bX(t_n) = \bx(t_n)$, which is ACL and positive almost everywhere on $\mR^d$ according to Assumption \ref{assumption: SDE_solution_supp_abs}(b). It follows that the conditional distribution $P_{\Delta\bX(t_n; \bx(t_n))}$ of the increment is also ACL and positive almost everywhere on $\mR^d$.
Therefore, for any $A\subset \mR^d$,
\begin{align}\label{eq: equivalent_to_lebesgue}
    \int_{\bx\in A}1 \intd \bx = 0\quad  \text{ if and only if }\quad \int_{\bx \in A}1\cdot \frac{\intd P_{\Delta\bX(t_n; \bx(t_n))}(\bx)}{\intd \bx} \intd \bx= 0,
\end{align}
where $\frac{\intd P_{\Delta\bX(t_n; \bx(t_n))}(\bx)}{\intd \bx}$ is the Radon–Nikodym derivative (i.e., the density function) of $P_{\Delta\bX(t_n; \bx(t_n))}(\bx)$ w.r.t. the Lebesgue measure. This implies that $P_{\Delta\bX(t_n; \bx(t_n))}$ and the Lebesgue measure are equivalent.

Next, consider $P_{1:N_T|0}$. For any $0 \le m < n\le N_T$,  let $\bX(m:n):=(\bX(t_m), \bX(t_{m + 1}),\cdots, \bX(t_n))$, and denote by $P_{m:n|0}$ the joint distribution of $\bX(m:n)$ conditional on $\bX(t_0)\equiv\bx(t_0)$. Let $\bx(m:n) = (\bx(t_m), \cdots, \bx(t_n))$ be a realization of $\bX(m:n)$. Due to the Markov property of the Target SDE \eqref{eq: target_SDE}, we have
\begin{align}
    \intd P_{m:n|0}(\bx(m:n)) &= \intd \bigg[P_{m|0}(\bx(t_m))\prod_{j=m}^{n - 1}P\big(\bX(t_{j + 1})=\bx(t_{j + 1}) | \bX(m:j)=\bx(m:j)\big)\bigg] \nonumber\\
    &= \intd \bigg[P_{m|0}(\bx(t_m))\prod_{j=m}^{n - 1}P\big(\bX(t_{j + 1})=\bx(t_{j + 1}) | \bX(t_j)=\bx(t_j)\big)\bigg] \nonumber\\
    &= \intd P_{m|0}(\bx(t_m))\prod_{j=m}^{n - 1}\intd P_{j + 1 | j}\big(\bx(t_{j + 1}) | \bx(t_j)\big), \label{eq: target_Markov}
\end{align}
 where we recall that $P_{m|0}(\bx(t_m))$ and $P_{j + 1 | j}\big(\bx(t_{j + 1}) | \bx(t_j)\big)$ are ACL and positive almost everywhere on $\mR^d$ for any $m$ and $j$ because of Assumption \ref{assumption: SDE_solution_supp_abs}(b). Hence, $P_{m:n|0}$ is ACL and positive almost everywhere on $\mR^{d\times (n - m + 1)}$ for any $m < n$. In particular, $P_{1:N_T|0}$ is  ACL and positive almost everywhere on $\mR^{d\times N_T}$.
 Using a similar argument as in \eqref{eq: equivalent_to_lebesgue}, we deduce that $P_{1:N_T|0}$ is equivalent to the Lebesgue measure.

On the other hand, consider the distribution of the generated increment $\genP_{\Delta\genX(t_n; \bx(t_n))}$ for a given $t_n$ and $\bx(t_n)$. Note that $\Delta\genX(t_n; \bx(t_n))$ is the terminal state (i.e. $\tilde\bZ(\tau_K)$) at the last step of the generation process \eqref{eq: reverse_SDE_DDPM_simulation}, which means it is the sum of a random variable determined by the state of the second last step of \eqref{eq: reverse_SDE_DDPM_simulation} (i.e. $ \tilde\bZ(\tau_{K - 1})$)
and an independent Gaussian variable (with mean $0$ and variance $g^2(T_g - \tau_{K - 1})\Delta\tau \cdot I_d$).
It follows that $\genP_{\Delta\genX(t_n; \bx(t_n))}$ is ACL and positive almost everywhere on $\mR^d$ by Corollary 2.2 of \cite{bhattacharya2007basic}, and consequently  also equivalent to the Lebesgue measure for each $n$.

By the autoregressive generation \eqref{eq: AR_generation}, the conditional distribution $\genP_{n + 1|n}(\cdot |\hat\bx(t_n))$ of $\genX(t_{n + 1})$ conditional on $\genX(t_n)=\hat\bx(t_n)$ is ACL and positive almost everywhere on $\mR^d$ for each $n$.
 Using similar arguments leading to  \eqref{eq: equivalent_to_lebesgue} and \eqref{eq: target_Markov}, we conclude that the distribution $\genP_{1:N_T|0}$ of the entire generated path $\big(\genX(t_1, \cdots, \genX(t_{N_T})\big)$ is equivalent to the Lebesgue measure and positive almost everywhere on $\mR^{d\times N_T}$.

Finally, we establish the decomposition result \eqref{eq: path_KL_decomposition} for the KL-divergence between the target and generated paths. Note that given $\bX(t_0)\equiv \bx(t_0)$, the Markov property holds for both $\bX(1:N_T)$ and $\genX(1:N_T)$. Then, the chain rule of KL-divergence \citep[Exercise 3.2]{wainwright2019high} implies that %\footnote{A proof of the chain rule property can be found in this \href{https://high-dimensional-statistics.github.io/2020/09/05/exercise-3.2.html}{note}.}
\begin{align*}
    \KL\big(P_{1:N_T|0} || \genP_{1:N_T|0}\big) &= \sum_{n=0}^{N_T - 1}\mE_{\bx(0:n)\sim \bX(0:n)| \bX(t_0)} \bigg[\KL\big(P_{\bX(t_{n + 1}) |\bX(0:n)=\bx(0:n)} \big|\big| \genP_{\genX(t_{n + 1}) |\genX(0:n)=\bx(0:n)}\bigg]\\
    &=\sum_{n=0}^{N_T - 1}\mE_{\bx(0:n)\sim \bX(0:n)| \bX(t_0)} \bigg[\KL\big(P_{\bX(t_{n + 1}) |\bX(t_n)=\bx(t_n)} \big|\big| \genP_{\genX(t_{n + 1}) |\genX(t_n)=\bx(t_n)}\bigg]\\
    &= \sum_{n=0}^{N_T - 1}\mE_{\bx(t_n)\sim \bX(t_n)| \bX(t_0)} \bigg[\KL\big(P_{\Delta\bX(t_n; \bx(t_n))} \big|\big| \genP_{\Delta\genX(t_n; \bx(t_n))}\bigg]\\
    &= \sum_{n=0}^{N_T - 1}\mE_{\bX(t_n)\sim P_{n|0}}\bigg[\KL\big(P_{\Delta\bX(t_n; \bX(t_n))} \big|\big| \genP_{\Delta\genX(t_n; \bX(t_n))}\big)\bigg].
\end{align*}
This completes the proof.
\end{proof}

\begin{figure}[htp]
\centering
\begin{tikzpicture}[
    auto,
    node distance=1cm and 1.5cm,
    block/.style={
        rectangle,
        rounded corners,
        draw,
        text width=8em,
        text centered,
        minimum height=2em
    }
]

% Define the nodes
\node[block, text width=8em] (lemma1) {Lemma \ref{lemma: KL_bound_decomposition}\\ Error sources identification};

\node[block, right=0.8cm of lemma1, text width=7em] (lemma6) {Lemma \ref{lemma: KL_dt}\\ KL evolution};

\node[block, below=0.3cm of lemma1, text width=8em] (lemma2) {Lemma \ref{lemma: initial_dist_error_bound}\\ Initialization error};
\node[block, below=0.3cm of lemma2, text width=8em] (lemma3) {Lemma \ref{lemma: state_discrete_error_bound}\\ State \\ discretization};
\node[block, below=0.3cm of lemma3, text width=8em] (lemma4) {Lemma \ref{lemma: score_discretization_error_bound}\\ Score\\ discretization};
\node[block, below=0.3cm of lemma4, text width=8em] (lemma5) {Lemma \ref{lemma: BM_discretization_error_bound}\\ Brownian motion discretization};

\node[block, right=0.7cm of lemma4, text width=7em] (lemma7) {Lemma \ref{lemma: score_diff_bound_large_t}\\ Stochastic localization for $\tau > 0$};

\node[block, right=0.7cm of lemma4, yshift=2.3cm, text width=7em] (lemma8) {Lemma \ref{lemma: lipschitz_score_diff_bound}\\ Lipschitz score for small $\tau$};

\node[block, left=0.6cm of lemma3, text width=11em] (prop2) {Proposition \ref{prop: condition_diffusion_KL_bound}\\ Target bound on\\ {\small$\mE_{\bc\sim\bC}\big[\KL(p_\tar(\cdot | \bc) || \tilde p_{\tau_K}^\theta)\big]$}};

\node[block, right=4.2cm of lemma2, yshift=0.3cm, text width=10em] (lemma12) {Lemma \ref{lemma: diffusion_time_change}\\ Diffusion model as Stochastic localization};
\node[block, below=0.6cm of lemma12, text width=10em] (lemma11) {Lemma \ref{lemma: hessian_score_bound}\\ Score Hessian bound};
\node[block, below=0.6cm of lemma11, text width=9em] (lemma9) {Lemma \ref{lemma: stochastic_char_score_diff}\\ Space and time discretization error};
\node[block, below=0.6cm of lemma9, text width=9em] (lemma10) {Lemma \ref{lemma: score_function_norm_bound}\\ Score norm bound};

% Draw the arrows
\begin{scope}[-Triangle]
    \draw (lemma6.west) -- (lemma1.east);
    \draw (lemma1.west) -- (prop2.20);

    \draw (lemma2.west) -- (prop2.10);
    \draw (lemma3.west) -- (prop2.east);
    \draw (lemma4.west) -- (prop2.-10);
    \draw (lemma5.west) -- (prop2.-20);

    \draw (lemma7.west) -- (lemma4.east);
    \draw (lemma8.west) -- (lemma4.20);
    % \draw (lemma8.west) -- (lemma5.10);
    \draw (lemma10.west) -- (lemma5.4);

    \draw (lemma9.west) -- (lemma7.7);

    \draw (lemma10.north) -- (lemma9.south);
    \draw (lemma11.south) -- (lemma9.north);

    \draw (lemma12.south) -- (lemma11.north);
    \draw (lemma10.west) -- (lemma7.-10);

\end{scope}

\end{tikzpicture}
\caption{Roadmap of Proving Proposition \ref{prop: condition_diffusion_KL_bound}}
\label{fig:placeholder}
\end{figure}

\section{Proof of Proposition \ref{prop: condition_diffusion_KL_bound}}\label{sec: convergence_proof}

We first define some constants that will be used throughout the proof. The forward equation \eqref{eq: forward_SDE_DDPM}  admits an analytical solution
\begin{align}
    \bY(s) &= e^{-\int_\tau^sf(v)\intd v}\bY(\tau) + \int_{\tau}^s e^{-\int_u^sf(v)\intd v}g(u) \intd \bB(u) \nonumber \\
    & \overset{d}{=}  e^{-\frac{a}{4}(s^2 - \tau^2) - \frac{b}{2}(s - \tau)}\bY(\tau) + \sqrt{1 - e^{-\frac{a}{2}(s^2 - \tau^2) - b(s - \tau)}}\cdot\zeta, \label{eq: DDPM_conditional_relation}
\end{align}
for any $0 < \tau < s < T_g$, where $\zeta\sim\mathcal{N}(0, I_d)$ is independent of $\bY(\tau)$. Let
\begin{align}\label{eq: forward_SDE_coeff}
    \lambda_{\tau, s} := e^{-\frac{a}{4}(s^2 - \tau^2) - \frac{b}{2}(s - \tau)}, \quad \lambda_\tau:= \lambda_{0,\tau}, \quad \sigma^2_{\tau, s} := 1 - e^{-\frac{a}{2}(s^2 - \tau^2) - b(s - \tau)},\quad \text{and}  \quad  \sigma_{\tau}^2 = \sigma^2_{0,\tau}.
\end{align}
 We then have $\bY(s) | \bY(\tau) \sim \mathcal{N}(\lambda_{\tau, s}\bY(\tau), \sigma^2_{\tau, s}I_d)$, and we denote by $q_{s|\tau}(\cdot | \by, \bc)$ the corresponding Gaussian density when $\bY(\tau) = \by$ under the conditional $\bc$.

The proof of Proposition \ref{prop: condition_diffusion_KL_bound} is long, accomplished through  several lemmas. For reader's convenience, Figure~\ref{fig:placeholder} provides the relations and logical flows of various lemmas in proving
Proposition \ref{prop: condition_diffusion_KL_bound}.

\begin{proof}[Overview of the Proof of Proposition \ref{prop: condition_diffusion_KL_bound}]

Given a fixed condition $\bc$, we first state Lemma \ref{lemma: KL_bound_decomposition} below, which decomposes $\KL\big(p_\tar(\cdot | \bc) || \tilde p^\theta_{\tau_K}(\cdot | \bc)\big)$ and identifies various sources of approximation errors.

\begin{lemma}\label{lemma: KL_bound_decomposition}
For a given condition $\bc$, denote by $\tilde p^\theta_{\tau_K}(\cdot | \bc)$ the marginal density of the generated $\tilde\bZ(\tau_K)$ of the reverse process \eqref{eq: reverse_SDE_DDPM_simulation}. Under the conditions of Proposition \ref{prop: condition_diffusion_KL_bound}, we have
\begin{align}
        &\KL\big(p_\tar(\cdot | \bc) || \tilde p^\theta_{\tau_K}(\cdot | \bc)\big) \nonumber\\
        &\lesssim \KL\big(q_{T_g}(\cdot | \bc) || \mathcal{N}(0, I_d)\big) +  \sum_{k=1}^{K} \int_{\tau_{k-1}}^{\tau_k}\mE_{\bY(\tau)\sim q_\tau(\cdot | \bc), \bY(\tau_k)\sim q_{\tau_k|\tau}(\cdot |\bY(\tau), \bc)}\big[||f(\tau)\bY(\tau) - f(\tau_k)\bY(\tau_k)||^2\big]\intd\tau \nonumber\\
        &\qquad + \sum_{k=1}^{K} \int_{\tau_{k-1}}^{\tau_k}\mE_{\bY(\tau)\sim q_\tau(\cdot | \bc), \bY(\tau_k)\sim q_{\tau_k | \tau}(\cdot | \bY(\tau), \bc)}\big[||g^2(\tau)\nabla\log q_\tau(\bY(\tau) | \bc) - g^2(\tau_k)\nabla\log q_{\tau_k}(\bY(\tau_k) | \bc)||^2\big]\intd\tau\nonumber\\
        &\qquad +  \sum_{k=1}^{K}\int_{\tau_{k - 1}}^{\tau_{k}}[g^2(\tau_k) - g^2(\tau)]^2\mE_{\bY(\tau)\sim q_{\tau}(\cdot | \bc)} \big[\big\|\nabla\log q_\tau(\bY(\tau) | \bc)\big\|^2\big] \intd\tau \nonumber \\
        &\qquad + \Delta \tau\sum_{k=1}^{K}\mE_{\bY\sim q_{\tau_k}(\cdot | \bc)}\big[||g^2(\tau_k)\nabla\log q_{\tau_k}(\bY(\tau_k) | \bc) - g^2(\tau_k)s_\theta(\tau_k, \bY(\tau_k), \bc)||^2\big].\label{eq: KL_bound_decomposition}
\end{align}
\end{lemma}

The proof of Lemma \ref{lemma: KL_bound_decomposition} can be found in Appendix \ref{sec: error_identification}. We next discuss how to bound each term on the right-hand-side of \eqref{eq: KL_bound_decomposition}.

The first term of \eqref{eq: KL_bound_decomposition} is the initialization error arising from the use of the Gaussian noise, instead of the terminal state of the forward process, as the initialization of the generation process \eqref{eq: reverse_SDE_DDPM_simulation}. We bound this initialization error in Lemma \ref{lemma: initial_dist_error_bound}, leveraging on the exponential convergence property of the forward OU process \eqref{eq: forward_SDE_DDPM}. The proof of Lemma~\ref{lemma: initial_dist_error_bound} is given in Appendix \ref{sec: Gaussian_initialization_error}.

\begin{lemma}\label{lemma: initial_dist_error_bound}
For the terminal marginal density $q_{T_g}(\cdot | \bc)$ induced by the forward process defined in \eqref{eq: forward_SDE_DDPM}, if the moment condition \eqref{eq: convergence_finite_moment} is satisfied, then we have:
    \begin{align*}
        \mE_{\bc\sim\bC}\bigg[\KL\big(q_{T_g}(\cdot | \bc) || \mathcal{N}(0, I_d)\big)\bigg]\lesssim e^{-\frac{a}{2}T_g^2 - bT_g}\big[M_2 + de^{-\frac{a}{2}T_g^2 - bT_g}\big].
    \end{align*}
\end{lemma}

Next, we bound the second term of \eqref{eq: KL_bound_decomposition}, which accounts for the state discretization error of using $f(\tau_k)\bY(\tau_k)$ to approximate $f(\tau)\bY(\tau)$ for $\tau\in[\tau_{k - 1}, \tau_k]$. Lemma \ref{lemma: state_discrete_error_bound} shows that this error can be controlled under the finite moment condition \eqref{eq: convergence_finite_moment}. The proof, deferred to Appendix \ref{sec: state_discrretization_error}, is based on applying It\^{o}'s formula to $f(\tau)\bY(\tau)$.

\begin{lemma}\label{lemma: state_discrete_error_bound}
    For the forward process $\bY$ in \eqref{eq: forward_SDE_DDPM}, if the moment condition \eqref{eq: convergence_finite_moment} is satisfied, then we have:
    \begin{align*}
    &\mE_{\bc\sim\bC}\bigg[\sum_{k=1}^{K} \int_{\tau_{k -1}}^{\tau_k}\mE_{\bY(\tau)\sim q_\tau(\cdot | \bc), \bY(\tau_k)\sim q_{\tau_k | \tau}(\cdot | \bY(\tau), \bc)}\big[||f(\tau)\bY(\tau) - f(\tau_k)\bY(\tau_k)||^2\big]\intd\tau\bigg] \\
    &\lesssim \Delta\tau(M_2 + d)(T_g + 1)^5.
\end{align*}
\end{lemma}

The third error term in \eqref{eq: KL_bound_decomposition} represents the time and space discretization error for approximating the score function $g^2(\tau)\nabla\log q_{\tau}(\bY(\tau)|\bc)$ by $g^2(\tau_k)\nabla\log q_{\tau_k}(\bY(\tau_k)| \bc)$ when $\tau\in[\tau_{k - 1}, \tau_k]$. We bound this error in Lemma \ref{lemma: score_discretization_error_bound}. Our proof is inspired by that of Theorem 5 of \cite{chen2023improved}, but we improve their dependence on the data dimension $d$ by applying the stochastic localization technique introduced in \cite{bentonnearly}; see Appendix \ref{sec: score_discretization_error} for the complete proof.

\begin{lemma}\label{lemma: score_discretization_error_bound}
Suppose the first condition of Proposition~\ref{prop: condition_diffusion_KL_bound} holds. Then for the forward process $\{\bY(\tau): \tau\in[0, T_g]\}$ in \eqref{eq: forward_SDE_DDPM} and the corresponding score function $\{\nabla\log q_\tau(\cdot | \bc): \tau\in[0, T_g]\}$, we have
    \begin{align*}
        &\mE_{\bc\sim\bC}\bigg[\sum_{k=1}^{K}\int_{\tau_{k - 1}}^{\tau_{k}} \mE_{\bY(\tau)\sim q_{\tau}(\cdot | \bc),\bY(\tau_k)\sim q_{\tau_k | \tau}(\cdot | \bY(\tau), \bc)}[\big\| g^2(\tau)\nabla\log q_{\tau}(\bY(\tau)|\bc) -  g^2(\tau_k)\nabla\log q_{\tau_k}(\bY(\tau_k)| \bc) \big\|]^2\intd \tau\bigg]\\
        &\lesssim M_2\Delta\tau (T_g + 1)^2 + d\Delta\tau \big[L_1 + (T_g + 1)^4 \big].
    \end{align*}
\end{lemma}

The fourth term of \eqref{eq: KL_bound_decomposition} is the discretization error of simulating the Brownian motion $g(\tau)\intd\bB(\tau)$ in \eqref{eq: reverse_SDE_DDPM} by $g(\tau_k)\sqrt{\Delta\tau}\zeta_k \overset{d}{=}g(\tau_k)\intd\bB(\tau)$ from $\tau_k$ to $\tau_{k + 1}$ in \eqref{eq: reverse_SDE_DDPM_simulation}. 
We provide a bound on this error in Lemma \ref{lemma: BM_discretization_error_bound}, whose proof is given in Appendix \ref{sec: BM_discrete_error}.

\begin{lemma}\label{lemma: BM_discretization_error_bound}
    Under the same condition as in Lemma~\ref{lemma: score_discretization_error_bound}, we have
    \begin{align*}
        &\mE_{\bc\sim\bC}\bigg[\sum_{k=1}^{K}\int_{\tau_{k - 1}}^{\tau_{k}}[g^2(\tau_k) - g^2(\tau)]^2\mE_{\bY(\tau)\sim q_{\tau}(\cdot | \bc)} \big[\big\|\nabla\log q_\tau(\bY(\tau) | \bc)\big\|^2\big] \intd\tau\bigg]\\
        & \quad \lesssim d(\Delta\tau)^2[L_1 + T_g + 1].
    \end{align*}
\end{lemma}

The last term of \eqref{eq: KL_bound_decomposition} captures the score matching error out of the neural network approximation of the true score function.
Given condition \eqref{eq: convergence_score_matching}, we have
\begin{align}
    &\Delta \tau\sum_{k=1}^{K}\mE_{\bc\sim\bC}\bigg[\mE_{\bY\sim q_{\tau_k}(\cdot | \bc)}\big[||g^2(\tau_k)\nabla\log q_{\tau_k}(\bY(\tau_k) | \bc) - g^2(\tau_k)s_\theta(\tau_k, \bY(\tau_k), \bc)||^2\big]\bigg]\nonumber\\
    &\le (T_g + 1)^3\varepsilon_{\text{score}}^2. \label{eq: score_matching_error_bound}
\end{align}

Finally, by combining the results of Lemma \ref{lemma: KL_bound_decomposition} through Lemma \ref{lemma: BM_discretization_error_bound}, along with the bound \eqref{eq: score_matching_error_bound} on the score matching error, we obtain the error bound in \eqref{eq: condition_diffusion_KL_bound}, proving Proposition \ref{prop: condition_diffusion_KL_bound}.
\end{proof}

In the following five subsections, we present the proofs of Lemmas \ref{lemma: KL_bound_decomposition} through \ref{lemma: BM_discretization_error_bound}.

\subsection{Proof of Lemma~\ref{lemma: KL_bound_decomposition}}\label{sec: error_identification}  
To prove Lemma~\ref{lemma: KL_bound_decomposition}, we first consider the reverse SDE \eqref{eq: reverse_SDE_DDPM} and the discrete simulation \eqref{eq: reverse_SDE_DDPM_simulation} when $\tau\in(\tau_k, \tau_{k + 1}]$. Given $\tilde\bY(\tau_k) = \tilde\bz\in\mR^d$, rewrite \eqref{eq: reverse_SDE_DDPM} as
\begin{align}\label{eq: piece_reverse_SDE}
    \intd\tilde{\bY}(\tau) = [\tilde f(\tau)\tilde\bY(\tau) + \tilde g^2(\tau) \nabla\log q_{T_g - \tau}(\tilde \bY(\tau) | \bc)]\intd\tau + \tilde g(\tau) \intd\tilde\bB(\tau),\ %\tilde\bY(\tau_k) = \tilde\bz,
\end{align}
where $\tilde f(\tau) := f(T_g - \tau)$ and $\tilde g(\tau) := g(T_g - \tau)$. On $(\tau_k, \tau_{k + 1}]$, write \eqref{eq: reverse_SDE_DDPM_simulation} starting from $\tilde\bZ(\tau_k) = \tilde\bz$ as
\begin{align}\label{eq: piece_reverse_SDE_simulation}
    \intd\tilde{\bZ}(\tau) = [\tilde f(\tau_k)\tilde\bZ(\tau_k) + \tilde g^2(\tau_k) s_\theta(T_g - \tau_k, \tilde\bZ(\tau_k), \bc)]\intd\tau + \tilde g(\tau_k) \intd\tilde\bB(\tau),\ %\tilde\bZ(\tau_k) = \tilde\bz,
\end{align}
where, for notational simplicity, we use the same Brownian motion $\tilde\bB$  as we only care about distributions. To facilitate presentation, we introduce additional notations summarized in Table \ref{table: KL_decomposition_notation}:
\begin{table}[htp]
    \centering
    \begin{tabular}{c c c c c}
        \toprule
        Dynamics & State & \makecell{Marginal \\ Distribution ($\tau$)} & \makecell{Conditional \\ Distribution ($\tau < s$)} & Comment \\
        \hline
        Forward SDE \eqref{eq: forward_SDE_DDPM} & $\bY(\tau)$ & $q_\tau(\cdot | \bc)$ & $q_{s | \tau}(\cdot | \bc)$ &  \makecell{$q_0(\cdot | \bc) = p_\tar(\cdot | \bc)$; \\ $q_\tau(\cdot | \bc) = \tilde p_{T_g - \tau}$}\\\midrule
        Reverse SDE \eqref{eq: piece_reverse_SDE} & $\tilde\bY(\tau)$ & $\tilde p_\tau$ & $\tilde p_{s |\tau}$ &  \makecell{omit $\bc$ for simplicity;\\ $\tilde p_{\tau} = q_{T_g - \tau}(\cdot | \bc)$}\\\midrule
        Simulated SDE \eqref{eq: piece_reverse_SDE_simulation} & $\tilde\bZ(\tau)$ & $\tilde p_\tau^\theta$ & $\tilde p^\theta_{s|\tau}$  & omit $\bc$ for simplicity\\
        \bottomrule
    \end{tabular}
    \caption{Notations for the proof of Lemma \ref{lemma: KL_bound_decomposition}.}
    \label{table: KL_decomposition_notation}
\end{table}

The key ingredient of the proof is the following Lemma \ref{lemma: KL_dt}, which characterizes the KL-divergence between conditional densities of the two It\^{o} processes, \eqref{eq: piece_reverse_SDE} and \eqref{eq: piece_reverse_SDE_simulation}, with different drift and diffusion coefficients. Note that \cite[Lemma 6, Lemma 7(2)]{chen2023improved} also provide KL-divergence characterizations, which however are for the case where two processes share the same diffusion coefficients and not applicable to our setting. For the same reason, we cannot apply the Girsanov theorem to characterize the KL-divergence between two path measures as in \cite{bentonnearly}. 
The proof of Lemma~\ref{lemma: KL_dt} is provided in Appendix \ref{sec: KL_dt_proof}.

\begin{lemma}\label{lemma: KL_dt}
Given a condition $\bc$, consider the conditional densities $\tilde p_{\tau | \tau_k}(\cdot | \tilde\bz)$ and $\tilde p^\theta_{\tau | \tau_k}(\cdot | \tilde\bz)$ induced by \eqref{eq: piece_reverse_SDE} and \eqref{eq: piece_reverse_SDE_simulation}, respectively. We have the following results.
\begin{enumerate}
    \item[(a)] If $\tilde\bY(\tau_k)$ follows the marginal density $\tilde p_{\tau_k}$ of the reverse SDE \eqref{eq: reverse_SDE_DDPM}, then
    \begin{align*}%\label{eq: limit_KL}
        \mE_{\tilde\bY(\tau_k)\sim\tilde p_{\tau_k}} \bigg[\lim_{\tau \to \tau_k+}\KL\big(\tilde p_{\tau | \tau_k}(\cdot | \tilde\bY(\tau_k)) || \tilde p^\theta_{\tau | \tau_k}(\cdot | \tilde\bY(\tau_k))\big)\bigg] = 0.
    \end{align*}

    \item[(b)] For any given initial $\tilde\bz$ and $\tau\in(\tau_k, \tau_{k + 1}]$,
    \begin{align}%\label{eq: density_KL_dt_bound}
        &\frac{\intd}{\intd \tau}\KL\big(\tilde p_{\tau | \tau_k}(\cdot | \tilde\bz) || \tilde p^\theta_{\tau | \tau_k}(\cdot | \tilde\bz)\big) \nonumber\\
        & \le \frac{1}{C_{KL}}\mE_{\tilde\bY(\tau)}\big\|\tilde f(\tau)\tilde\bY(\tau) - \tilde f(\tau_k)\tilde\bz + \tilde g^2(\tau)\nabla\log q_{T_g - \tau}(\tilde\bY(\tau)|\bc) -  \tilde g^2(\tau_k)s_\theta(T_g - \tau_k, \tilde\bz, \bc)\big\|^2 \nonumber\\
        & \qquad +\frac{[\tilde g^2(\tau_k) - \tilde g^2(\tau)]^2}{4C_{KL}}\mE_{\tilde\bY(\tau)} \big[\big\|\nabla\log\tilde p_\tau(\tilde\bY(\tau))\big\|^2\big] \nonumber\\
        &\qquad + \frac{1}{2}\bigg[C_{KL} - \tilde g^2(\tau_k)\bigg] \mE_{\tilde\bY(\tau)}\bigg\| \nabla\log\frac{\tilde p_{\tau | \tau_k}(\tilde\bY(\tau)| \tilde\bz) }{\tilde p^\theta_{\tau | \tau_k}(\tilde\bY(\tau) | \tilde\bz) }\bigg\|^2, \label{eq: density_KL_dt_bound}
    \end{align}
    for any fixed constant $C_\KL > 0$, where the expectation is taken w.r.t. the conditional density $\tilde\bY(\tau)\sim \tilde p_{\tau | \tau_k}(\cdot | \tilde\bz)$ of the reverse SDE  \eqref{eq: piece_reverse_SDE} .
\end{enumerate}
\end{lemma}

Now, we are ready to prove Lemma \ref{lemma: KL_bound_decomposition}.
\begin{proof}[Proof of Lemma \ref{lemma: KL_bound_decomposition}]

We apply the result of Lemma \ref{lemma: KL_dt}(b) and set $C_{KL} = b$. Note that $\tilde g^2(\tau) = g^2(T_g - \tau) = a(T_g - \tau) + b \ge b$ for any $\tau\in[0, T_g]$. Then, integrating on both sides of \eqref{eq: density_KL_dt_bound} from $\tau_k$ to $\tau_{k + 1}$, we have
\begin{align*}
    &\KL\big(\tilde p_{\tau_{k + 1} | \tau_k}(\cdot | \tilde\bz) || \tilde p^\theta_{\tau_{k + 1} | \tau_k}(\cdot | \tilde\bz\big) - \lim_{\tau\to\tau_k+}\KL\big(\tilde p_{\tau | \tau_k}(\cdot | \tilde\bz) || \tilde p^\theta_{\tau | \tau_k}(\cdot | \tilde\bz\big)\\
    &\le \frac{1}{b}\int_{\tau_k}^{\tau_{k + 1}} \mE_{\tilde\bY(\tau)\sim\tilde p_{\tau | \tau_k}(\cdot | \tilde\bz)} \big\|\tilde f(\tau)\tilde\bY(\tau) - \tilde f(\tau_k)\tilde\bz + \tilde g^2(\tau)\nabla\log q_{T_g - \tau}(\tilde\bY(\tau)|\bc) -  \tilde g^2(\tau_k)s_\theta(T_g - \tau_k, \tilde\bz, \bc)\big\|^2\intd \tau\\
    &\qquad + \frac{1}{4b}\int_{\tau_k}^{\tau_{k + 1}} [\tilde g^2(\tau_k) - \tilde g^2(\tau)]^2\mE_{\tilde\bY(\tau)\sim\tilde p_{\tau | \tau_k}(\cdot | \tilde\bz)} \big[\big\|\nabla\log\tilde p_\tau(\tilde\bY(\tau))\big\|^2\big] \intd\tau.
\end{align*}
Since the above result holds when $\tilde\bY(\tau_k)= \tilde\bZ(\tau_k) = \tilde \bz$ for any fixed $\tilde\bz\in\mR^d$, we can integrate on both sides of the above inequality w.r.t. $\tilde \bz = \tilde\bY(\tau_k)\sim \tilde p_{\tau_k}$. Then, by Lemma \ref{lemma: KL_dt}(a), we have
\begin{align}
        &\mE_{\tilde\bY(\tau_k)\sim \tilde p_{\tau_k}} \big[\KL\big(\tilde p_{\tau_{k + 1} | \tau_k}(\cdot | \tilde\bY(\tau_k)) || \tilde p^\theta_{\tau_{k  +1} | \tau_k}(\cdot | \tilde\bY(\tau_k)\big) \big] \nonumber\\
        &\le  \frac{1}{b}\int_{\tau_k}^{\tau_{k + 1}}\mE_{(\tilde\bY(\tau_k), \tilde\bY(\tau))}\bigg[\big\|\tilde g^2(\tau)\nabla\log q_{T_g - \tau}(\tilde\bY(\tau)|\bc) -  \tilde g^2(\tau_k)s_\theta(T_g - \tau_k, \tilde\bY(\tau_k), \bc) \nonumber\\
        &\qquad \qquad\qquad\qquad\qquad\qquad\qquad + \tilde f(\tau)\tilde\bY(\tau) - \tilde f(\tau_k) \tilde\bY(\tau_k) \big\|^2\bigg]\intd \tau\nonumber\\
        &\qquad + \frac{1}{4b}\int_{\tau_k}^{\tau_{k + 1}} [\tilde g^2(\tau_k) - \tilde g^2(\tau)]^2\mE_{\tilde\bY(\tau)\sim \tilde p_\tau} \big[\big\|\nabla\log\tilde p_\tau(\tilde\bY(\tau))\big\|^2\big] \intd\tau,\label{eq: DDPM_simulation_step_bound}
\end{align}
where the expectation in the first term on the right-hand side is taken w.r.t. $\tilde\bY(\tau_k)\sim\tilde p_{\tau_k}$ and $\tilde\bY(\tau)\sim\tilde p_{\tau|\tau_k}(\cdot | \tilde\bY(\tau_k))$.

On the other hand, consider the reverse SDE \eqref{eq: piece_reverse_SDE} starting from $\tilde\bY(\tau_k)\sim\tilde p_{\tau_k}$, and the simulated SDE \eqref{eq: piece_reverse_SDE_simulation} starting from $\tilde\bZ(\tau_k)\sim\tilde p^\theta_{t_k}$.
Using the chain rule of KL divergence, we have the following result:
\begin{equation}\label{eq: KL_chain_rule}
    \KL\big(\tilde p_{\tau_{k + 1}} || \tilde p^\theta_{\tau_{k + 1}}\big) \le \KL\big(\tilde p_{\tau_{k}} || \tilde p^\theta_{\tau_{k}}\big) + \mE_{\tilde\bY(\tau_k)\sim\tilde p_{\tau_k}}\big[\KL\big(\tilde p_{\tau_{k + 1} | \tau_k}(\cdot | \tilde\bY(\tau_k)) || \tilde p^\theta_{\tau_{k + 1} | \tau_k}(\cdot | \tilde\bY(\tau_k)\big) \big].
\end{equation}

With \eqref{eq: DDPM_simulation_step_bound} and \eqref{eq: KL_chain_rule}, we have
\begin{align*}
    &\KL(p_\tar(\cdot | \bc) || \tilde p^\theta_{\tau_K}(\cdot | \bc)) = \KL(q_0(\cdot | \bc) || \tilde p^\theta_{\tau_K}(\cdot | \bc)) = \KL\big(\tilde p_{T_g} || \tilde p^\theta_{\tau_K}\big) = \KL\big(\tilde p_{\tau_K} || \tilde p^\theta_{\tau_K}\big)\\
    &\overset{\eqref{eq: KL_chain_rule}}{\le} \KL\big(\tilde p_0 || \tilde p^\theta_0\big) + \sum_{k = 0}^{K - 1}\mE_{\tilde\bY(\tau_k)\sim\tilde p_{\tau_k}}\big[\KL\big(\tilde p_{\tau_{k + 1} | \tau_k}(\cdot | \tilde\bY(\tau_k)) || \tilde p^\theta_{\tau_{k + 1} | \tau_k}(\cdot | \tilde\bY(\tau_k)\big) \big]\\
    &\overset{\eqref{eq: DDPM_simulation_step_bound}}{\le} \KL\big(q_{T_g}(\cdot | \bc) || \mathcal{N}(0, I_d)\big) + \frac{1}{4b}\sum_{k=0}^{K - 1}\int_{\tau_k}^{\tau_{k + 1}}[\tilde g^2(\tau_k) - \tilde g^2(\tau)]^2\mE_{\tilde\bY(\tau)\sim\tilde p_{\tau}} \big[\big\|\nabla\log\tilde p_\tau(\tilde\bY(\tau))\big\|^2\big] \intd\tau\\
    &\qquad + \frac{1}{b}\sum_{k = 0}^{K - 1}\int_{\tau_k}^{\tau_{k + 1}}\mE_{\tilde\bY(\tau_k)\sim\tilde p_{\tau_k},\tilde\bY(\tau)\sim\tilde p_{\tau | \tau_k}(\cdot | \tilde\bY(\tau_k))}\bigg[\big\|\tilde g^2(\tau)\nabla\log q_{T_g - \tau}(\tilde\bY(\tau)|\bc) -  \tilde g^2(\tau_k)s_\theta(T_g - \tau_k, \tilde\bY(\tau_k), \bc) \\
    &\qquad\qquad\qquad\qquad\qquad\qquad\qquad\qquad + \tilde f(\tau)\tilde\bY(\tau) - \tilde f(\tau_k)\tilde\bY(\tau_k)\big\|^2\bigg]\intd \tau\\
    &\le \KL\big(q_{T_g}(\cdot | \bc) || \mathcal{N}(0, I_d)\big) + \frac{1}{4b}\sum_{k=1}^{K}\int_{\tau_{k - 1}}^{\tau_{k}}[g^2(\tau_k) - g^2(\tau)]^2\mE_{\bY(\tau)\sim q_{\tau}(\cdot | \bc)} \big[\big\|\nabla\log q_\tau(\bY(\tau) | \bc)\big\|^2\big] \intd\tau\\
    &\qquad + \frac{1}{b}\sum_{k = 1}^{K}\int_{\tau_{k - 1}}^{\tau_{k}}\mE_{\bY(\tau)\sim q_{\tau}(\cdot | \bc),\bY(\tau_k)\sim q_{\tau_k | \tau}(\cdot | \bY(\tau_k), \bc)}\bigg[\big\|g^2(\tau)\nabla\log q_{\tau}(\bY(\tau)|\bc) -  g^2(\tau_k)s_\theta(\tau_k, \bY(\tau_k), \bc) \\
    &\qquad\qquad\qquad\qquad\qquad\qquad\qquad\qquad + f(\tau)\bY(\tau) - f(\tau_k)\bY(\tau_k)\big\|^2\bigg]\intd \tau\\
    &\lesssim \KL\big(q_{T_g}(\cdot | \bc) || \mathcal{N}(0, I_d)\big)  + \sum_{k=1}^{K}\int_{\tau_{k - 1}}^{\tau_{k}}\mE_{\bY(\tau)\sim q_{\tau}(\cdot | \bc),\bY(\tau_k)\sim q_{\tau_k | \tau}(\cdot | \bY(\tau_k), \bc)}\big[\big\|f(\tau_k)\bY(\tau_k) - f(\tau)\bY(\tau) \big\|^2\big] \intd \tau\\
    &\qquad + \sum_{k=1}^{K}\int_{\tau_{k - 1}}^{\tau_{k}} \mE_{\bY(\tau)\sim q_{\tau}(\cdot | \bc),\bY(\tau_k)\sim q_{\tau_k | \tau}(\cdot | \bY(\tau_k), \bc)}\big[\big\| g^2(\tau)\nabla\log q_{\tau}(\bY(\tau)|\bc) -  g^2(\tau_k)\nabla\log q_{\tau_k}(\bY(\tau_k)| \bc) \big\|\big]^2\intd \tau\\
    &\qquad +  \sum_{k=1}^{K}\int_{\tau_{k - 1}}^{\tau_{k}}[g^2(\tau_k) - g^2(\tau)]^2\mE_{\bY(\tau)\sim q_{\tau}(\cdot | \bc)} \big[\big\|\nabla\log q_\tau(\bY(\tau) | \bc)\big\|^2\big] \intd\tau\\
    &\qquad + \Delta\tau\sum_{k=1}^{K}\mE_{\bY(\tau_k)\sim q_{\tau_k}(\cdot | \bc)} \big\| g^2(\tau_k)\nabla\log q_{\tau_k}(\bY(\tau_k)|\bc) -  g^2(\tau_k)s_\theta(\tau_k, \bY(\tau_k), \bc) \big\|^2,
\end{align*}
where the last step is obtained by adding and subtracting a term of $g^2(\tau_k)\nabla\log q_{\tau_k}(\bY(\tau_k) | \bc)$, and using the fact that $\|\bx + \by + \bz\|^2 \le 9\|\bx\|^2 + 9\|\by\|^2 + 9\|\bz\|^2$. The proof is hence completed.
\end{proof}

\subsection{Proof of Lemma~\ref{lemma: initial_dist_error_bound} }\label{sec: Gaussian_initialization_error}  %Error 1 - Gaussian initialization

\begin{proof}[Proof of Lemma \ref{lemma: initial_dist_error_bound}]
We adapt the proof of Proposition 4 of \cite{bentonnearly} to our setting where the forward OU process is time-inhomogeneous. Recall that $q_{\tau | 0}(\cdot | \by(0),  \bc)$ is the conditional distribution of $\bY(\tau)$ given $\bY(0) = \by(0)$ under condition $\bc$, and $q_{\tau | 0}(\cdot | \by(0),  \bc) \sim \mathcal{N}(\lambda_\tau \bY(0), \sigma_\tau^2I_d)$ from \eqref{eq: DDPM_conditional_relation} and \eqref{eq: forward_SDE_coeff}. Then, one can directly compute that
\begin{align*}
    \KL(q_{\tau | 0}(\cdot | \by(0),  \bc) || \mathcal{N}(0, I_d)) = \frac{1}{2}\big(d\log\sigma_\tau^{-2} - d + d\sigma_\tau^2 + \|\lambda_\tau\by(0)\|^2\big).
\end{align*}
By the convexity of the KL divergence and Jensen's inequality, we have
\begin{align*}
    \KL\big(q_{T_g}(\cdot | \bc) || \mathcal{N}(0, I_d)\big) &= \KL\bigg(\int_{\by(0)\in\mR^d} q_{T_g | 0}(\cdot | \by(0), \bc) \intd q_0(\by(0) | \bc) \bigg|\bigg| \mathcal{N}(0, I_d)\bigg)\\
    &\le \int_{\by(0)\in\mR^d}\KL\big(q_{T_g | 0}(\cdot | \by(0), \bc) || \mathcal{N}(0, I_d)\big)\intd q_0(\by(0) | \bc)\\
    & =\frac{1}{2}\bigg(d\log\sigma_{T_g}^{-2} - d + d\sigma_{T_g}^2 + \lambda_{T_g}^2\mE_{\bY(0)\sim q_{0}(\cdot | \bc)}\|\bY(0)\|^2\bigg)\\
    &\lesssim de^{-aT_g^2 - 2bT_g} + e^{-\frac{a}{2}T_g^2 - bT_g}\big(\mE_{\bY(0)\sim p_\tar(\cdot | \bc)}\|\bY(0)\|^2\big)
\end{align*}
where we use \eqref{eq: forward_SDE_coeff} and the approximation that $-\ln(1 - x)\lesssim x + \frac{x^2}{2}$ for $x\in(-1, 1)$. Applying condition \eqref{eq: convergence_finite_moment}, we immediately obtain
\begin{align*}
    \mE_{\bc\sim\bC}\bigg[\KL\big(q_{T_g}(\cdot | \bc) || \mathcal{N}(0, I_d)\big)\bigg] \lesssim e^{-\frac{a}{2}T_g^2 - bT_g}\big[M_2 + de^{-\frac{a}{2}T_g^2 - bT_g}\big].
\end{align*}
This completes the proof of Lemma \ref{lemma: initial_dist_error_bound}.
\end{proof}

\subsection{Proof of Lemma \ref{lemma: state_discrete_error_bound}}\label{sec: state_discrretization_error}

\begin{proof}[Proof of Lemma \ref{lemma: state_discrete_error_bound}]
By It\^{o}'s formula, with $\bY(\tau)$ following the forward SDE \eqref{eq: forward_SDE_DDPM}, we have
\begin{align*}
    \intd(f(\tau)\bY(\tau)) = f'(\tau)\bY(\tau)\intd\tau + f(\tau)\intd\bY(\tau) = \bigg[\frac{a}{2} - \frac{1}{4}(a\tau + b)^2\bigg]\bY(\tau)\intd\tau + \frac{1}{2}(a\tau + b)^{\frac{3}{2}}\intd \bB(\tau).
\end{align*}
Let $u(\tau) := \frac{a}{2} - \frac{1}{4}(a\tau + b)^2$ and $v(\tau):=\frac{1}{2}(a\tau + b)^{\frac{3}{2}}$. Then
\begin{align*}
    &\mE_{\bY(\tau)\sim q_\tau(\cdot | \bc), \bY(\tau_k)\sim q_{\tau_k | \tau}(\cdot | \bY(\tau), \bc)}\big[||f(\tau)\bY(\tau) - f(\tau_k)\bY(\tau_k)||^2\big] \\
    &= \mE_{\bY(\tau)\sim q_\tau(\cdot | \bc), (\bB)} \bigg[\bigg|\bigg|\int_{\tau}^{\tau_k}u(s)\bY(s)\intd s + \int_{\tau}^{\tau_k}v(s)\intd\bB(s)\bigg|\bigg|^2\bigg] \\
    &\lesssim \mE_{\bY(\tau)\sim q_\tau(\cdot | \bc), (\bB)} \bigg[\bigg|\bigg|\int_{\tau}^{\tau_k}u(s)\bY(s)\intd s\bigg|\bigg|^2 + \bigg|\bigg|\int_{\tau}^{\tau_k}v(s)\intd\bB(s)\bigg|\bigg|^2\bigg] \\
    &\lesssim  d\int_{\tau}^{\tau_k}v^2(s) \intd s+ \left(\int_{\tau}^{\tau_k}u^2(s)\intd s\right)\left(\int_{\tau}^{\tau_k}\mE_{\bY(s)\sim q_s(\cdot | \bc)}[||\bY(s)||^2]\intd s\right) ,
\end{align*}
where the last inequality follows from the Cauchy--Schwarz inequality. Note that  $\bY(s) | \bY(0) \sim \mathcal{N}(\lambda_s\bY(0), \sigma_s^2I_d)$ for $\bY(0)\sim p_\tar(\cdot | \bc)$. Since $\lambda^2_s, \sigma_s^2\le 1$ for any $s\in[0, T_g]$, it follows that
\begin{align*}
    \mE_{\bY(s)\sim q_s(\cdot | \bc)}[||\bY(s)||^2] \le \mE_{\bY(0)\sim p_\tar(\cdot | \bc)}[\|\bY(0)\|^2 ] + d.
\end{align*}
Meanwhile, given that $\int_{\tau}^{\tau_k}u^2(s)\intd s \lesssim \int_{\tau}^{\tau_k}(s + 1)^4\intd s \lesssim (\Delta\tau)(\tau + 1)^4$ and $\int_{\tau}^{\tau_k}v^2(s)\intd s \lesssim (\Delta\tau)(\tau + 1)^3$, we derive
\begin{align*}
    & \mE_{\bc\sim\bC}\bigg[\sum_{k=1}^{K} \int_{\tau_{k - 1}}^{\tau_k} \mE_{\bY(\tau)\sim q_\tau(\cdot | \bc), \bY(\tau_k)\sim q_{\tau_k | \tau}(\cdot | \bY(\tau), \bc)}\big[||f(\tau)\bY(\tau) - f(\tau_k)\bY(\tau_k)||^2\big]\intd\tau\bigg] \\
    &\lesssim \sum_{k=1}^{K} \int_{\tau_{k - 1}}^{\tau_k} \bigg[d(\Delta\tau)(\tau + 1)^3 +(\Delta\tau)^2 (M_2 + d)(\tau + 1)^4\bigg]\intd\tau \\
    &\lesssim \Delta\tau(M_2 + d)\int_{0}^{T_g}(\tau + 1)^4\intd \tau\\
    &\lesssim \Delta\tau(M_2 + d)(T_g + 1)^5,
\end{align*}
which completes the proof.
\end{proof}

\subsection{Proof of Lemma \ref{lemma: score_discretization_error_bound}}\label{sec: score_discretization_error}
Lemma \ref{lemma: score_discretization_error_bound} bounds the (cumulative) time and space discretization error from approximating the score function $g^2(\tau)\nabla\log q_{\tau}(\bY(\tau)|\bc)$ by $g^2(\tau_k)\nabla\log q_{\tau_k}(\bY(\tau_k)| \bc)$ for $\tau\in[\tau_{k - 1}, \tau_k]$. To establish this bound, we consider the case when $\tau$ is small and when $\tau$ is large separately.

Define an index threshold:
\begin{align}\label{eq: expoenential_threshold}
    k_1 := \min\{0\le k \le K: \tau_k \ge \min\{1, T_g/2\}\}.
\end{align}
With $k_1$, we adopt the following approximation for $\sigma_\tau^2$ throughout the proof in this subsection,
\begin{align}\label{eq: expoenetial_approximation}
    \sigma_\tau^2 = 1 - \exp(-a\tau^2/2 - b\tau) \asymp \begin{cases}
        a\tau^2/2 + b\tau,\ &\text{for } \tau < \tau_{k_1},\\
        1,& \text{otherwise}.
    \end{cases}
\end{align}

The following result bounds the discretization error when $\tau$ is small, based on the Lipschitz property of $\nabla\log q_0(\cdot | \bc)$. In particular, it can be shown that the score $\nabla\log q_\tau(\cdot | \bc)$ is still Lipschitz for small $\tau$, an important property that will be used in the proof. The proof of Lemma~\ref{lemma: lipschitz_score_diff_bound} is inspired by \cite{chen2023improved} and is deferred to Appendix \ref{sec: lipschitz_score_diff_bound}.

\begin{lemma}\label{lemma: lipschitz_score_diff_bound}
    Assume $\nabla\log p_\tar(\cdot|\bc)$ is $L(\bc)$-Lipschitz continuous for a given condition $\bc$. If for some $1\le k \le k_1$, there exists $s \in [\tau_{k - 1}, \tau_k]$ satisfying $\sigma_{s}^2\le \frac{\lambda_{s}}{2L(\bc)}$, then
    \begin{align*}
        & \int_{\tau_{k - 1}}^{s}\mE_{\bY(\tau)\sim q_\tau(\cdot | \bc), \bY(\tau_k)\sim q_{\tau_k | \tau}(\cdot | \bY(\tau), \bc)}\big[\|g^2(\tau)\nabla\log q_\tau(\bY(\tau) | \bc) - g^2(\tau_k)\nabla\log q_{\tau_k}(\bY(\tau_k) | \bc)\|^2\big]\intd \tau\\
        &\lesssim d(\Delta\tau)^2L(\bc)(L(\bc) + 1)(\tau_k + 1)^6.
    \end{align*}
\end{lemma}

However, the condition $\sigma_\tau^2 \le \lambda_\tau/(2L(\bc))$ may not hold for large $\tau$, and the score function is no longer Lipschitz continuous. To get around, we apply the stochastic localization technique introduced in \cite{bentonnearly} to control the error (which improves the technique in \citealp{chen2023improved}). In particular, we have the following result, whose proof is given in Appendix \ref{sec: stochastic_local_score_diff_bound}.

\begin{lemma}\label{lemma: score_diff_bound_large_t}
    Given a condition $\bc$, for any $1\le k_0 \le k_1$ and any $s\in(\tau_{k_0 - 1}, \tau_{k_0}]$, we have
    \begin{align*}
    &\int_{s}^{\tau_{k_0}}\mE_{\bY(\tau)\sim q_\tau(\cdot | \bc), \bY(\tau_{k_0})\sim q_{\tau_{k_0} | \tau}(\cdot | \bY(\tau), \bc)}\big[\|g^2(\tau)\nabla\log q_{\tau}(\bY(\tau) | \bc) - g^2(\tau_{k_0})\nabla\log q_{\tau_{k_0}}(\bY(\tau_{k_0}) | \bc ) \|^2\big] \intd\tau\\
    &\quad +\sum_{k=k_0 + 1}^{K}\int_{\tau_{k - 1}}^{\tau_k}\mE_{\bY(\tau)\sim q_\tau(\cdot | \bc), \bY(\tau_k)\sim q_{\tau_k | \tau}(\cdot | \bY(\tau), \bc)}\big[\|g^2(\tau)\nabla\log q_{\tau}(\bY(\tau) | \bc) - g^2(\tau_k)\nabla\log q_{\tau_k}(\bY(\tau_k) | \bc ) \|^2\big] \intd\tau\\
    &\lesssim \Delta\tau(T_g + 1)^2\mE_{\bY(0)\sim p_\tar(\cdot | \bc)}\big[\|\bY(0)\|^2\big] + d\Delta\tau\bigg[\ln\bigg(\frac{1}{s} \bigg) + \frac{1}{s} + (T_g + 1)^4 \bigg].
    \end{align*}
\end{lemma}

Now, we are ready to prove Lemma \ref{lemma: score_discretization_error_bound}.

\begin{proof}[Proof of Lemma \ref{lemma: score_discretization_error_bound}]

From \eqref{eq: forward_SDE_coeff}, we note that $\lambda_\tau\sigma_\tau^{-2}$ is decreasing in $\tau$, with  $\lim_{\tau \to 0+}\lambda_\tau\sigma_\tau^{-2} = +\infty$ and $\lim_{\tau \to \infty}\lambda_\tau\sigma_\tau^{-2} = 0$. Hence, for the given condition $\bc$, there must exist  $\tau_0$ such that $\lambda_{\tau_0}\sigma_{\tau_0}^{-2} = 2L(\bc)$. This allows us to apply Lemma \ref{lemma: lipschitz_score_diff_bound} for $\tau < \min\{\tau_0, \tau_{k_1}\}$. However, when $L(\bc)$ is small, $\tau_0$ can be large. To control the magnitude of (the sum of) $(\tau_k + 1)^6$ in Lemma \ref{lemma: lipschitz_score_diff_bound}, we introduce a larger Lipschitz constant
\begin{align}\label{eq: lipschitz_constant}
    L_2(\bc) := L(\bc) + \lambda_{\min\{1,T_g/2\}}\sigma_{\min\{1, T_g/2\}}^{-2}.
\end{align}
It is clear that $\nabla\log q_0(\cdot | \bc)$ is also $L_2(\bc)$-Lipschitz continuous. We can then define
\begin{align*}%\label{eq: lipschitz_threshold}
    \tau_{lipz}(\bc) = \max\big\{0\le \tau\le T_g: \lambda_\tau\sigma_\tau^{-2} \ge 2L_2(\bc)\big\}.
\end{align*}
Note that $\lambda_{\min\{1, T_g/2\}}\sigma_{\min\{1, T_g/2\}}^{-2} \le 2L_2(\bc)$ and $\lambda_\tau\sigma_\tau^{-2}$ is decreasing in $\tau$. Thus, for any given and fixed condition $\bc$, $\tau_{lipz}(\bc) < \min\{1, T_g/2\} \le \tau_{k_1}.$ Let $k_{0}(\bc) = \min\{1\le k \le K: \tau_k \ge \tau_{lipz}(\bc)\}\le k_1$.

We can now apply Lemma \ref{lemma: lipschitz_score_diff_bound} with settings of $s = \tau_1, \cdots, \tau_{k_{0}(\bc) - 1}$ and $s = \tau_{lipz}(\bc)$. Then, for $\bY(\tau)\sim q_{\tau}(\cdot | \bc),\bY(\tau_k)\sim q_{\tau_k | \tau}(\cdot | \bY(\tau), \bc)$,
\begin{align*}
    & \sum_{k=1}^{k_{0}(\bc) - 1}\int_{\tau_{k - 1}}^{\tau_{k}}\mE_{(\bY(\tau), \bY(\tau_k))} \big[\|g^2(\tau)\nabla\log q_\tau(\bY(\tau) | \bc) - g^2(\tau_k)\nabla\log q_{\tau_k}(\bY(\tau_k) | \bc)\|^2\big]\intd \tau \\
    & \qquad + \int_{\tau_{k_{0}(\bc) - 1}}^{\tau_{lipz}(\bc)}\mE_{(\bY(\tau), \bY(\tau_{k_0(\bc)}))} \big[\|g^2(\tau)\nabla\log q_\tau(\bY(\tau) | \bc) - g^2(\tau_{k_0(\bc)})\nabla\log q_{\tau_{k_0(\bc)}}(\bY(\tau_{k_0(\bc)}) | \bc)\|^2\big]\intd \tau\\
    &\lesssim d(\Delta\tau)^2L_2(\bc)(L_2(\bc) + 1)\bigg[\sum_{k=1}^{k_0(\bc)} (\tau_k + 1)^6\bigg]\\
    &\lesssim d(\Delta\tau)^2L_2(\bc)(L_2(\bc) + 1)k_0(\bc),
     % &\lesssim d(\Delta\tau)^2L(\bc)(L(\bc) + 1)k_0(\bc).
\end{align*}
where we use the fact that $\tau_{k_0(\bc)}\le \tau_{k_1} \le 1 + \Delta\tau$. Note that when $\tau = \tau_{lipz}(\bc)\le 1$, we can apply \eqref{eq: expoenetial_approximation} to obtain
\begin{align*}
    2L_2(\bc) \asymp \frac{\lambda_\tau}{\sigma_\tau^2} = \frac{e^{-\frac{a}{4}\tau^2 - \frac{b}{2}\tau}}{1 - e^{-\frac{a}{2}\tau^2 - b\tau}} \asymp \frac{1 - a\tau^2/4 - b\tau /2}{a\tau^2/2 + b\tau} = \frac{1}{a\tau^2/2 + b\tau} - \frac{1}{2}  \asymp \frac{1}{\tau} - \frac{1}{2},
\end{align*}
yielding
\begin{align}\label{eq: lipz_index_approximation}
    \tau_{lipz}(\bc) \asymp \frac{2}{4L_2(\bc) + 1},\quad k_0(\bc) \asymp \frac{2}{(4L_2(\bc) + 1)\Delta\tau}.
\end{align}
Therefore,
\begin{align}
    &  \mE_{\bc\sim\bC}\bigg[\sum_{k=1}^{k_{0}(\bc) - 1}\int_{\tau_{k - 1}}^{\tau_{k}}\mE_{(\bY(\tau), \bY(\tau_k))} \big[\|g^2(\tau)\nabla\log q_\tau(\bY(\tau) | \bc) - g^2(\tau_k)\nabla\log q_{\tau_k}(\bY(\tau_k) | \bc)\|^2\big]\intd \tau\bigg] \nonumber\\
    & \quad + \mE_{\bc\sim\bC}\bigg[\int_{\tau_{k_{0}(\bc) - 1}}^{\tau_{lipz}(\bc)}\mE_{(\bY(\tau), \bY(\tau_{k_0(\bc)}))} \big[\|g^2(\tau)\nabla\log q_\tau(\bY(\tau) | \bc) - g^2(\tau_{k_0(\bc)})\nabla\log q_{\tau_{k_0(\bc)}}(\bY(\tau_{k_0(\bc)}) | \bc)\|^2\big]\intd \tau\bigg] \nonumber\\
    &\lesssim \mE_{\bc\sim\bC}\big[d(\Delta\tau)^2L_2(\bc)(L_2(\bc) + 1)k_0(\bc)\big] \nonumber\\
    &\overset{\eqref{eq: lipz_index_approximation}}{\lesssim} d\Delta\tau\mE_{\bc\sim\bC}\big[L_2(\bc) + 1\big]\nonumber\\
    &=  d\Delta\tau\mE_{\bc\sim\bC}\big[L(\bc) + \lambda_{\min\{1, T_g/2\}}\sigma_{\min\{1, T_g/2\}}^{-2} + 1\big] \nonumber\\
    % &\le  d\Delta\tau\mE_{\bc\sim\bC}\big[L(\bc) + 1\big] \nonumber\\
    &\lesssim d\Delta\tau(L_1 + 1),\label{eq: score_error_bound_1}
\end{align}
where $L_1$ is defined in \eqref{eq: convergence_lipschitz}, and the last step follows from the approximation that
\begin{align}\label{eq: threshold_approximation}
    \lambda_{\min\{1, T_g/2\}}\sigma_{\min\{1, T_g/2\}}^{-2} \asymp  \frac{1}{a(\min\{1, T_g/2\})^2/2 + b\min\{1, T_g/2\}} - \frac{1}{2} \asymp 1.
\end{align}

On the other hand, for those $k \ge k_0(\bc)$, Lemma \ref{lemma: score_diff_bound_large_t} implies that
\begin{align}
    & \mE_{\bc\sim\bC}\bigg[\int_{\tau_{lipz}(\bc)}^{\tau_{k_0(\bc)}} \mE_{(\bY(\tau), \bY(\tau_{k_0(\bc)}))}\big[\|g^2(\tau)\nabla\log q_{\tau}(\bY(\tau) | \bc) - g^2(\tau_{k_0(\bc)})\nabla\log q_{\tau_{k_0(\bc)}}(\bY(\tau_{k_0(\bc)}) | \bc ) \|^2\big] \intd\tau\bigg]\nonumber\\
    & \quad + \mE_{\bc\sim\bC}\bigg[\sum_{k=k_0(\bc) + 1}^{K }\int_{\tau_{k - 1}}^{\tau_k}\mE_{(\bY(\tau), \bY(\tau_k))}\big[\|g^2(\tau)\nabla\log q_{\tau}(\bY(\tau) | \bc) - g^2(\tau_k)\nabla\log q_{\tau_k}(\bY(\tau_k) | \bc ) \|^2\big] \intd\tau\bigg]\nonumber\\
    &\lesssim \Delta\tau (T_g + 1)^2 \mE_{\bc\sim\bC} \big[\mE_{\bY(0)\sim p_\tar(\cdot | \bc)} \big[\|\bY(0)\|^2\big]\big] + d\Delta\tau\mE_{\bc\sim\bC}\bigg[\ln\bigg(\frac{1}{\tau_{lipz}(\bc)} \bigg) + \frac{1}{\tau_{lipz}(\bc)} + (T_g + 1)^4 \bigg] \nonumber\\
    &\overset{\eqref{eq: lipz_index_approximation}}{\lesssim} M_2\Delta\tau (T_g + 1)^2 + d\Delta\tau \mE_{\bc\sim\bC}\big[\ln(2L_2(\bc) + 1/2) + 2L_2(\bc) + 1/2 + (T_g + 1)^4 \big] \nonumber\\
    &\lesssim M_2\Delta\tau (T_g + 1)^2 + d\Delta\tau \big[L_1 + (T_g + 1)^4 \big],\label{eq: score_error_bound_2}
\end{align}
where in the last step we use the fact that $\ln(x + 1/2) \le x + 1/2$ for any $x\ge 0$ and the approximation in \eqref{eq: threshold_approximation}. The proof is complete by combining \eqref{eq: score_error_bound_1} and \eqref{eq: score_error_bound_2}.
\end{proof}

\subsubsection{Proof of Lemma \ref{lemma: lipschitz_score_diff_bound}}\label{sec: lipschitz_score_diff_bound}

\begin{proof}[Proof of Lemma \ref{lemma: lipschitz_score_diff_bound}]
    Given a condition $\bc$, we can compute,  for $\tau < \tau_k, \bY(\tau)\sim q_\tau(\cdot | \bc)$ and $\bY(\tau_k)\sim q_{\tau_k | \tau}(\cdot | \bY(\tau), \bc)$,
\begin{align}
    & \mE_{\bY(\tau)\sim q_\tau(\cdot | \bc), \bY(\tau_k)\sim q_{\tau_k | \tau}(\cdot | \bY(\tau), \bc)}\big[\|g^2(\tau)\nabla\log q_\tau(\bY(\tau) | \bc) - g^2(\tau_k)\nabla\log q_{\tau_k}(\bY(\tau_k) | \bc)\|^2\big]\nonumber\\
    &\le 2\mE_{\bY(\tau)}\big[\|(g^2(\tau) - g^2(\tau_k))\nabla\log q_\tau(\bY(\tau) | \bc)\|^2] \nonumber\\
    &\qquad  + 2\mE_{(\bY(\tau), \bY(\tau_k))}\big[g^4(\tau_k)\|\nabla\log q_\tau(\bY(\tau) | \bc) - \nabla\log q_{\tau_k}(\bY(\tau_k) | \bc)\|^2\big]\nonumber\\
    &\overset{\text{(i)}}{\le} 2\big[\big(g^2(\tau) - g^2(\tau_k)\big)^2 + 2g^4(\tau_k)\big(\lambda_{\tau, \tau_k}^{-1} - 1\big)^2\big] \mE_{\bY(\tau)}\big[\|\nabla\log q_\tau(\bY(\tau) | \bc)\|^2]\nonumber\\
    &\qquad  + 8\mE_{(\bY(\tau), \bY(\tau_k))}\big[g^4(\tau_k)\|\nabla\log q_\tau(\bY(\tau) | \bc) - \nabla\log q_{\tau}(\lambda_{\tau, \tau_k}^{-1}\bY(\tau_k) | \bc)\|^2\big]\nonumber\\
    &\overset{\text{(ii)}}{\lesssim} (\tau_k - \tau)^2(\tau_k + 1)^4 \mE_{\bY(\tau)}\big[\|\nabla\log q_\tau(\bY(\tau) | \bc)\|^2]\nonumber\\
    &\qquad  + (\tau_k + 1)^2\mE_{(\bY(\tau), \bY(\tau_k))}\big[\|\nabla\log q_\tau(\bY(\tau) | \bc) - \nabla\log q_{\tau}(\lambda_{\tau, \tau_k}^{-1}\bY(\tau_k) | \bc)\|^2\big] \label{eq: score_error_target}
\end{align}
where the inequality (i) is from Lemma 11 of \cite{chen2023improved}, and the inequality (ii) employs the approximation that for small $\tau_k - \tau$,
\begin{align*}
    \lambda^{-1}_{\tau, \tau_k} - 1 = e^{\frac{a}{4}(\tau_k^2 - \tau^2) + \frac{b}{2}(\tau_k - \tau)} - 1\lesssim \frac{a}{4}(\tau_k^2 - \tau^2) + \frac{b}{2}(\tau_k - \tau) \lesssim (\tau_k - \tau)(\tau_k + 1).
\end{align*}

Next, from the condition that $\sigma_{s}^2\le \frac{\lambda_{s}}{2L(\bc)}$ and the fact that $\lambda_\tau\sigma_\tau^{-2}$ is decreasing in $\tau$, we know that $\sigma_{\tau}^2\le \frac{\lambda_{\tau}}{2L(\bc)}$ for all $\tau \le s$, indicating that Lemma 14 of \cite{chen2023improved} holds and $\nabla\log q_\tau(\cdot | \bc)$ is $(2L(\bc)\lambda_\tau^{-1})$-Lipschitz on $\mR^d$ for all $\tau \le s$. Then,
\begin{align*}
    &\mE_{\bY(\tau)\sim q_\tau(\cdot | \bc), \bY(\tau_k)\sim q_{\tau_k | \tau}(\cdot | \bY(\tau), \bc)}[\|\nabla\log q_\tau(\bY(\tau) | \bc) - \nabla\log q_{\tau}(\lambda_{\tau, \tau_k}^{-1}\bY(\tau_k) | \bc)\|^2] \\
    &\le 4L^2(\bc)\lambda_\tau^{-2}\mE_{(\bY(\tau), \bY(\tau_k))} \big[\|\bY(\tau)  - \lambda_{\tau, \tau_k}^{-1} \bY(\tau_k)\|^2\big]\\
    &= 4dL^2(\bc)\lambda_\tau^{-2} \lambda_{\tau,\tau_k}^{-2} \sigma^2_{\tau, \tau_k} = 4d L^2(\bc)e^{\frac{a}{2}\tau_k^2 + b\tau_k}\big[1 - e^{-\frac{a}{2}(\tau_k^2 - \tau^2) - b(\tau_k - \tau)}\big]\\
    &\overset{\eqref{eq: expoenetial_approximation}}{\lesssim} d L^2(\bc)(1 + a\tau_k^2/2 + b\tau_k)\cdot(\tau_k - \tau)(a\tau_k + b) \\
    &\lesssim d L^2(\bc)(\tau_k - \tau)(\tau_k + 1)^3,
\end{align*}
where we use the condition that $k \le k_1$ so that we can apply \eqref{eq: expoenetial_approximation} to approximate $(\lambda_\tau \lambda_{\tau,\tau_k})^{-2}$. Next, the Lipschitz continuity of $\nabla\log q_\tau(\cdot | \bc)$ implies that $\mE_{\bY(\tau)\sim q_\tau(\cdot | \bc)}||\nabla\log q_\tau(\bY(\tau) | \bc)\|^2 \le 2dL(\bc)\lambda_\tau^{-1}$ as in Lemma 21 of \cite{chen2023improved}. Hence, we can infer from \eqref{eq: score_error_target} that
\begin{align*}
    & \int_{\tau_{k - 1}}^{s}\mE_{\bY(\tau)\sim q_\tau(\cdot | \bc), \bY(\tau_k)\sim q_{\tau_k | \tau}(\cdot | \bY(\tau), \bc)}\big[\|g^2(\tau)\nabla\log q_\tau(\bY(\tau) | \bc) - g^2(\tau_k)\nabla\log q_{\tau_k}(\bY(\tau_k) | \bc)\|^2\big]\intd \tau \\
    &\lesssim \int_{\tau_{k - 1}}^{s}(\tau_k - \tau)^2(\tau_k + 1)^4 dL(\bc)\lambda_\tau^{-1} + d L^2(\bc)(\tau_k - \tau)(\tau_k + 1)^5 \intd \tau\\
    &\lesssim \int_{\tau_{k - 1}}^{s}(\tau_k - \tau)^2(\tau_k + 1)^6 dL(\bc) + d L^2(\bc)(\tau_k - \tau)(\tau_k + 1)^5 \intd \tau\\
    &\lesssim d(\Delta\tau)^2L(\bc)(L(\bc) + 1)(\tau_k + 1)^6,
\end{align*}
where the second inequality is obtained by applying \eqref{eq: expoenetial_approximation} to approximate $\lambda_\tau^{-1}$. The proof is complete.
\end{proof}

\subsubsection{Proof of Lemma \ref{lemma: score_diff_bound_large_t}}\label{sec: stochastic_local_score_diff_bound}

The following Lemma \ref{lemma: score_function_norm_bound} follows immediately from Lemmas 20 and 21 in \cite{chen2023improved}.

\begin{lemma}\label{lemma: score_function_norm_bound}
    For any given condition $\bc$ and $\tau \in (0, T_g]$, and the marginal density $q_\tau(\cdot | \bc)$ defined by the forward process \eqref{eq: forward_SDE_DDPM}, we have
    \begin{align*}
        \mE_{\bY(\tau)\sim q_\tau(\cdot | \bc)}\|\nabla\log q_\tau(\bY(\tau) | \bc)\|^2 \le \frac{d}{\sigma_\tau^2},
    \end{align*}
    where $\sigma_\tau^2$ is the conditional variance defined in \eqref{eq: forward_SDE_coeff}.
\end{lemma}

We next apply the stochastic localization technique in \cite{bentonnearly} (see Lemmas~3 to 6 therein) to derive a bound on the space and time discretization error of the score function in Lemma \ref{lemma: stochastic_char_score_diff} below. This result is the key to the proof of Lemma \ref{lemma: score_diff_bound_large_t}, whose proof is placed at the end of this subsection due to its length.

\begin{lemma}\label{lemma: stochastic_char_score_diff}
    For any given condition $\bc$ and $\tau \in (0, T_g]$, let $q_{0|\tau}(\cdot | \by(\tau), \bc)$ be the conditional distribution of $\bY(0)$ given $\bc$ and $\bY(\tau)=\by(\tau)$, and $\bSimga_\tau(\by(\tau), \bc) = \cov_{\bY(0)\sim q_{0|\tau}(\cdot | \by(\tau), \bc)}\big(\bY(0)\big)$ be the corresponding conditional covariance matrix. Then for any $\tau > 0$, when $\tau\in[\tau_{k - 1}, \tau_k]$, we have
    \begin{align*}
    &\mE_{\bY(\tau)\sim q_\tau(\cdot | \bc), \bY(\tau_k)\sim q_{\tau_k | \tau}(\cdot | \bY(\tau), \bc)}\big[\|\nabla\log q_{\tau}(\bY(\tau) | \bc) - \nabla\log q_{\tau_k}(\bY(\tau_k) | \bc ) \|^2\big]\\
    &\le \int_{\tau}^{\tau_k} f(s)d\sigma^{-2}_{\tau_k} + g^2(s)d\sigma^{-4}_{s} \intd s\\
    &\qquad + \sigma_{\tau_k}^{-4}\mE_{\bY(\tau_k)\sim q_{\tau_k}(\cdot | \bc)}[\text{trace}(\bSimga_{\tau_k}(\bY(\tau_k), \bc))] - \sigma_{\tau}^{-4}\mE_{\bY(\tau)\sim q_\tau(\cdot | \bc)}[\text{trace}(\bSimga_{\tau}(\bY(\tau), \bc))].
\end{align*}
\end{lemma}

Now, we prove Lemma \ref{lemma: score_diff_bound_large_t}.

\begin{proof}[Proof of Lemma \ref{lemma: score_diff_bound_large_t}]
For notational simplicity, we introduce
\begin{align*}
    \tau'_{k_0 - 1} = s,\ \tau'_{k} = \tau_k,\ \text{for }k=k_0, \cdots, K,
\end{align*}
which means that $\tau_{k}' - \tau_{k - 1}' \le \Delta\tau$ for any $k=k_0, \cdots, K$.

Now, given the condition $\bc$, note that for $\bY(\tau)\sim q_\tau(\cdot | \bc)$ and $\bY(\tau_k')\sim q_{\tau_k' | \tau}(\cdot | \bY(\tau), \bc)$,
\begin{align*}
    & \mE_{(\bY(\tau), \bY(\tau_k'))} \big[\|g^2(\tau)\nabla\log q_\tau(\bY(\tau) | \bc) - g^2(\tau_k')\nabla\log q_{\tau_k'}(\bY(\tau_k') | \bc)\|^2\big]\\
    &\le 2\mE_{\bY(\tau)}\big[\|(g^2(\tau) - g^2(\tau_k'))\nabla\log q_\tau(\bY(\tau) | \bc)\|^2]\\
    &\qquad  + 2\mE_{(\bY(\tau), \bY(\tau_k'))} \big[g^4(\tau_k')\|\nabla\log q_\tau(\bY(\tau) | \bc) - \nabla\log q_{\tau_k'}(\bY(\tau_k') | \bc)\|^2\big]\\
    &\lesssim d(\Delta\tau)^2\sigma_\tau^{-2} +  \mE_{(\bY(\tau), \bY(\tau_k'))} \big[g^4(\tau_k')\|\nabla\log q_\tau(\bY(\tau) | \bc) - \nabla\log q_{\tau_k'}(\bY(\tau_k') | \bc)\|^2\big],
\end{align*}
where the last inequality is from Lemma \ref{lemma: score_function_norm_bound}. Then, since $\tau_k' = \tau_k$ for $k\ge k_0$, by Lemma \ref{lemma: stochastic_char_score_diff} we have
\begin{align}
    & \int_{s}^{\tau_{k_0}}\mE_{\bY(\tau)\sim q_\tau(\cdot | \bc), \bY(\tau_{k_0})\sim q_{\tau_{k_0} | \tau}(\cdot | \bY(\tau), \bc)}\big[\|g^2(\tau)\nabla\log q_{\tau}(\bY(\tau) | \bc) - g^2(\tau_{k_0})\nabla\log q_{\tau_{k_0}}(\bY(\tau_{k_0}) | \bc ) \|^2\big] \intd\tau\nonumber\\
    &\qquad + \sum_{k=k_0 + 1}^{K}\int_{\tau_{k - 1}}^{\tau_k}\mE_{(\bY(\tau), \bY(\tau_k))}\big[\|g^2(\tau)\nabla\log q_\tau(\bY(\tau) | \bc) - g^2(\tau_k)\nabla\log q_{\tau_k}(\bY(\tau_k) | \bc)\|^2\big]\intd \tau \nonumber\\
    &=\sum_{k=k_0}^{K}\int_{\tau_{k - 1}'}^{\tau_k'}\mE_{(\bY(\tau), \bY(\tau_k'))}\big[\|g^2(\tau)\nabla\log q_\tau(\bY(\tau) | \bc) - g^2(\tau_k')\nabla\log q_{\tau_k'}(\bY(\tau_k') | \bc)\|^2\big]\intd \tau \nonumber\\
    & \lesssim \sum_{k=k_0}^{K} \int_{\tau_{k -1}'}^{\tau_k'} \bigg[d(\Delta\tau)^2\sigma_\tau^{-2}  + \int_{\tau}^{\tau_k'}g^4(\tau_k')f(s)d\sigma^{-2}_{\tau_k'} + g^4(\tau_k')g^2(s)d\sigma^{-4}_{s} \intd s \bigg]\intd \tau \label{eq: score_diff_part_1}\\
    &\qquad  + \sum_{k=k_0}^{K}\int_{\tau_{k - 1}'}^{\tau_k'} g^4(\tau_k')\big(\sigma_{\tau_k'}^{-4}\mE_{\bY(\tau_k')}[\text{trace}(\bSimga_{\tau_k'})] - \sigma_{\tau}^{-4}\mE_{\bY(\tau)}[\text{trace}(\bSimga_{\tau})]\big) \intd\tau, \label{eq: score_diff_part_2}
\end{align}
where we omit the dependency of $\bSimga_\tau$ on $\bY(\tau)$ and $\bc$ for simplicity since the context is clear.

We first bound \eqref{eq: score_diff_part_1}. Recall that $f(\tau) = \frac{1}{2}(a\tau + b), g(\tau) = \sqrt{a\tau + b}$. In addition, $\sigma_\tau^{-2}$ is decreasing in $\tau$ from \eqref{eq: forward_SDE_coeff} and we have its approximation based on \eqref{eq: expoenetial_approximation}. Then
\begin{align}
    &\sum_{k=k_0}^{K} \int_{\tau_{k -1}'}^{\tau_k'} \bigg[d(\Delta\tau)^2\sigma_\tau^{-2}  + \int_{\tau}^{\tau_k'}g^4(\tau_k')f(s)d\sigma^{-2}_{\tau_k'} + g^4(\tau_k')g^2(s)d\sigma^{-4}_{s} \intd s \bigg]\intd \tau \nonumber\\
    &\lesssim \sum_{k=k_0}^{K}\int_{\tau_{k - 1}'}^{\tau_k'} \bigg[d(\Delta\tau)^2\sigma_{\tau_{k - 1}'}^{-2} + \int_{\tau}^{\tau_k'} g^4(\tau_k')f(\tau_k')d\sigma^{-2}_{\tau_{k - 1}'} + g^4(\tau_k')g^2(\tau_k')d\sigma^{-4}_{\tau_{k - 1}'} \intd s \bigg]\intd \tau \nonumber\\
    &\lesssim \sum_{k=k_0}^{k_1} \bigg[\frac{d(\Delta\tau)^3}{a(\tau_{k - 1}')^2/2 + b\tau_{k - 1}'}+ \frac{d(\Delta\tau)^2(a\tau_k' + b)^3}{a(\tau_{k - 1}')^2/2 + b\tau_{k - 1}'} + \frac{d(\Delta\tau)^2(a\tau_k' + b)^3}{[a(\tau_{k - 1}')^2/2 + b\tau_{k - 1}']^2}\bigg] + \sum_{k_1 + 1}^{K} d(\Delta\tau)^2(a\tau_k' + b)^3\nonumber\\
    &\lesssim d\Delta\tau\int_{\tau_{k_0 - 1}'}^{\tau_{k_1}'}  \bigg[1 + \tau + \frac{1}{\tau} + \frac{1}{\tau^2}\bigg] \intd\tau + d\Delta\tau\int_{\tau_{k_1}'}^{T_g}  (\tau + 1)^3 \intd\tau\nonumber\\
    &\lesssim d\Delta\tau\bigg[\ln\bigg(\frac{1}{\tau_{k_0 - 1}'} \bigg) + \frac{1}{\tau_{k_0 - 1}'} + (T_g + 1)^4 \bigg]\nonumber\\
    &= d\Delta\tau\bigg[\ln\bigg(\frac{1}{s} \bigg) + \frac{1}{s} + (T_g + 1)^4 \bigg].\label{eq: score_diff_bound_1}
\end{align}

We next bound \eqref{eq: score_diff_part_2}. Noting that both $\sigma_\tau^2$ and $\mE_{\bY(\tau)}[\text{trace}(\bSimga_{\tau})]$ are increasing in $\tau$ (see \eqref{eq: posterior_cov_trace_dt} and the discussion thereafter), we  rearrange the summation to have
\begin{align}
    &\sum_{k=k_0}^{K}\int_{\tau_{k - 1}'}^{\tau_k'}g^4(\tau_k') \big(\sigma_{\tau_k'}^{-4}\mE_{\bY(\tau_k')}[\text{trace}(\bSimga_{\tau_k}')] - \sigma_{\tau}^{-4}\mE_{\bY(\tau)}[\text{trace}(\bSimga_{\tau})]\big)\intd\tau \nonumber\\
    &\le \sum_{k=k_0}^{K}g^4(\tau_k')\Delta\tau \big(\sigma_{\tau_k'}^{-4}\mE_{\bY(\tau_k')}[\text{trace}(\bSimga_{\tau_k'})] - \sigma_{\tau_k'}^{-4}\mE_{\bY(\tau_{k - 1}')}[\text{trace}(\bSimga_{\tau_{k - 1}'})]\big)\nonumber\\
    &\lesssim  \sum_{k=k_0 - 1}^{k_1} \Delta\tau\mE_{\bY(\tau_k')}[\text{trace}(\bSimga_{\tau_k'})] \big(g^4(\tau_k')\sigma_{\tau_k'}^{-4} - g^4(\tau_{k + 1}')\sigma_{\tau_{k + 1}'}^{-4}\big) \nonumber\\
    &\qquad + g^4(T_g)\sigma_{\tau_{k_1}'}^{-4}\Delta\tau \mE_{\bY(\tau_{k_1 + 1}')}[\text{trace}(\bSimga_{\tau_{k_1 + 1}'})]  \nonumber\\
    &\qquad + g^4(T_g)\sigma_{\tau_{k_1}'}^{-4}\Delta\tau \bigg[\sum_{k=k_1 + 2}^{K}\big(\mE_{\bY(\tau_{k}')}[\text{trace}(\bSimga_{\tau_{k}'})] - \mE_{\bY(\tau_{k - 1}')}[\text{trace}(\bSimga_{\tau_{k - 1}'})]\big)\bigg] \nonumber\\
    &\overset{\eqref{eq: expoenetial_approximation}}{\lesssim} \Delta\tau(T_g + 1)^2\mE_{\bY(T_g)}[\text{trace}(\bSimga_{T_g})] \nonumber\\
    &\qquad +\sum_{k=k_0 - 1}^{k_1} \Delta\tau\mE_{\bY(\tau_k')}[\text{trace}(\bSimga_{\tau_k'})]\bigg[\frac{(a\tau_k' + b)^2}{[a(\tau_k')^2/2 + b\tau_k']^{2}} - \frac{(a\tau_{k + 1}' + b)^2}{[a(\tau_{k + 1}')^2/2 + b\tau_{k + 1}']^{2}}\bigg] \nonumber\\
    &\lesssim \Delta\tau(T_g + 1)^2\mE_{\bY(T_g)}[\text{trace}(\bSimga_{T_g})] + \sum_{k=k_0 - 1}^{k_1} \Delta\tau\mE_{\bY(\tau_k')}[\text{trace}(\bSimga_{\tau_k'})]\cdot\bigg[\frac{1}{(\tau_k')^2} - \frac{1}{(\tau_{k + 1}')^2}\bigg] \nonumber\\
    &\lesssim \Delta\tau(T_g + 1)^2\mE_{\bY(T_g)}[\text{trace}(\bSimga_{T_g})] + \sum_{k=k_0 - 1}^{k_1} \frac{(\Delta\tau)^2}{(\tau_k')^3}\cdot\mE_{\bY(\tau_k')}[\text{trace}(\bSimga_{\tau_k'})]. \label{eq: posterior_trace_diff}
\end{align}
Recall that $\bSimga_{\tau} = \bSimga_{\tau}(\bY(\tau), \bc)$ is the conditional variance of $\bY(0)$ given $\bY(\tau)$ and $\bc$. Then
\begin{align}\label{eq: posterior_trace_bound_1}
    \mE_{\bY(T_g)}[\text{trace}(\bSimga_{T_g})] \le \mE_{\bY(T_g)}\big[\mE_{\bY(0)|\bY(T_g)}\|\bY(0)\|^2\big] = \mE_{\bY(0)\sim p_\tar(\cdot | \bc)}\big[\|\bY(0)\|^2\big].
\end{align}
On the other hand, $\bY(\tau) | \bY(0) \sim \mathcal{N}\big(\lambda_\tau\bY(0), \sigma_\tau^2I_d\big)$ from \eqref{eq: DDPM_conditional_relation}. Then, for $\tau < \tau_{k_1}' = \tau_{k_1}$ in \eqref{eq: expoenential_threshold}, using a similar approximation as in \eqref{eq: expoenetial_approximation} we have
\begin{align}
    \mE_{\bY(\tau)}[\text{trace}(\bSimga_{\tau})] &= \mE_{\bY(\tau)}\big[\text{trace}\big(\cov_{\bY(0)|\bY(\tau)}[\bY(0)]\big)\big]\nonumber\\
    &=\lambda_{\tau}^{-2}\mE_{\bY(\tau)} \big[\text{trace}\big(\cov_{\bY(0)|\bY(\tau)}[\lambda_{\tau}\bY(0) - \bY(\tau)]\big)\big] \nonumber\\
    &\le \lambda_{\tau}^{-2}\mE_{(\bY(0), \bY(\tau))} \big[\|\lambda_{\tau}\bY(0) - \bY(\tau)\|^2\big] \nonumber\\
    &= \lambda_\tau^{-2}d\sigma_\tau^2 = d\big(e^{a\tau^2/2 + b\tau} - 1\big)\nonumber\\
    &\asymp d(a\tau^2/2 + b\tau). \label{eq: posterior_trace_bound_2}
\end{align}
Substituting \eqref{eq: posterior_trace_bound_1} and \eqref{eq: posterior_trace_bound_2} into \eqref{eq: posterior_trace_diff}, we have
\begin{align}
    &\sum_{k=k_0}^{K}\int_{\tau_{k - 1}'}^{\tau_k'}g^4(\tau_k') \big(\sigma_{\tau_k'}^{-4}\mE_{\bY(\tau_k')}[\text{trace}(\bSimga_{\tau_k'})] - \sigma_{\tau}^{-4}\mE_{\bY(\tau)}[\text{trace}(\bSimga_{\tau})]\big)\intd\tau \nonumber\\
    &\lesssim  \Delta\tau(T_g + 1)^2 \mE_{\bY(0)}\big[\|\bY(0)\|^2\big] + \sum_{k=k_0 - 1}^{k_1} \frac{(\Delta\tau)^2d[a(\tau_k')^2/2 + b\tau_k'
    ]}{(\tau_k')^3} \nonumber \\
    &\lesssim \Delta\tau(T_g + 1)^2 \mE_{\bY(0)}\big[\|\bY(0)\|^2\big]  + d\Delta\tau  \int_{\tau_{k_0 - 1}'}^{\tau_{k_1}'}\bigg[\frac{1}{\tau} + \frac{1}{\tau^2}\bigg] \intd\tau \nonumber\\
    &\lesssim \Delta\tau(T_g + 1)^2\mE_{\bY(0)}\big[\|\bY(0)\|^2\big]  + d\Delta\tau\bigg[\ln\bigg(\frac{1}{\tau_{k_0 - 1}'} \bigg) + \frac{1}{\tau_{k_0 - 1}'}\bigg]\nonumber\\
    &= \Delta\tau(T_g + 1)^2\mE_{\bY(0)\sim p_\tar(\cdot | \bc)}\big[\|\bY(0)\|^2\big]  + d\Delta\tau\bigg[\ln\bigg(\frac{1}{s} \bigg) + \frac{1}{s}\bigg]. \label{eq: score_diff_bound_2}
\end{align}
The proof is complete by combining \eqref{eq: score_diff_bound_1} and \eqref{eq: score_diff_bound_2}.
\end{proof}

It remains to prove Lemma \ref{lemma: stochastic_char_score_diff}. To this end, we need a bound on the expectation of $\|\nabla^2\log q_\tau(\bY_\tau|\bc)\|_F^2$, which is provided in Lemma \ref{lemma: hessian_score_bound} below. Its proof, deferred to Appendix \ref{sec: stochastic_localization}, is based on the stochastic localization technique.

Recall that $q_{0|\tau}(\cdot | \by(\tau), \bc)$ is the conditional distribution of $\bY(0)$ given $\bc$ and $\bY(\tau)=\by(\tau)$, and $\bSimga_\tau(\by(\tau),\bc) = \cov_{\bY(0)\sim q_{0|\tau}(\cdot | \by(\tau), \bc)} \big(\bY(0) | \by(\tau)\big)$  %\textcolor{red}{missing $c$?}
    is the corresponding conditional covariance matrix.
\begin{lemma}\label{lemma: hessian_score_bound}
    For any given condition $\bc$ and any $\tau \in (0, T_g]$, we have
    \begin{align*}
        \mE_{\bY(\tau)\sim q_\tau(\cdot | \bc)}\big[\|\nabla^2\log q_\tau(\bY_\tau|\bc)\|_F^2\big]\le d\sigma_\tau^{-4} + \frac{1}{g^2(\tau)}\cdot \frac{\intd}{\intd\tau}\bigg(\sigma_\tau^{-4}\mE_{\bY(\tau)\sim q_\tau(\cdot | \bc)}[\text{trace}(\bSimga_\tau(\bY(\tau), \bc))]\bigg).
    \end{align*}
    \end{lemma}

\begin{proof}[Proof of Lemma \ref{lemma: stochastic_char_score_diff}]
Following \cite{bentonnearly}, we first consider the stochastic dynamic of the score function $\big\{\nabla\log q_{T_g - \tau}(\tilde{\bY}(\tau) | \bc ): \tau\in[0, T_g)\big\}$ under a given condition $\bc$.  Recall that the reverse SDE \eqref{eq: reverse_SDE_DDPM} of diffusion models is
\begin{align*}
     \intd\tilde{\bY}(\tau) = [\tilde f(\tau)\tilde\bY(\tau) + \tilde g^2(\tau) \nabla\log q_{T_g - \tau}(\tilde \bY(\tau) | \bc)]\intd\tau + \tilde g(\tau) \intd\tilde\bB(\tau),
\end{align*}
where $\tilde f(\tau) := f(T_g - \tau)$ and $\tilde g(\tau) := g(T_g - \tau)$. 
As in the proof of Lemma 3 of \cite{bentonnearly}, $\nabla\log q_{T_g - \tau}(\cdot | \bc)$ is smooth for $\tau\in[0, T_g)$, and by It\^{o}'s formula we can write
\begin{align}
        \intd\big[\nabla\log q_{T_g - \tau}(\tilde{\bY}(\tau) | \bc )\big] &= \bigg\{\nabla^2\log q_{T_g - \tau}(\tilde{\bY}(\tau) | \bc )\cdot\big[\tilde f(\tau)\tilde\bY(\tau) + \tilde g^2(\tau) \nabla\log q_{T_g - \tau}(\tilde \bY(\tau) | \bc)\big]\nonumber\\
        &\qquad \quad + \frac{1}{2}\tilde g^2(\tau)\cdot \Delta \big(\nabla\log q_{T_g - \tau}(\tilde{\bY}(\tau) | \bc )\big)\bigg\}\intd\tau + \frac{\intd\big[\nabla\log q_{T_g - \tau}(\tilde{\bY}(\tau) | \bc )\big]}{\intd \tau} \intd \tau \nonumber\\
        &\qquad +  \tilde g(\tau)\nabla^2\log q_{T_g - \tau}(\tilde{\bY}(\tau) | \bc ) \intd\tilde\bB(\tau).\label{eq: score_function_ito}
\end{align}
To further simplify \eqref{eq: score_function_ito}, consider the Fokker–Planck equation for the forward process \eqref{eq: forward_SDE_DDPM}:
\begin{align*}
    \intd q_\tau(\by | \bc) = \bigg(-\nabla\cdot\big[(-f(\tau)q_\tau(\by | \bc)\by \big] + \frac{1}{2}\Delta\big[g^2(\tau)q_\tau(\by|\bc)\big]\bigg)\intd\tau,
\end{align*}
leading to
\begin{align*}
    \intd(\log q_\tau(\by | \bc)) &= \frac{1}{q_\tau(\by | \bc)}\bigg(-\nabla\cdot\big[(-f(\tau) q_\tau(\by | \bc)\by\big] + \frac{1}{2} \Delta\big[g^2(\tau)q_\tau(\by|\bc)\big]\bigg)\intd\tau\\
    &= \bigg(f(\tau)d + f(\tau)\by^T\nabla\log q_\tau(\by | \bc) + \frac{1}{2}g^2(\tau)\Delta\log  q_\tau(\by | \bc) + \frac{1}{2}g^2(\tau)\|\nabla\log q_\tau(\by | \bc)\|^2\bigg)\intd \tau.
\end{align*}
It follows that
\begin{equation}\label{eq: score_function_time_derivative}
    \begin{split}
        &\frac{\intd\big[\nabla\log q_{T_g - \tau}(\by | \bc )\big]}{\intd \tau} = \nabla\frac{\intd(\log q_{T_g - \tau}(\by | \bc))}{\intd \tau} \\
        &= -\bigg(\tilde f(\tau)\nabla\log q_{T_g - \tau}(\by | \bc) + \tilde f(\tau)\nabla^2\log q_{T_g - \tau}(\by | \bc)\cdot\by \\
        & \quad \qquad + \frac{1}{2}\tilde g^2(\tau)\nabla\big(\Delta\log  q_{T_g - \tau}(\by | \bc)\big) + \tilde g^2(\tau)\nabla^2\log q_{T_g - \tau}(\by | \bc)\cdot \nabla\log q_{T_g - \tau}(\by | \bc)\bigg).
    \end{split}
\end{equation}
Substituting \eqref{eq: score_function_time_derivative} into \eqref{eq: score_function_ito} and simplifying, we obtain
\begin{align}\label{eq: score_function_process}
    \intd\big[\nabla\log q_{T_g - \tau}(\tilde{\bY}(\tau) | \bc )\big] = -\tilde f(\tau)\nabla\log q_{T_g - \tau}(\tilde\bY(\tau) | \bc) \intd \tau + \tilde g(\tau)\nabla^2\log q_{T_g - \tau}(\tilde{\bY}(\tau) | \bc ) \intd\tilde\bB(\tau).
\end{align}

Now, we establish the desired bound. First, for  fixed $s \in (0, \tau)$, we have
\begin{align*}
    &\intd\big[(\nabla\log q_{T_g - s}(\tilde{\bY}(s) | \bc ))^T\nabla\log q_{T_g - \tau}(\tilde{\bY}(\tau) | \bc )\big] \\
    &= -\tilde f(\tau)(\nabla\log q_{T_g - s}(\tilde{\bY}(s) | \bc ))^T\nabla\log q_{T_g - \tau}(\tilde\bY(\tau) | \bc) \intd \tau \\
    &\qquad + \tilde g(\tau)(\nabla\log q_{T_g - s}(\tilde{\bY}(s) | \bc ))^T\nabla^2\log q_{T_g - \tau}(\tilde{\bY}(\tau) | \bc ) \intd\tilde\bB(\tau).
\end{align*}
By Lemma \ref{lemma: hessian_score_bound}, $\int_{s}^\tau \mE_{\tilde\bY(u)}\big[\|\nabla^2\log q_{T_g - u}(\tilde\bY(u))\|^2_F\big]\intd u < \infty$ where $\tilde\bY(u)\sim q_{T_g - u}(\cdot | \bc)$, implying that the final term in the above equation is a square-integrable martingale. Integrating on both sides and taking expectations w.r.t. the joint distribution of $(\tilde\bY(\tau), \tilde\bY(s))$ induced by  \eqref{eq: reverse_SDE_DDPM}, we get
\begin{equation}\label{eq: score_process_corss_term_dt}
    \begin{split}
        &\frac{\intd}{\intd\tau}\mE_{(\tilde\bY(\tau), \tilde\bY(s))}\big[(\nabla\log q_{T_g - s}(\tilde{\bY}(s) | \bc ))^T\nabla\log q_{T_g - \tau}(\tilde{\bY}(\tau) | \bc )\big]\\
        &= -\mE_{(\tilde\bY(\tau), \tilde\bY(s))}\big[\tilde f(\tau)(\nabla\log q_{T_g - s}(\tilde{\bY}(s) | \bc ))^T\nabla\log q_{T_g - \tau}(\tilde\bY(\tau) | \bc) \big].
    \end{split}
\end{equation}

Next, if setting $\tilde\Lambda(\tau) = \exp\big(\int_0^\tau \tilde f(s)\intd s \big)$ and  applying It\^{o}'s formula to \eqref{eq: score_function_process}, we obtain
\begin{align*}
    \intd\big[\tilde\Lambda(\tau)\nabla\log q_{T_g - \tau}(\tilde{\bY}(\tau) | \bc )\big] = \tilde\Lambda(\tau)\tilde g(\tau)\nabla^2\log q_{T_g - \tau}(\tilde{\bY}(\tau) | \bc ) \intd\tilde\bB(\tau).
\end{align*}
We know from Lemma \ref{lemma: hessian_score_bound} that $\int_{s}^\tau\tilde\Lambda^2(u)\mE_{\tilde\bY(u)} \big[\|\nabla^2\log q_{T_g - u}(\tilde\bY(u))\|^2_F\big]\intd u < \infty$ for any fixed $s > 0$. By It\^{o}'s isometry and chain rule, we obtain
\begin{align*}
    &\tilde\Lambda^2(\tau)\tilde g^2(\tau)\mE_{\tilde\bY(\tau)}\big[\|\nabla^2\log q_{T_g - \tau}(\tilde\bY(\tau) | \bc)\|^2_F\big]\\
    &= \frac{\intd}{\intd\tau}\mE_{(\tilde\bY(\tau), \tilde\bY(s))}\big[\|\tilde\Lambda(\tau)\nabla\log q_{T_g - \tau}(\tilde{\bY}(\tau) | \bc ) - \tilde\Lambda(s)\nabla\log q_{T_g - s}(\tilde{\bY}(s) | \bc )\|^2\big] \nonumber\\
    &= \tilde\Lambda^2(\tau)\cdot \frac{\intd}{\intd\tau}\mE_{\tilde\bY(\tau)}\big[\|\nabla\log q_{T_g - \tau}(\tilde{\bY}(\tau) | \bc ) ||^2 + 2\tilde f(\tau)\tilde\Lambda^2(\tau) \mE_{\tilde\bY(\tau)}\big[\|\nabla\log q_{T_g - \tau}(\tilde{\bY}(\tau) | \bc ) ||^2\\
    &\qquad - 2\tilde\Lambda(\tau)\tilde\Lambda(s)\cdot \frac{\intd}{\intd\tau}\mE_{(\tilde\bY(\tau), \tilde\bY(s))}\big[(\nabla\log q_{T_g - \tau}(\tilde{\bY}(\tau) | \bc ))^T\nabla\log q_{T_g - s}(\tilde{\bY}(s) | \bc )\big] \\
    &\qquad - 2\tilde f(\tau)\tilde\Lambda(\tau)\tilde\Lambda(s)\mE_{(\tilde\bY(\tau), \tilde\bY(s))}\big[(\nabla\log q_{T_g - \tau}(\tilde{\bY}(\tau) | \bc ))^T\nabla\log q_{T_g - s}(\tilde{\bY}(s) | \bc )\big]\\
    &\overset{\eqref{eq: score_process_corss_term_dt}}{=}  \tilde\Lambda^2(\tau)\cdot \frac{\intd}{\intd\tau}\mE_{\tilde\bY(\tau)}\big[\|\nabla\log q_{T_g - \tau}(\tilde{\bY}(\tau) | \bc ) ||^2 + 2\tilde f(\tau)\tilde\Lambda^2(\tau)\mE_{\tilde\bY(\tau)}\big[\|\nabla\log q_{T_g - \tau}(\tilde{\bY}(\tau) | \bc ) ||^2.
\end{align*}
It follows that
\begin{equation}\label{eq: F_norm_characterization}
    \begin{split}
        & \frac{\intd}{\intd\tau}\mE_{\tilde\bY(\tau)}\big[\|\nabla\log q_{T_g - \tau}(\tilde{\bY}(\tau) | \bc ) ||^2 + 2\tilde f(\tau)\mE_{\tilde\bY(\tau)}\big[\|\nabla\log q_{T_g - \tau}(\tilde{\bY}(\tau) | \bc ) ||^2\\
        &= \tilde g^2(\tau)\mE_{\tilde\bY(\tau)}\big[\|\nabla^2\log q_{T_g - \tau}(\tilde\bY(\tau) | \bc)\|^2_F\big].
    \end{split}
\end{equation}

For any fixed $s > 0$ and $\tau > s$,  \eqref{eq: score_process_corss_term_dt} and \eqref{eq: F_norm_characterization} lead to
\begin{align*}
    &\frac{\intd}{\intd\tau}\mE_{(\tilde\bY(\tau), \tilde\bY(s))} \big[\|\nabla\log q_{T_g - \tau}(\tilde{\bY}(\tau) | \bc ) - \nabla\log q_{T_g - s}(\tilde{\bY}(s) | \bc )\|^2\big]\\
    &= \frac{\intd}{\intd\tau}\mE_{\tilde\bY(\tau)}\big[\|\nabla\log q_{T_g - \tau}(\tilde{\bY}(\tau) | \bc )\|^2\big] - 2\frac{\intd}{\intd\tau}\mE_{(\tilde\bY(\tau), \tilde\bY(s))}\big[(\nabla\log q_{T_g - \tau}(\tilde{\bY}(\tau) | \bc ))^T\nabla\log q_{T_g - s}(\tilde{\bY}(s) | \bc )\big]\\
    &= \tilde g^2(\tau)\mE_{\tilde\bY(\tau)}\big[\|\nabla^2\log q_{T_g - \tau}(\tilde\bY(\tau) | \bc)\|^2_F\big] - 2\tilde f(\tau)\mE_{\tilde\bY(\tau)}\big[\|\nabla\log q_{T_g - \tau}(\tilde{\bY}(\tau) | \bc ) ||^2\\
    &\qquad + 2\mE_{(\tilde\bY(\tau), \tilde\bY(s))}\big[\tilde f(\tau)(\nabla\log q_{T_g - s}(\tilde{\bY}(s) | \bc ))^T\nabla\log q_{T_g - \tau}(\tilde\bY(\tau) | \bc) \big]\\
    &= \tilde g^2(\tau)\mE_{\tilde\bY(\tau)}\big[\|\nabla^2\log q_{T_g - \tau}(\tilde\bY(\tau) | \bc)\|^2_F\big] \\
    &\qquad - 2\tilde f(\tau)\mE_{(\tilde\bY(\tau), \tilde\bY(s))}\big[\|\nabla\log q_{T_g - \tau}(\tilde{\bY}(\tau) | \bc ) - \nabla\log q_{T_g - s}(\tilde{\bY}(s) | \bc )||^2\big]\\
    &\qquad + 2\tilde f(\tau)\mE_{(\tilde\bY(\tau), \tilde\bY(s))} \big[(\nabla\log q_{T_g - s}(\tilde{\bY}(s) | \bc ) - \nabla\log q_{T_g - \tau}(\tilde{\bY}(\tau) | \bc ))^T\nabla\log q_{T_g - s}(\tilde{\bY}(s) | \bc )\big].
\end{align*}
However,  by Young's inequality,
\begin{align*}
    &\mE_{(\tilde\bY(\tau), \tilde\bY(s))}\big[(\nabla\log q_{T_g - s}(\tilde{\bY}(\tau) | \bc ) - \nabla\log q_{T_g - \tau}(\tilde{\bY}(s) | \bc ))^T\nabla\log q_{T_g - s}(\tilde{\bY}(s) | \bc )\big]\\
    &\le \frac{1}{2}\bigg[\mE_{(\tilde\bY(\tau), \tilde\bY(s))}\big[\|\nabla\log q_{T_g - \tau}(\tilde{\bY}(\tau) | \bc ) - \nabla\log q_{T_g - s}(\tilde{\bY}(s) | \bc )\|^2\big] + \mE_{\tilde\bY(s)}\big[\|\nabla\log q_{T_g - s}(\tilde{\bY}(s) | \bc )\|^2\big]\bigg].
\end{align*}
Therefore, we have
\begin{align*}
    &\frac{\intd}{\intd\tau}\mE_{(\tilde\bY(\tau), \tilde\bY(s))} \big[\|\nabla\log q_{T_g - \tau}(\tilde{\bY}(\tau) | \bc ) - \nabla\log q_{T_g - s}(\tilde{\bY}(s) | \bc )\|^2\big]\\
    &\le \tilde f(\tau)\mE_{\tilde\bY(s)}\big[\|\nabla\log q_{T_g - s}(\tilde{\bY}(s) | \bc )\|^2\big] + \tilde g^2(\tau)\mE_{\tilde\bY(\tau)}\big[\|\nabla^2\log q_{T_g - \tau}(\tilde\bY(\tau) | \bc)\|^2_F\big].
\end{align*}
Let $s=\tau_k$ and $\tau\in[\tau_k, \tau_{k + 1}]$. Integrate both sides of the above equation from $\tau_k$ to $\tau$ and use Lemma \ref{lemma: score_function_norm_bound} and Lemma \ref{lemma: hessian_score_bound} to obtain
\begin{align*}
    &\mE_{(\tilde\bY(\tau), \tilde\bY(\tau_k))}\big[\|\nabla\log q_{T_g - \tau}(\tilde{\bY}(\tau) | \bc ) - \nabla\log q_{T_g - \tau_k}(\tilde{\bY}(\tau_k) | \bc )\|^2\big]\\
    &\le \int_{\tau_k}^{\tau}\tilde f(u)\mE_{\tilde\bY(\tau_k)}\big[\|\nabla\log q_{T_g - \tau_k}(\tilde{\bY}(\tau_k) | \bc )\|^2\big] + \tilde g^2(u)\mE_{\tilde\bY(u)}\big[\|\nabla^2\log q_{T_g - u}(\tilde\bY(u) | \bc)\|^2_F\big]\intd u\\
    &\le \int_{\tau_k}^{\tau}\tilde f(u)d\sigma^{-2}_{T_g - \tau_k} + \tilde g^2(u)d\sigma^{-4}_{T_g - u} - \frac{\intd}{\intd r}\big(\sigma_{T - r}^{-4}\mE_{\tilde\bY(r)\sim q_{T_g - r}(\cdot | \bc)}[\text{trace}(\bSimga_{T_g - r})]\big)\big|_{r=u} \intd u\\
    &\le \int_{\tau_k}^{\tau}\tilde f(u)d\sigma^{-2}_{T_g - \tau_k} + \tilde g^2(u)d\sigma^{-4}_{T_g - u} \intd u \\
    &\qquad + \sigma_{T - \tau_k}^{-4}\mE_{\bY(T_g - \tau_k)}[\text{trace}(\bSimga_{T_g - \tau_k})] - \sigma_{T - \tau}^{-4}\mE_{\bY(T_g - \tau)}[\text{trace}(\bSimga_{T_g - \tau})],
\end{align*}
where we omit to write out the dependency of $\bSimga_\tau$ on $\bY(\tau)$ and $\bc$ for simplicity.

Finally, for the forward process $\bY$ in \eqref{eq: forward_SDE_DDPM}, since $\tilde\bY(\tau) = \bY(T_g - \tau)$ for any $\tau\in[0, T_g]$, we have for $\tau >0$ and $\tau\in[\tau_{k - 1}, \tau_k]$,
\begin{align*}
    &\mE_{(\bY(\tau), \bY(\tau_k))}\big[\|\nabla\log q_{\tau}(\bY(\tau) | \bc) - \nabla\log q_{\tau_k}(\bY(\tau_k) | \bc ) \|^2\big]\\
    &\le \int_{\tau}^{\tau_k} f(s)d\sigma^{-2}_{\tau_k} + g^2(s)d\sigma^{-4}_{s} \intd s + \big(\sigma_{\tau_k}^{-4}\mE_{\bY(\tau_k)}[\text{trace}(\bSimga_{\tau_k})] - \sigma_{\tau}^{-4}\mE_{\bY(\tau)}[\text{trace}(\bSimga_{\tau})]\big).
\end{align*}
This completes our proof.
\end{proof}

\subsection{Proof of Lemma \ref{lemma: BM_discretization_error_bound}}\label{sec: BM_discrete_error}

\begin{proof}[Proof of Lemma \ref{lemma: BM_discretization_error_bound}]
    For a given condition $\bc$, recall the definition of $\tau_{lipz}(\bc), k_0(\bc)$ and $k_1$ in the proof of Lemma \ref{lemma: score_discretization_error_bound} in Appendix \ref{sec: score_discretization_error}. By Lemma 14 and Lemma 21 of \cite{chen2023improved} and our Lemma \ref{lemma: score_function_norm_bound}, we can derive
\begin{align*}
    &\sum_{k=1}^{K}\int_{\tau_{k - 1}}^{\tau_{k}}[g^2(\tau_k) - g^2(\tau)]^2\mE_{\bY(\tau)\sim q_{\tau}(\cdot | \bc)} \big[\big\|\nabla\log q_\tau(\bY(\tau) | \bc)\big\|^2\big] \intd\tau\\
    &\lesssim \sum_{k=1}^{K}\int_{\tau_{k - 1}}^{\tau_{k}}(\Delta\tau)^2\mE_{\bY(\tau)\sim q_{\tau}(\cdot | \bc)} \big[\big\|\nabla\log q_\tau(\bY(\tau) | \bc)\big\|^2\big] \intd\tau\\
    &\lesssim \sum_{k=1}^{k_0(\bc) - 1} \bigg[\int_{\tau_{k - 1}}^{\tau_{k}}(\Delta\tau)^2\cdot dL(\bc)\lambda_\tau^{-1} \intd\tau \bigg] + \int_{\tau_{k_0(\bc) - 1}}^{\tau_{lipz}(\bc)}(\Delta\tau)^2\cdot dL(\bc)\lambda_\tau^{-1} \intd\tau \\
    &\qquad + \int_{\tau_{lipz}(\bc)}^{\tau_{k_0(\bc)}}(\Delta\tau)^2\cdot d\sigma^{-2}_\tau \intd\tau + \sum_{k_0(\bc) + 1}^K\bigg[\int_{\tau_{k - 1}}^{\tau_{k}}(\Delta\tau)^2\cdot d\sigma^{-2}_\tau \intd\tau\bigg]\\
    &\overset{\eqref{eq: expoenetial_approximation} }{\lesssim} \int_{0}^{\tau_{lipz}(\bc)} d(\Delta\tau)^2L(\bc)(1 + a\tau^2/4 + b\tau/2) \intd\tau + \int_{\tau_{lipz}(\bc)}^{\tau_{k_1}} \frac{d(\Delta\tau)^2}{a\tau^2/2 + b\tau} \intd\tau + \int_{\tau_{k_1}}^{T_g} d(\Delta\tau)^2 \intd\tau\\
    &\lesssim d(\Delta\tau)^2L(\bc)\cdot \tau_{lipz}(\bc) + d(\Delta\tau)^2\int_{\tau_{lipz}(\bc)}^{\tau_{k_1}}\bigg(\frac{1}{\tau} + \frac{1}{\tau^2} \bigg)\intd\tau + d(\Delta\tau)^2T_g\\
    &\overset{\eqref{eq: lipz_index_approximation}}{\lesssim} d(\Delta\tau)^2\bigg(L(\bc) + 2L_2(\bc) + 1/2 + \ln\big(2L_2(\bc) + 1/2\big) + T_g \bigg)\\
    &\overset{\eqref{eq: lipschitz_constant}}{\lesssim} d(\Delta\tau)^2\bigg(L(\bc) + T_g + 1 \bigg).
\end{align*}
Taking expectation on both sides w.r.t. $\bc\sim\bC$, we obtain
\begin{align*}
    &\mE_{\bc\sim\bC}\bigg[\sum_{k=1}^{K}\int_{\tau_{k - 1}}^{\tau_{k}}[g^2(\tau_k) - g^2(\tau)]^2\mE_{\bY(\tau)\sim q_{\tau}(\cdot | \bc)} \big[\big\|\nabla\log q_\tau(\bY(\tau) | \bc)\big\|^2\big] \intd\tau\bigg]\\
    &\lesssim d(\Delta\tau)^2\mE_{\bc\sim\bC}\bigg[L(\bc) + T_g + 1\bigg]\\
    &\lesssim d(\Delta\tau)^2[L_1 + T_g + 1].
\end{align*}
The proof is therefore complete.
\end{proof}

\section{Additional Proofs}%\label{sec: technical_details}
In this section, we prove Lemmas \ref{lemma: KL_dt} and  \ref{lemma: hessian_score_bound} that are used in Appendix \ref{sec: convergence_proof} for the proof of Proposition \ref{prop: condition_diffusion_KL_bound}.

\subsection{Proof of Lemma \ref{lemma: KL_dt}}\label{sec: KL_dt_proof}

\begin{proof}[Proof of Lemma \ref{lemma: KL_dt}]
Denote the following drift functions of \eqref{eq: piece_reverse_SDE} and \eqref{eq: piece_reverse_SDE_simulation} for simplicity:
\begin{align*}
    \tilde F(\by, \tau) := \tilde f(\tau)\by + \tilde g^2(\tau) \nabla\log q_{T_g - \tau}(\by | \bc),\quad \tilde F_\theta(\tilde\bz):= \tilde f(\tau_k)\tilde\bz + \tilde g^2(\tau_k) s_\theta(T_g - \tau_k, \tilde\bz, \bc).
\end{align*}

We first prove Lemma \ref{lemma: KL_dt}(a). Consider an auxiliary process $\{\tilde\bY^a(\tau): \tau\in(\tau_k, \tau_{k + 1}]\}$, which also starts from $\tilde{\bY}^a(\tau_k) = \tilde\bz$, with the dynamic
\begin{align}
    \intd\tilde{\bY}^a(\tau) &= [\tilde f(\tau_k)\tilde\bz + \tilde g^2(\tau_k) \nabla\log q_{T_g - \tau_k}(\tilde \bz | \bc)]\intd\tau + \tilde g(\tau) \intd\tilde\bB(\tau):=\tilde F_a(\tilde\bz)\intd\tau + \tilde g(\tau) \intd \tilde\bB(\tau). \label{eq: auxiliary_reverse_SDE}
\end{align}
Denote by $\tilde p_{\tau | \tau_k}^a(\cdot | \tilde\bz)$ the conditional density of $\tilde{\bY}^a(\tau)$  given $\tilde{\bY}^a(\tau_k) = \tilde\bz$. Then, $\tilde p_{\tau | \tau_k}^a(\cdot | \tilde\bz)$ is a Gaussian density with mean $\tilde \bz + (\tau - \tau_k)\tilde F_a(\tilde\bz)$ and covariance $\int_{\tau_k}^{\tau}\tilde g^2(s)\intd s\cdot I_d$.

Note that, for any $\tau\in(\tau_k, \tau_{k + 1}]$ and $\tilde\bz\in\mR^d$,
\begin{align}
    \KL\big(\tilde p_{\tau | \tau_k}(\cdot | \tilde\bz) || \tilde p^\theta_{\tau | \tau_k}(\cdot | \tilde\bz)\big)  &= \int_{\mR^d} \tilde p_{\tau | \tau_k}(\by | \tilde\bz) \log\frac{\tilde p_{\tau | \tau_k}(\by | \tilde\bz)}{\tilde p_{\tau | \tau_k}^\theta(\by | \tilde\bz)}\intd\by\nonumber\\
    &= \KL\big(\tilde p_{\tau | \tau_k}(\cdot | \tilde\bz) || \tilde p_{\tau | \tau_k}^a(\cdot | \tilde\bz)\big) + \int_{\mR^d} \tilde p_{\tau | \tau_k}(\by | \tilde\bz) \log\frac{\tilde p_{\tau | \tau_k}^a(\by | \tilde\bz)}{\tilde p^\theta_{\tau | \tau_k}(\by | \tilde\bz)}\intd\by. \label{eq: limit_KL_decomposion}
\end{align}

Using a similar argument as in the proof of Lemma 7(2) in \cite{chen2023improved}, we can obtain that, for almost every $\tilde\bz\in\mR^d$,
\begin{align*}
    \lim_{\tau\to\tau_k+}\KL(\tilde p_{\tau | \tau_k}(\cdot | \tilde\bz)|| \tilde p^a_{\tau | \tau_k}(\cdot | \tilde\bz)) = 0.
\end{align*}
It follows that
\begin{align}\label{eq: limit_KL_auxiliary}
    \mE_{\tilde\bY(\tau_k)\sim \tilde p_{\tau_k}}\bigg[\lim_{\tau\to\tau_k+}\KL(\tilde p_{\tau | \tau_k}(\cdot | \tilde\bY(\tau_k))|| \tilde p^a_{\tau | \tau_k}(\cdot | \tilde\bY(\tau_k)))\bigg] = 0.
\end{align}

On the other hand, both $\tilde p^a_{\tau | \tau_k}(\cdot | \tilde\bz)$ and $\tilde p^\theta_{\tau | \tau_k}(\cdot | \tilde\bz)$ are Gaussian densities, and  $\int^{\tau}_{\tau_k}\tilde g^2(u) \intd u =[\tilde g^2(\tau_k) - a(\tau - \tau_k)/2](\tau - \tau_k) \le \tilde g^2(\tau_k)(\tau - \tau_k)$. Then, for every $\tilde\bz\in\mR^d$,
\begin{align*}
    &\lim_{\tau \to \tau_k+}\int_{\mR^d} \tilde p_{\tau | \tau_k}(\by | \tilde\bz) \log\frac{\tilde p_{\tau | \tau_k}^a(\by | \tilde\bz)}{\tilde p^\theta_{\tau | \tau_k}(\by | \tilde\bz)}\intd\by\nonumber\\
    &= \lim_{\tau \to \tau_k+} \int_{\mR^d} \tilde p_{\tau | \tau_k}(\by | \tilde\bz)\bigg[ - \frac{\|\by - \tilde\bz - (\tau - \tau_k)\tilde F_a(\tilde\bz)\|^2}{\int^{\tau}_{\tau_k}\tilde g^2(u)\intd u} + \frac{\|\by - \tilde\bz - (\tau - \tau_k)\tilde F_\theta(\tilde\bz)\|^2}{\tilde g^2(\tau_k)(\tau - \tau_k)}\bigg]\intd\by\nonumber\\
    &= \lim_{\tau \to \tau_k+} \int_{\mR^d} \tilde p_{\tau | \tau_k}(\by | \tilde\bz) \frac{\tilde g^2(\tau_k) - a(\tau - \tau_k)/2 - \tilde g^2(\tau_k)}{[\tilde g^2(\tau_k) - a(\tau - \tau_k)/2]\cdot \tilde g^2(\tau_k)(\tau - \tau_k)}\|\by - \tilde\bz\|^2\intd\by\nonumber\\
    &\qquad + \lim_{\tau \to \tau_k+}\int_{\mR^d}2\tilde p_{\tau | \tau_k}(\by | \tilde\bz)\bigg[\frac{\tilde F_a(\tilde\bz)}{\tilde g^2(\tau_k) - a(\tau - \tau_k)/2} - \frac{\tilde F_\theta(\tilde\bz)}{\tilde g^2(\tau_k)}\bigg]^T(\by - \tilde\bz)\intd\by\nonumber\\
    &\qquad + \lim_{\tau \to \tau_k+}\int_{\mR^d}\tilde p_{\tau | \tau_k}(\by | \tilde\bz)\bigg[\frac{\|\tilde F_\theta(\tilde\bz)\|^2(\tau - \tau_k)}{\tilde g^2(\tau_k)} - \frac{\|\tilde F_a(\tilde\bz)\|^2(\tau -\tau_k)}{\tilde g^2(\tau_k) - a(\tau - \tau_k)/2}\bigg]\intd\by\nonumber\\
    &\le  \lim_{\tau \to \tau_k+}\int_{\mR^d}2\tilde p_{\tau | \tau_k}(\by | \tilde\bz)\bigg[\frac{\tilde F_a(\tilde\bz)}{\tilde g^2(\tau_k) - a(\tau - \tau_k)/2} - \frac{\tilde F_\theta(\tilde\bz)}{\tilde g^2(\tau_k)}\bigg]^T(\by - \tilde\bz)\intd\by\nonumber\\
    &\le 2\lim_{\tau \to \tau_k+} \bigg\|\frac{\tilde F_a(\tilde\bz)}{\tilde g^2(\tau_k) - a(\tau - \tau_k)/2} - \frac{\tilde F_\theta(\tilde\bz)}{\tilde g^2(\tau_k)}\bigg\| \big(\mE_{\tilde \bY(\tau)\sim \tilde p_{\tau | \tau_k}(\cdot | \tilde\bz)} \big[\|\tilde\bY(\tau) - \tilde\bz\|^2\big] \big)^{1/2}\nonumber\\
    &= 0.
\end{align*}
It follows that
\begin{align}
    \mE_{\tilde\bY(\tau_k)\sim \tilde p_{\tau_k}}\bigg[\lim_{\tau \to \tau_k+}\int_{\mR^d} \tilde p_{\tau | \tau_k}(\by | \tilde\bY(\tau_k)) \log\frac{\tilde p_{\tau | \tau_k}^a(\by | \tilde\bY(\tau_k))}{\tilde p^\theta_{\tau | \tau_k}(\by | \tilde\bY(\tau_k))}\intd\by\bigg] \le 0.\label{eq: limit_KL_Gaussian}
\end{align}
Combining \eqref{eq: limit_KL_decomposion}, \eqref{eq: limit_KL_auxiliary} and \eqref{eq: limit_KL_Gaussian}, and using the non-negativity property of KL-divergence, we obtain
\begin{align*}%\label{eq: limit_KL}
        \mE_{\tilde\bY(\tau_k)\sim\tilde p_{\tau_k}} \bigg[\lim_{\tau \to \tau_k+}\KL\big(\tilde p_{\tau | \tau_k}(\cdot | \tilde\bY(\tau_k)) || \tilde p^\theta_{\tau | \tau_k}(\cdot | \tilde\bY(\tau_k))\big)\bigg]= 0,
        % &\le 2\bigg[\big(\mE_{\bY(0)\sim q_0(\cdot | \bc)}[\|\bY(0)\|^2] + d\big)\cdot \mE_{\tilde\bY(\tau_k)}\big\|\nabla\log q_{T_g - \tau_k}(\tilde\bY(\tau_k)|\bc) -  s_\theta(T_g - \tau_k, \tilde\bY(\tau_k), \bc)\big\|^2\bigg]^{1/2}.
\end{align*}
which proves Lemma \ref{lemma: KL_dt}(a).

Next, we consider bounding $\frac{\intd}{\intd\tau}\KL(\tilde p_{\tau | \tau_k}(\cdot | \tilde\bz)|| \tilde p^\theta_{\tau | \tau_k}(\cdot | \tilde\bz)$ for Lemma \ref{lemma: KL_dt}(b). For any given $\tilde\bz\in\mR^d$, we use the following notations for simplicity,
\begin{align*}
    \bar p_\tau:= \tilde p_{\tau | \tau_k}(\cdot | \tilde\bz), \quad \bar p^\theta_\tau:= \tilde p^\theta_{\tau | \tau_k}(\cdot | \tilde\bz).
\end{align*}

Using the same argument as in the proof of Lemma 6 in \cite{chen2023improved}, we have 
\begin{align}
    \frac{\intd}{\intd\tau}\KL(\bar p_\tau || \bar p^\theta_\tau) = \int_{\by\in\mR^d} \log \frac{\bar p_\tau}{\bar p_\tau^\theta}\cdot \frac{\partial \bar p_\tau}{\partial\tau}(\by)\intd \by - \int_{\by\in\mR^d} \frac{\bar p_\tau}{\bar p_\tau^\theta}\cdot \frac{\partial \bar p^\theta_\tau}{\partial\tau}(\by)\intd \by. \label{eq: density_KL_dt}
\end{align}
Applying the Fokker-Planck equation to \eqref{eq: piece_reverse_SDE} and \eqref{eq: piece_reverse_SDE_simulation}, we get the evolutions of $\bar p_\tau$ and $\bar p^\theta_\tau$ as
\begin{align*}
    \frac{\partial \bar p_\tau}{\partial\tau}(\by) &= \nabla\cdot\bigg[-\tilde F(\by, \tau) \bar p_\tau(\by) + \frac{1}{2}\tilde g^2(\tau)\nabla \bar p_\tau(\by)\bigg],\\
    \frac{\partial \bar p^\theta_\tau}{\partial\tau}(\by) &= \nabla\cdot\bigg[-\tilde F_\theta(\tilde\bz) \bar p^\theta_\tau(\by) + \frac{1}{2}\tilde g^2(\tau_k)\nabla \bar p^\theta_\tau(\by)\bigg].
\end{align*}

It again follows from the Fokker-Planck equation that the first term of \eqref{eq: density_KL_dt} can be expressed as
\begin{align*}
    \int_{\by\in\mR^d} \log \frac{\bar p_\tau}{\bar p_\tau^\theta}\cdot \frac{\partial \bar p_\tau}{\partial\tau}(\by)\intd \by &= \int_{\by\in\mR^d} \bar p_\tau(\by) [\tilde F(\by, \tau)]^T\nabla\log\frac{\bar p_\tau}{\bar p_\tau^\theta}(\by)\intd \by  - \frac{\tilde g^2(\tau)}{2}\int_{\by\in\mR^d} [\nabla\bar p_\tau(\by)]^T\nabla\log\frac{\bar p_\tau}{\bar p_\tau^\theta}(\by) \intd \by,
\end{align*}
and the second term of \eqref{eq: density_KL_dt}  as
\begin{align*}
    \int_{\by\in\mR^d} \frac{\bar p_\tau}{\bar p_\tau^\theta}\cdot \frac{\partial \bar p^\theta_\tau}{\partial\tau}(\by)\intd \by = \int_{\by\in\mR^d} \bar p^\theta_\tau(\by) [\tilde F_\theta(\tilde\bz)]^T \nabla\frac{\bar p_\tau}{\bar p_\tau^\theta}(\by) \intd \by - \frac{\tilde g^2(\tau_k)}{2}\int_{\by\in\mR^d} [\nabla\bar p^\theta_\tau(\by)]^T\nabla\frac{\bar p_\tau}{\bar p_\tau^\theta}(\by)\intd \by.
\end{align*}

We compute
\begin{align}
    &\int_{\by\in\mR^d} \bar p_\tau(\by) [\tilde F(\by, \tau)]^T\nabla\log\frac{\bar p_\tau}{\bar p_\tau^\theta}(\by)\intd \by - \int_{\by\in\mR^d} \bar p^\theta_\tau(\by) \tilde F_\theta(\tilde\bz)^T\nabla\frac{\bar p_\tau}{\bar p_\tau^\theta}(\by) \intd \by \nonumber\\
    &= \int_{\by\in\mR^d} \bar p_\tau(\by) [\tilde F(\by, \tau)]^T\nabla\log\frac{\bar p_\tau}{\bar p_\tau^\theta}(\by)\intd \by - \int_{\by\in\mR^d} \bar p_\tau(\by) \tilde F_\theta(\tilde\bz)^T\nabla\log\frac{\bar p_\tau}{\bar p_\tau^\theta}(\by) \intd \by\nonumber\\
    &= \int_{\by\in\mR^d} \bar p_\tau(\by) [\tilde F(\by, \tau) - \tilde F_\theta(\tilde\bz)]^T\nabla\log\frac{\bar p_\tau}{\bar p_\tau^\theta}(\by)\intd \by \nonumber\\
    &\le \int_{\by\in\mR^d} \bar p_\tau(\by) \bigg[\frac{1}{C_{KL}}\big\|\tilde F(\by, \tau) - \tilde F_\theta(\tilde\bz)\big\|^2 + \frac{C_\KL}{4}\bigg\|\nabla\log\frac{\bar p_\tau}{\bar p_\tau^\theta}(\by)\bigg\|^2\bigg]\intd \by \nonumber\\
    &= \frac{1}{C_{KL}}\mE_{\bar\bY(\tau)\sim\bar p_\tau} \big[\big\|\tilde F(\bar\bY(\tau), \tau) - \tilde F_\theta(\tilde\bz)\big\|^2\big] + \frac{C_\KL}{4}\mE_{\bar\bY(\tau)\sim\bar p_\tau} \bigg[\bigg\|\nabla\log\frac{\bar p_\tau(\bar\bY(\tau))}{\bar p_\tau^\theta(\bar\bY(\tau))}\bigg\|^2\bigg], \label{eq: KL_dt_part1}
\end{align}
where the inequality is from the fact that $\bx^T\by \le \frac{1}{C_{\KL}}\|\bx\|^2 + \frac{C_{\KL}}{4}\|\by\|^2$ for any $\bx, \by\in\mR^d$ and any fixed constant $C_{\KL} > 0$.

In addition, we have
\begin{align}
    &\frac{\tilde g^2(\tau_k)}{2}\int_{\by\in\mR^d} [\nabla\bar p^\theta_\tau(\by)]^T\nabla\frac{\bar p_\tau}{\bar p_\tau^\theta}(\by)\intd \by - \frac{\tilde g^2(\tau)}{2}\int_{\by\in\mR^d} [\nabla\bar p_\tau(\by)]^T\nabla\log\frac{\bar p_\tau}{\bar p_\tau^\theta}(\by) \intd \by \nonumber\\
    &= \frac{\tilde g^2(\tau_k)}{2}\int_{\by\in\mR^d} \bar p^\theta_\tau(\by)[\nabla\log\bar p^\theta_\tau(\by)]^T\nabla\frac{\bar p_\tau}{\bar p_\tau^\theta}(\by)\intd \by  - \frac{\tilde g^2(\tau)}{2}\int_{\by\in\mR^d}\bar  p_\tau(\by)[\nabla\log\bar p_\tau(\by)]^T\nabla\log\frac{\bar p_\tau}{\bar p_\tau^\theta}(\by) \intd \by\nonumber\\
    &= \frac{\tilde g^2(\tau_k)}{2}\int_{\by\in\mR^d}\bar p_\tau(\by)[\nabla\log\bar p^\theta_\tau(\by)]^T\nabla\log\frac{\bar p_\tau}{\bar p_\tau^\theta}(\by)\intd \by - \frac{\tilde g^2(\tau)}{2}\int_{\by\in\mR^d} \bar p_\tau(\by)[\nabla\log\bar p_\tau(\by)]^T\nabla\log\frac{\bar p_\tau}{\bar p_\tau^\theta}(\by) \intd \by\nonumber\\
    &= -\frac{1}{2}\int_{\by\in\mR^d} \bar p_\tau(\by)[\tilde g^2(\tau)\nabla\log\bar p_\tau(\by) - \tilde g^2(\tau_k)\nabla\log\bar p^\theta_\tau(\by)]^T \nabla\log\frac{\bar p_\tau}{\bar p_\tau^\theta}(\by)\intd \by\nonumber\\
    &= \frac{1}{2}\int_{\by\in\mR^d} \bar p_\tau(\by)[\tilde g^2(\tau_k) - \tilde g^2(\tau)][\nabla\log\bar p_\tau(\by)]^T \nabla\log\frac{\bar p_\tau}{\bar p_\tau^\theta}(\by)\intd \by  - \frac{1}{2}\int_{\by\in\mR^d} \bar p_\tau(\by) \tilde g^2(\tau_k)\bigg\|\nabla\log\frac{\bar p_\tau}{\bar p_\tau^\theta}(\by)\bigg\|^2\intd \by\nonumber\\
    &\le \frac{1}{4C_{\KL}}\int_{\by\in\mR^d} \bar p_\tau(\by)[\tilde g^2(\tau_k) - \tilde g^2(\tau)]^2\big\|\nabla\log\bar p_\tau(\by)\big\|^2\intd\by \nonumber\\
    &\qquad + \bigg[\frac{C_{\KL}}{4} - \frac{1}{2} \tilde g^2(\tau_k)\bigg] \int_{\by\in\mR^d} \bar p_\tau(\by) \bigg\|\nabla\log\frac{\bar p_\tau}{\bar p_\tau^\theta}(\by)\bigg\|^2\intd \by\nonumber\\
    &= \frac{[\tilde g^2(\tau_k) - \tilde g^2(\tau)]^2}{4C_{\KL}} \mE_{\bar\bY(\tau) \sim \bar p_\tau} \big[\big\|\nabla\log\bar p_\tau(\bar\bY(\tau))\big\|^2\big] \nonumber\\
    &\qquad + \frac{1}{2}\bigg[\frac{C_{\KL}}{2}- \tilde g^2(\tau_k)\bigg] \mE_{\bar\bY(\tau) \sim \bar p_\tau}\bigg\|\nabla\log\frac{\bar p_\tau(\bar\bY(\tau))}{\bar p_\tau^\theta(\bar\bY(\tau))}\bigg\|^2, \label{eq: KL_dt_part2}
\end{align}
where the inequality is due to $\bx^T\by \le \frac{1}{2C_{\KL}}\|\bx\|^2 + \frac{C_{\KL}}{2}\|\by\|^2$ for any $\bx, \by\in\mR^d$ and any fixed constant $C_{\KL} > 0$.

Finally, by combining \eqref{eq: KL_dt_part1} and \eqref{eq: KL_dt_part2}, we can infer from \eqref{eq: density_KL_dt} that \eqref{eq: density_KL_dt_bound} holds. This completes the proof of Lemma~\ref{lemma: KL_dt}.
\end{proof}

\subsection{Proof of Lemma \ref{lemma: hessian_score_bound} via Stochastic Localization}\label{sec: stochastic_localization}
In this section, we briefly introduce the technique of stochastic localization, based on which we prove Lemma \ref{lemma: hessian_score_bound}. The main idea of the proof follows \cite{bentonnearly}, but we consider diffusion models with time-{\it inhomogeneous} forward processes. Throughout this section, we consider a given condition $\bc$ and the target distribution $p_\tar(\cdot | \bc)$.

We first  discuss stochastic localization; for details, see e.g. \cite{bentonnearly, montanari2023sampling}.
Give a random variable $\xi\sim p_\tar(\cdot | \bc)$, the stochastic localization considers a process $\big(\bU(s)\big)_{s\ge0}$ defined by
\begin{align}\label{eq: stochastic_localization}
    \bU(s) = s\xi + \bar\bB(s),\ \text{for } s \ge 0,
\end{align}
where $\bar\bB$ is a $d$-dimensional standard Brownian motion independent of $\xi$. Denote the posterior covariance of $\xi$ given $\bU(s)$ and $\bc$ by
$\bA_s(\bU(s), \bc) = \cov_{\xi|(\bU(s), \bc)}(\xi)$.
Then, we have the following characterization {\cite[Proposition 2]{bentonnearly}}:
\begin{equation}\label{eq: stochastic_posterior_cov_dt}
    \frac{\intd}{\intd s}\mE_{\bU(s)}\big[\bA_s\big] = -\mE_{\bU(s)}\big[\bA_s^2\big],\quad \text{for } s\ge 0,
\end{equation}
where the expectation is taken w.r.t. the marginal distribution of $\bU(s)$ given $\bc$, and we omit the dependence of $\bA_s$ on $\bU(s)$ and $\bc$ for notational simplicity.

We then have the following result, which shows that the forward OU process $\bY$ can be viewed as a stochastic localization under a time change. The proof of this lemma is deferred to the end of this section.

\begin{lemma}\label{lemma: diffusion_time_change}
    Under condition $\bc$, if we define the process $\bY$ and $\bU$ according to \eqref{eq: forward_SDE_DDPM} and \eqref{eq: stochastic_localization}, respectively, and consider a time change
    \begin{equation}\label{eq: diffusion_localization_time_change}
        \tau(s) = \inf\big\{u: \exp(au^2/2 + bu) - 1 > s^{-1}\big\},
    \end{equation}
    then $\{\bU(s): s\ge 0\}$ and $\{ s\lambda_{\tau(s)}^{-1}\bY(\tau(s)): s\ge 0\}$ have the same law, where $\lambda_\tau$ is defined in \eqref{eq: forward_SDE_coeff}.
\end{lemma}

Denote by $q_{0|\tau}(\cdot | \by(\tau), \bc)$ the posterior distribution of $\bY(0)$ given $\bc$ and $\bY(\tau)=\by(\tau)$. Let the posterior covariance matrix be $\bSimga_\tau(\by(\tau), \bc) := \cov_{\bY(0)\sim q_{0|\tau}(\cdot | \by(\tau), \bc)}\big[\bY(0)\big]$, for which we may omit its dependence on $\by(\tau)$ and $\bc$ when the context is clear. Then, for any $s > 0$, when $\tau = \tau(s)$, it follows from Lemma \ref{lemma: diffusion_time_change} that
\begin{align*}
    \mE_{\bY(\tau)}\big[\bSimga_{\tau}^2\big] &= \int_{\by\in\mR^d} [\cov(\bY(0) | \bY(\tau) = \by, \bc)]^2 \intd P_{\bY(\tau)}(\by)\\
    &= \int_{\by\in\mR^d} [\cov(\bY(0) | s\lambda_\tau^{-1}\bY(\tau) = s\lambda_\tau^{-1}\by, \bc)]^2 \intd P_{s\lambda_\tau^{-1}\bY(\tau)}(s\lambda_\tau^{-1}\by)\\
    & = \int_{\by\in\mR^d} [\cov(\xi | \bU(s) = s\lambda_\tau^{-1}\by, \bc)]^2 \intd P_{\bU(s)}(s\lambda_\tau^{-1}\by)\\
    &=\int_{\bu\in\mR^d} [\cov(\xi | \bU(s) = \bu, \bc)]^2 \intd P_{\bU(s)}(\bu)\\
    &= \mE_{\bU(s)}\big[\bA_s^2\big],
\end{align*}
and similarly $\mE_{\bU(s)}\big[\bA_s\big] = \mE_{\bY(\tau)}\big[\bSimga_{\tau}\big]$, where the expectations involving  $\bSimga_{\tau} = \bSimga_{\tau}(\bY(\tau), \bc)$ are taken w.r.t. $\bY(\tau)\sim q_\tau(\cdot | \bc)$ of the forward process \eqref{eq: forward_SDE_DDPM}, and $P_{\bM}$ is the distribution function of a general random variable $\bM$.
Equation \eqref{eq: stochastic_posterior_cov_dt} yields
\begin{align}\label{eq: posterior_cov_dt}
    \mE_{\bY(\tau)}\big[\bSimga_{\tau}^2\big] = \mE_{\bU(s)}\big[\bA_s^2\big] = -\frac{\intd}{\intd s}\mE_{\bU(s)}\big[\bA_s\big] = -\frac{\intd}{\intd s}\mE_{\bY(\tau(s))}\big[\bSimga_{\tau(s)}\big] = -\frac{\intd \tau(s)}{\intd s}\frac{\intd}{\intd \tau}\mE_{\bY(\tau)}\big[\bSimga_{\tau}\big].
\end{align}
Note that $\sigma_\tau^2 = 1 - \exp(-a\tau^2/2 - b\tau)$, and $\frac{\intd(\sigma_\tau^2)}{\intd \tau} = 2\sigma_\tau \frac{\intd \sigma_\tau}{\intd \tau} := 2\sigma_\tau\dot{\sigma}_\tau = (a\tau + b)\exp(-a\tau^2/2 - b\tau)$. Hence,  for $\tau = \tau(s)$,
\begin{align*}
    s^{-1} = \exp(a\tau^2/2 + b\tau) - 1 = \frac{1}{1 - \sigma_\tau^2} - 1 = \frac{\sigma_\tau^2}{1 - \sigma_\tau^2},
\end{align*}
which indicates that $\frac{\intd s}{\intd \tau} = \big[-2\dot\sigma_\tau\sigma_\tau\cdot \sigma_\tau^2 - 2\dot\sigma_\tau\sigma_\tau(1 - \sigma_\tau^2)\big]/\sigma_\tau^4 = -2\dot\sigma_\tau/\sigma_\tau^3$. Then, by taking the trace of both sides of \eqref{eq: posterior_cov_dt}, we obtain
\begin{align}\label{eq: posterior_cov_trace_dt}
    \mE_{\bY(\tau)}\big[\text{trace}(\bSimga_\tau^2)\big] = -\frac{\intd \tau(s)}{\intd s}\frac{\intd}{\intd \tau}\mE_{\bY(\tau)}\big[\text{trace}(\bSimga_{\tau})\big] = \frac{\sigma_\tau^3}{2\dot\sigma_\tau}\frac{\intd}{\intd \tau}\mE_{\bY(\tau)}\big[\text{trace}(\bSimga_{\tau})\big],
\end{align}
which also implies that $\mE_{\bY(\tau)}\big[\text{trace}(\bSimga_{\tau})\big]$ is increasing in $\tau$ since $\sigma_\tau\dot\sigma_\tau \ge 0$ for all $\tau\ge 0$.

Now we are ready to prove Lemma \ref{lemma: hessian_score_bound}.
\begin{proof}[Proof of Lemma \ref{lemma: hessian_score_bound}]
Recall that $\bY(\tau) | \bY(0)\sim\mathcal{N}(\lambda_\tau\bY(0), \sigma_\tau^2I_d)$. Using a similar argument as in the proof of Lemma 5 of \cite{bentonnearly}, we obtain
\begin{align*}
    \nabla^2\log q_\tau(\by_\tau|\bc) = -\sigma_\tau^{-2}I_d + \sigma_\tau^{-4}\cov_{q_{0|\tau}(\cdot | \by(\tau), \bc)}\big(\by(\tau) - \lambda_\tau\bY(0)\big) = -\sigma_\tau^{-2}I_d + \sigma_\tau^{-4}\lambda_\tau^2\bSimga_\tau.
\end{align*}
Noting that $\|A\|_F^2 = \text{trace}(A^TA)$ for a matrix $A$ and that $1 > \lambda_\tau^2 = \exp(-a\tau^2/2 - b\tau) = 2\sigma_\tau\dot\sigma_\tau/(a\tau + b)$, we can directly compute
\begin{align*}
    &\mE_{\bY(\tau)\sim q_\tau(\cdot | \bc)}\big[\|\nabla^2\log q_\tau(\bY_\tau|\bc)\|_F^2\big] \\
    &= d\sigma_\tau^{-4} - 2\sigma_\tau^{-6}\lambda_\tau^2\mE_{\bY(\tau)}[\text{trace}(\bSimga_\tau)] + \sigma_\tau^{-8}\lambda_\tau^4\mE_{\bY(\tau)}[\text{trace}(\bSimga_\tau^2)]\\
    &\le d\sigma_\tau^{-4} - 2\sigma_\tau^{-6}\lambda_\tau^2\mE_{\bY(\tau)}[\text{trace}(\bSimga_\tau)] + \sigma_\tau^{-8}\lambda_\tau^2\mE_{\bY(\tau)}[\text{trace}(\bSimga_\tau^2)]\\
    &\overset{\eqref{eq: posterior_cov_trace_dt}}{=} d\sigma_\tau^{-4} - \frac{4\sigma_\tau^{-5}\dot\sigma_\tau}{a\tau + b}\mE_{\bY(\tau)}[\text{trace}(\bSimga_\tau)] + \frac{2\sigma_\tau^{-7}\dot\sigma_\tau}{a\tau + b}\cdot \frac{\sigma_\tau^3}{2\dot\sigma_\tau}\frac{\intd}{\intd\tau}\mE_{\bY(\tau)}[\text{trace}(\bSimga_\tau)]\\
    &= d\sigma_\tau^{-4} - \frac{4\sigma_\tau^{-5}\dot\sigma_\tau}{a\tau + b}\mE_{\bY(\tau)}[\text{trace}(\bSimga_\tau)] + \frac{\sigma_\tau^{-4}}{a\tau + b}\frac{\intd}{\intd\tau}\mE_{\bY(\tau))}[\text{trace}(\bSimga_\tau)]\\
    &= d\sigma_\tau^{-4} + \frac{1}{a\tau + b}\cdot \frac{\intd}{\intd\tau}\bigg(\sigma_\tau^{-4}\mE_{\bY(\tau)\sim q_\tau(\cdot | \bc)}[\text{trace}(\bSimga_\tau)]\bigg).
\end{align*}
The proof is complete by virtue of $g^2(\tau) = a\tau + b$.
\end{proof}

\begin{proof}[Proof of Lemma \ref{lemma: diffusion_time_change}]
Letting $\Lambda(\tau) = \exp\big(\int_{0}^\tau f(s)\intd s\big)$ and applying It\^{o}'s formula to the forward SDE \eqref{eq: forward_SDE_DDPM} under the condition $\bc$, we have
\begin{align*}
    \intd [\Lambda(\tau)\bY(\tau)] = f(\tau)\Lambda(\tau)\bY(\tau)\intd\tau + \Lambda(\tau)\big[-f(\tau)\bY(\tau)\intd\tau + g(\tau)\intd \bB(\tau)\big] = \Lambda(\tau)g(\tau)\intd \bB(\tau).
\end{align*}
By the Dambis-Dubins–Schwarz theorem {\citep[Theorem 3.4.6]{karatzas2014brownian}}, there is a process $(\hat\bB(s))_{s\ge 0}$ such that under the time change
\begin{align*}%\label{eq: martingale_time_change}
    \kappa(s) := \inf\bigg\{\tau: \int_{0}^\tau\Lambda^2(u)g^2(u)\intd u > s\bigg\} = \inf\bigg\{\tau: \exp\big(a\tau^2/2 + b\tau\big) - 1> s\bigg\},
\end{align*}
we have
\begin{align*}
    \hat\bB(e^{a\tau^2/2 + b\tau} - 1) = \int_{0}^\tau\Lambda(u)g(u)\intd \bB(u),
\end{align*}
where $(\hat\bB(s))_{s\ge 0}$ is a standard Brownian motion w.r.t. the filtration $\big(\mathcal{F}_{\kappa(s)}\big)_{s\ge0}$, while $(\mathcal{F}_\tau)_{\tau \ge 0}$ is the natural filtration generated by $(\bB(\tau))_{\tau \ge 0}$. Then, for any $s > 0$, we have
\begin{align*}
    \Lambda(\kappa(s))\bY(\kappa(s)) = \bY(0) + \int_{0}^{\kappa(s)}\Lambda(u)g(u)\intd \bB(u) = \bY(0) + \hat\bB(s).
\end{align*}
Now, if we set $\bU(0) = 0$ and
\begin{align*}
    \bU(s) := s\Lambda(\kappa(s^{-1}))\bY(\kappa(s^{-1})) = s\bY(0) + s\hat\bB(s^{-1}),\ \text{for } s > 0,
\end{align*}
then we find that $\bU$ satisfies the definition of the stochastic localization in \eqref{eq: stochastic_localization}, because $\bY(0)\sim p_\tar(\cdot | \bc)$ and the law of $\big(s\hat\bB(s^{-1})\big)_{s\ge 0}$ is the same as the law of $\big(\hat\bB(s)\big)_{s\ge 0}$. The proof is complete by noting that $\kappa(s^{-1}) = \tau(s)$ in \eqref{eq: diffusion_localization_time_change} and $\Lambda(\tau) = \lambda_\tau^{-1}$ in \eqref{eq: forward_SDE_coeff}.
\end{proof}

\section{The $q-$Learning Algorithm for Continuous-Time Mean-Variance Portfolio Selection}\label{sec: application_simulation_more}

In this section, we present the RL algorithm, Algorithm \ref{alg: qL_mv}, implemented for the continuous-time mean-variance portfolio selection problem discussed in Section~\ref{sec:application}. Of the actor-critic type, it is a direct extension of Algorithm~5 in \cite{jia2023q} by taking into account of synthetic price paths. We refer to \cite{jia2023q} for further details about the theoretical foundation and the design principle of this so-called $q-$learning algorithm.

\begin{algorithm}[htbp]
\caption{Offline q-Learning Mean–Variance Algorithm with Synthetic Wealth Simulator}
\textbf{Inputs}: initial state $x(0)$, horizon $T$, time step $\Delta t$, number of steps $N_T = T/\Delta t$, number of episodes $M$, number of real asset price paths $H$, number of synthetic paths (generated at each episode) $H_s$, parameterized value function $J_{\varphi}(\cdot,\cdot; w)$ and policy $\pi_{\psi}(\cdot,\cdot; w)$, initial Lagrange multiplier $w$, learning rate functions $\alpha_{\varphi}(\cdot),\alpha_{\psi}(\cdot), \alpha_w(\cdot)$

\textbf{Required program}: a real wealth simulator $x' = Wealth_{\Delta t}(t,x,a)$ that takes current time--state pair $(t,x)$ and action $a$ as inputs and generates state $x'$ at time $t + \Delta t$ according to \eqref{eq: MV_wealth_process}, a synthetic wealth simulator $x'=Wealth_{\Delta t}(t, x, a; G(\theta))$ based on synthetic paths from a generative model $G(\theta)$, q-function $q_{\psi}(t, x, a; w) = \log\pi_\psi(t, x, a; w)$.

\textbf{Learning procedure}:
\begin{algorithmic}
\FOR{episode $m = 1$ \TO $M$}
\STATE Generate $H_s$ synthetic asset price paths by $G(\theta)$.
\STATE Initialize $n = 0$ and store $x^{(j)}(t_n) \leftarrow  x(0)$ for $j=1,..., H + H_s$.

\WHILE{$n < N_T$}
\FOR{path $j=1$ \TO $H + H_s$}
    \STATE{Generate action $a^{(j)}(t_n)\sim \pi_{\psi}(\cdot|t_n, x^{(j)}(t_n); w)$.

    Compute and store the test functions
    \vspace{-3mm}
    \[
        \xi^{(j)}(t_n)= \frac{\partial J_\varphi}{\partial \varphi}(t_{n},x^{(j)}(t_n); w),\quad \zeta^{(j)}(t_n) = \frac{\partial q_{\psi}}{\partial \psi}(t_{n}, x^{(j)}(t_n), a^{(j)}(t_n); w).
    \]
    \vspace{-5mm}
    \IF{$j \le H$}
        \STATE In the real wealth simulator: $x^{(j)}(t_{n + 1}) = Wealth_{\Delta t}(t_n, x^{(j)}(t_n), a^{(j)}(t_n)).$
    \ELSE
        \STATE In the synthetic wealth simulator: $x^{(j)}(t_{n + 1}) = Wealth_{\Delta t}(t_n, x^{(j)}(t_n), a^{(j)}(t_n); G(\theta)).$
    \ENDIF

    }

\ENDFOR
\STATE Update $n \leftarrow n + 1$.

\ENDWHILE	

\STATE Store the terminal wealth $x^{(j)}(T; m) \leftarrow x^{(j)}(t_{N_T})$ for $j=1,..., H + H_s$.

\STATE For path $j=1, \cdots H + H_s$, compute
\vspace{-3mm}
\begin{align*}
    \Delta \varphi^{(j)} = \sum_{n=0}^{N_T - 1} \xi^{(j)}(t_n) \big[
    J_\varphi(t_{n + 1}, x^{(j)}(t_{n + 1}); w) - J_\varphi(t_n, x^{(j)}(t_n); w) -q_{\psi}(t_n, x^{(j)}(t_n) ,a^{(j)}(t_n); w)\Delta t \big],\\
    \Delta \psi^{(j)} = \sum_{n=0}^{N_T - 1} \zeta^{(j)}(t_n)\big[
    J_\varphi(t_{n + 1}, x^{(j)}(t_{n + 1}); w) - J_\varphi(t_n, x^{(j)}(t_n); w) - q_{\psi}(t_n, x^{(j)}(t_n), a^{(j)}(t_n); w)\Delta t \big] .
\end{align*}

\vspace{-3mm}
\STATE Update $\varphi$ and $\psi$: $\varphi \leftarrow \varphi + \frac{\alpha_{\varphi}(m)}{H + H_s}\sum_{j=1}^{H + H_s}\Delta \varphi^{(j)},\ \psi \leftarrow \psi + \cdot\frac{\alpha_{\psi}(m)}{H + H_s} \sum_{j=1}^{H + H_s}\Delta \psi ^{(j)}$

\vspace{1mm}
\STATE Update $w$ (Lagrange multiplier) every $L$ episodes:
\IF{$m\equiv 0$ mod $L$}
    \STATE{$w\leftarrow w - \alpha_w \frac{1}{L}\sum_{i=m - L + 1}^m \frac{1}{H + H_s}\sum_{j=1}^{H + H_s}x^{(j)}(T; i)$}
\ENDIF

\ENDFOR
\end{algorithmic}
\label{alg: qL_mv}
\end{algorithm}

\section{Experiment Settings}\label{sec: experiment_settings}

In this section we outline the settings of our experiments. We first introduce the neural network structure we use for the score network $s_\theta(\tau_k, \bx, \bc)$, where the Multilayer Perceptron (MLP) is employed as the base model. To utilize the temporal information $\tau_k$ better, we apply the idea of Feature-wise Linear Modulation (FiLM) introduced in \cite{perez2018film}. We use the conditional linear layer $L(\bz, \tau_k)$, which is an element-wise scaling of a linear transformation, to replace the standard linear layer in MLP. Specifically, let $\mathbf{z} \in \mathbb{R}^{d_{\text{in}}}$ be the input vector and $k \in \{1, \dots, K\}$ be a discrete diffusion step. The output of $L(\bz, \tau_k)\in \mathbb{R}^{d_{\text{out}}}$ is given by:
\begin{align*}
    L(\bz, \tau_k) = \boldsymbol{\gamma}(k) \odot (\bW\bz + \mathbf{b}),
\end{align*}
where $\bW \in \mathbb{R}^{d_{\text{out}} \times d_{\text{in}}}$ and $\mathbf{b} \in \mathbb{R}^{d_{\text{out}}}$ are respectively the learnable weight and bias of the linear projection, $\boldsymbol{\gamma}:  \{1, \dots, K\}\to \mathbb{R}^{d_{\text{out}}}$ is a learnable time-step embedding vector, and $\odot$ denotes the Hadamard (element-wise) product. The neural network $s_\theta(\tau_k, \bx, \bc)$ is constructed by the composition of $m$ conditional linear layers $L_1, \cdots L_m$, and $m$ activation functions $A_1, \cdots, A_m$:
\begin{align*}
    s_\theta(\tau_k, \bx, \bc) = A_m\circ L_m(\cdot, \tau_k)\circ A_{m - 1}\circ L_{m - 1}(\cdot, \tau_k)\circ \cdots \circ A_1\circ L_1((\bx, \bc), \tau_k).
\end{align*}

Next, Tables \ref{table: score_network_setting_1d} and \ref{table: score_network_setting_100d} present the hyperparameters used to define the network structure for the 1-dimensional and 100-dimensional examples, respectively.
In addition, Table \ref{table: DDPM_settings_1d} and Table \ref{table: DDPM_settings_100d} display the hyperparameter settings of the diffusion models for the 1-dimensional and 100-dimensional examples, respectively.
Following a standard practice (see e.g. \citealt{songscore, ho2021classifier}), we select checkpoints at regular intervals during training and record the performance metric, which in our case is the KL divergence between samples of discrete observations from real and synthetic SDE paths. Training is halted when the performance metric plateaus, i.e., fails to improve across multiple checkpoints. We then report the minimum KL divergence value achieved over the course of training.

\begin{table}[H]
    % \captionsetup{justification=centering}
    \begin{minipage}{0.45\linewidth}
	\centering
        %\captionsetup{justification=center}
	\begin{tabular}{l | c c c c}
		\toprule
        Layer ID & $d_{\text{in}}$ & $d_{\text{out}}$ & \makecell{Activation\\ function}\\\hline
        1 (Input) & $1 + 1$ & 32 & Softplus\\
		2 (Hidden) & 32 & 32 & Softplus\\
		3 (Hidden) & 32& 16 & Softplus\\
        4 (Hidden) & 16 & 16 & Softplus\\
        5 (Output) & 16 & $1$ & - \\ \bottomrule
	\end{tabular}
	\caption{Neural network structure for the 1-dimensional examples in Sections~\ref{sec: numerical_experiment} and  \ref{sec:application}}
	\label{table: score_network_setting_1d}
	\end{minipage}%
    \hfill
    \begin{minipage}{0.45\linewidth}
	\centering
        %\captionsetup{justification=center}
	\begin{tabular}{l | c c c c}
		\toprule
        Layer ID & $d_{\text{in}}$ & $d_{\text{out}}$ & \makecell{Activation\\ function}\\\hline
        1 (Input) & $100 + 100$ & 256 & Softplus\\
		2 (Hidden) & 256 & 512 & Softplus\\
		3 (Hidden) & 512 & 512 & Softplus\\
        4 (Hidden) & 512 & 128 & Softplus\\
        5 (Output) & 128 & $100$ & - \\ \bottomrule
	\end{tabular}
	\caption{Neural network structure for the 100-dimensional example in Section~\ref{sec: numerical_experiment}}
	\label{table: score_network_setting_100d}
	\end{minipage}%
\end{table}

\begin{table}[htp]
    % \captionsetup{justification=centering}
    \begin{minipage}{0.45\linewidth}
	\centering
        %\captionsetup{justification=center}
	\begin{tabular}{l | l}
		\toprule Hyperparameters & Values\\
		\hline Drift coef. $f(\tau)$ & $\frac{1}{2}(19.9\tau + 0.1)$\\
		Diffusion coef. $g(\tau)$ & $\sqrt{19.9\tau + 0.1}$\\
        Diffusion horizon $T_g$ & 1\\
        Time step $\Delta\tau$& $10^{-2}$\\
        %Number of Episodes & 6000\\
        Batch Size $m$ & $H$\\
		Learning Rate $\alpha(j) $ & $10^{-3}$
        %\makecell[l]{$10^{-3}$ if $j < 4000$,\\ $10^{-5}$ o.w.}
        \\
        \bottomrule
	\end{tabular}
	\caption{Hyperparameters of diffusion models for the 1-dimensional examples, where $H$ is the number of training paths.}
	\label{table: DDPM_settings_1d}
	\end{minipage}%
    \hfill
    \begin{minipage}{0.45\linewidth}
	\centering
        %\captionsetup{justification=center}
	\begin{tabular}{l | l}
		\toprule Hyperparameters & Values\\
		\hline Drift coef. $f(\tau)$ & $\frac{1}{2}(19.9\tau + 0.1)$\\
		Diffusion coef. $g(\tau)$ & $\sqrt{19.9\tau + 0.1}$\\
        Diffusion horizon $T_g$ & 1\\
        Time step $\Delta\tau$& $10^{-3}$\\
        %Number of Episodes & 120,000\\
        Batch Size $m$ & $H$\\
		Learning Rate $\alpha(j) $ & $10^{-3}$
        %\makecell[l]{$10^{-3}$ if $j < 4000$,\\ $10^{-5}$ o.w.}
        \\
        \bottomrule
	\end{tabular}
	\caption{Hyperparameters of diffusion models for the 100-dimensional example, where $H$ is the number of training paths.}
	\label{table: DDPM_settings_100d}
	\end{minipage}%
\end{table}

% We define the conditional linear layer $f(\bx, k)$ as an element-wise scaling of a linear transformation. FiLM: Feature-wise Linear Modulation \cite{perez2018film}. Let $\mathbf{x} \in \mathbb{R}^{d_{\text{in}}}$ denote the input vector and $k \in \{1, \dots, T_g\}$ denote the discrete time step. The output $f(\bx, k)\in \mathbb{R}^{d_{\text{out}}}$ is given by:
% \begin{align*}
%     f(\bx, k) = \boldsymbol{\gamma}(k) \odot (\bW\bx + \mathbf{b}),
% \end{align*}

% where $\bW \in \mathbb{R}^{d_{\text{out}} \times d_{\text{in}}}$ and $\mathbf{b} \in \mathbb{R}^{d_{\text{out}}}$ are the learnable weight and bias of the linear projection, $\boldsymbol{\gamma}:  \{1, \dots, T_g\}\to \mathbb{R}^{d_{\text{out}}}$ is a learnable time-step embedding vector, and $\odot$ denotes the Hadamard (element-wise) product.

%%%%%%%%%%%%%%%%%%%%%%%%%%%%%%%%%%%%%%%%%%%%%%%%%%%%%%%%%%%%

%%%%%%%%%%%%%%%%%%%%%%%%%%%%%%%%%%%%%%%%%%%%%%%%%%%%%%%%%%%%

\end{document}